\documentclass{article}

 \usepackage[main, final]{neurips_2026}

\usepackage[utf8]{inputenc} 
\usepackage[T1]{fontenc}    
\usepackage{hyperref}       
\usepackage{url}            
\usepackage{amsmath}
\usepackage{amssymb}
\usepackage{mathtools}
\usepackage{amsthm}

\usepackage{enumitem}

\usepackage{booktabs}
\usepackage{multirow}
\usepackage{tabularx}
\usepackage{colortbl}
\usepackage{listings}
\usepackage{xcolor}

\definecolor{lightgray}{gray}{0.95}
\usepackage{tcolorbox}
\tcbuselibrary{breakable}
\usepackage{titletoc}
\usepackage{kotex}
\usepackage{subcaption}
\usepackage{pifont}
\newcommand{\cmark}{\ding{51}}
\newcommand{\xmark}{\ding{55}}
\usepackage{verbatim}
\title{Teaching VLMs What to Say, Not How to Reason: \\Rethinking Counterfactual Reasoning \\in Autonomous Driving}

\author{%
  Hayeon Oh and Donghwan Lee \\
  Department of Electrical Engineering \\
  Korea Advanced Institute of Science and Technology \\
  \texttt{\{hahao5252, donghwan\}@kaist.ac.kr} \\
}

\begin{document}

\maketitle

\begin{abstract}
End-to-end autonomous driving systems have been shown to suffer from shortcut learning and causal confusion, limiting their ability to generalize robustly under distribution shift. To mitigate these limitations, recent works based on Vision-Language Models (VLMs) have explored incorporating linguistic reasoning capabilities into driving decision-making, reporting improvements in driving performance. In particular, counterfactual (CF) reasoning has been proposed to enable self-reflective, high-level causal understanding. However, it remains unclear whether the language-based reasoning mitigates shortcut learning or merely shifts it from visual representations to linguistic patterns, creating an illusion of causal understanding. Moreover, existing approaches often conflate intervention-based reasoning with CF reasoning, leaving the effects of genuine CF reasoning unexplored. In this work, we examine what CF reasoning actually induces in VLM-based autonomous driving and observe that it shifts the action distribution toward more conservative behaviors while increasing uncertainty in reasoning and decision-making. This effect persists under reinforcement fine-tuning, suggesting that it is not merely a transient artifact. Furthermore, visual attention analysis shows that CF reasoning does not alter visual grounding, while causal intervention experiments reveal that masking causally relevant and non-causal factors in accident scenarios results in comparable decision change rates. Overall, these results suggest that CF reasoning does not induce genuine causal reasoning grounded in visual evidence. Instead, it acts as a linguistic shortcut that reshapes the action distribution, teaching models what to say but not how to reason, highlighting a gap between language-based reasoning and causal understanding in autonomous driving.
\end{abstract}

\section{Introduction}

Autonomous driving based on the conventional end-to-end paradigm has been consistently shown to suffer from inherent limitations in causal reasoning due to issues such as shortcut learning and causal confusion. To address these challenges, recent studies have explored Vision-Language Models (VLMs), where language-based intermediate reasoning steps such as chain-of-thought (CoT) \cite{e2e-vlm:drivelm, e2e-vlm:drivegpt4} and chain-of-causation (CoC) \cite{causal:e2e-vlm:coc-vla, causal:e2e-vlm:alpamayo} have been reported to improve driving performance. In particular, counterfactual (CF) reasoning \cite{causal:e2e-vlm:omnidrive, causal:cf_vla, robotics:cast, robotics:chop} has been regarded as a promising approach for enabling higher-level causal understanding through self-reflective reasoning, thereby strengthening the expectation that VLMs can acquire genuine causal reasoning capabilities.

However, according to the formal definition in \cite{causal:ladder_of_causation}, CF reasoning defined in prior works \cite{causal:e2e-vlm:omnidrive, causal:cf_vla} often does not correspond to retrospective reasoning that reconstructs past events and reasons about alternative causes. Instead, it is more commonly limited to intervention-level reasoning that considers how actions might change under the current observation \cite{causal:e2e-vlm:simlingo}. In other words, rather than reasoning over unobserved alternative worlds, it is closer to action adjustment conditioned on the current input. Due to this discrepancy, it remains unclear whether the reported performance gains truly originate from causal reasoning.

This raises the fundamental question: what effects does genuine retrospective CF reasoning induce in VLMs for autonomous driving \textbf{(RQ1)}? More specifically, does it induce genuine causal reasoning, or does the introduction of language as an additional modality instead act as a shortcut that reshapes the action distribution \textbf{(RQ2)}?

In this study, we address these questions by analyzing how CF reasoning, as an intermediate step prior to decision-making, influences model behavior and whether it leads to genuine causal understanding. Accordingly, evaluating these aspects requires scenarios with explicit causal factors. Furthermore, single-frame settings are insufficient to capture key causal cues, such as motion dynamics, trajectory evolution, and temporal risk accumulation, making temporal context essential. Accident video datasets provide a suitable setting for this evaluation, as they contain identifiable causal relationships and temporally ordered event progressions. Moreover, they enable retrospective counterfactual queries, such as "if the underlying cause had not been present, the accident would not have occurred," making them a suitable testbed for assessing genuine CF reasoning. 

Beyond scenario design, an appropriate action space and evaluation metrics are required to examine how CF reasoning influences decision-making. Conventional trajectory-based evaluations primarily focus on driving performance, making them less suitable for our objective of analyzing changes in the model's decision-making behavior and its underlying action distribution. In particular, a trajectory lies in a continuous action space, whereas a meta-action represents a discrete driving behavior. Moreover, trajectory is less semantically interpretable than meta-actions, making it difficult to interpret changes in the likelihood of semantically equivalent decisions. Consequently, meta-action provides a more interpretable basis for analyzing changes in reasoning exploration. Therefore, we adopt meta-action as the decision space for analysis, following prior work\cite{e2e-vlm:senna, e2e-vlm:alphadrive, causal:cf_vla, e2e-vlm:driveagent-r1, e2e-vlm:senna-2}. This enables consistent evaluation across likelihood-based analysis using perplexity and direct examination of changes in distributional diversity through policy entropy. Accordingly, we construct a testbed dataset by integrating existing accident video datasets \cite{dataset:ccd, dataset:nexar, dataset:coool} with explicit causal factors, temporal context, and meta-action annotations. Further details of the dataset are provided in Appendix~\ref{appendix:dataset}.

In conclusion, we reveal three key findings. First, we observe that CF reasoning induces a consistent shift toward more conservative action distributions, accompanied by increased entropy in language-based decision outputs, indicating that the model generates more diverse linguistic explanations while its decisions become biased toward conservative directions.The action distribution shift persists even after reinforcement learning fine-tuning (RFT), suggesting that it is not merely a training artifact. Second, we examine changes in visual attention with and without CF reasoning and find no meaningful differences in attention patterns, suggesting that the reasoning process operates independently of visual evidence. Third, in causal intervention experiments, VLMs exhibit similar levels of decision change regardless of whether causal or non-causal objects are masked, unlike humans selectively change their decisions only when causal factors are masked. This indicates that CF reasoning does not enable the model to selectively identify causal factors.

Overall, these results indicate that CF reasoning primarily reshapes action distributions rather than improving the grounding of causal reasoning in visual evidence. We further examine whether this effect extends to general driving scenarios and low-level decisions, as in prior work reporting performance gains from CF reasoning\cite{causal:e2e-vlm:omnidrive}. An ablation on CF reasoning data reveals that models trained with CF reasoning consistently exhibit lower rates of lateral maneuvers than those trained without it, suggesting that this shift also manifests under existing CF reasoning paradigms. Given the relatively low prevalence of lateral maneuvers in the evaluation trajectories, this shift may align with the dataset's action distribution, potentially explaining the reported performance gains. Taken together, our findings indicate that CF reasoning may not induce causal reasoning capabilities, but rather functions as a linguistic shortcut that shifts the existing policy distribution.

\section{Preliminaries}
\label{sec:preliminaries}
\subsection{Limitation of Driving Metrics for Evaluating Reasoning}

Metrics such as displacement error between predicted and expert trajectories \cite{dataset:nusc} and overall driving success rates \cite{dataset:navsim, dataset:bench2drive} are widely used for evaluating autonomous driving systems. However, while these metrics are effective for quantitative performance comparison, they do not directly reflect the underlying reasoning process behind model decisions. Recent studies further show that strong driving performance does not necessarily imply improved visual understanding or causal reasoning ability. In particular, models can achieve competitive results using only non-visual inputs such as ego status and navigation signals \cite{causal:e2e:ad-mlp, e2e-ad:is-ego-status}, and similar behaviors have also been observed in VLM-based approaches \cite{causal:e2e-vlm:omnidrive, causal:e2e-vlm:more_than_meets_the_eye}. These findings suggest that current performance gains may stem from spurious correlations or shortcut patterns in the data rather than genuine causal reasoning, making it difficult to isolate the effect of CF reasoning using standard driving metrics alone. Consequently, it remains unclear whether performance gains reflect improved reasoning or dataset-induced shortcuts, making it difficult to isolate the effect of CF reasoning. To address this issue, we adopt evaluation metrics designed to quantify latent reasoning behavior. Detailed definitions are provided in Appendix~\ref{appendix:implementations}.

\subsection{Metrics for Reasoning Capacity and Exploration}

\paragraph{Pass@k.} 
It measures whether a correct answer is included within the top-k generated outputs, and is commonly used to evaluate a model’s potential capability to solve a given problem \cite{metric:passk, limit-rlvr:does, metric:passk:example:atlas}. Unlike single-prediction metrics such as precision and recall, Pass@k evaluates multiple candidates per input, allowing us to distinguish between generation capability and selection behavior. In particular, Pass@1 reflects the model’s final decision, while larger values of k approximate the set of feasible decisions that the model can generate. This allows us to determine whether CF reasoning introduces new decisions or instead reshapes the distribution of existing actions.

\paragraph{Perplexity.} 
If the distribution of decision outcomes shifts under CF reasoning, it is necessary to determine whether such changes merely reflect a redistribution of probabilities or arise from substantive intervention in the decision generation process. To this end, we additionally analyze perplexity, which measures the likelihood assigned by a model to a given output sequence, and is defined in Appendix \ref{lst:perplexity}. This quantifies how naturally a model generates a given output sequence under its conditional distribution. This aims to distinguish whether the observed action distribution shift is simply a change in output probabilities or a consequence of CF reasoning acting as a conditioning factor in decision making. Specifically, if CF reasoning is merely an auxiliary process unrelated to decision making, both models are expected to assign similar likelihoods to the same Y. In contrast, if CF reasoning substantively contributes to decision generation, the model generating response with CF reasoning may assign a lower likelihood to the same output when evaluated without conditioning on the reasoning process, which would be reflected as higher perplexity.

\paragraph{RFT as a Verification Framework.}
The pass@k and perplexity analyses presented above are conducted on supervised fine-tuned (SFT) models that are instruction-tuned to perform CF reasoning prior to decision making. However, it remains unclear whether the observed behavioral patterns are merely artifacts. To address this question, we employ reinforcement fine-tuning (RFT) as a diagnostic framework. In autonomous driving, RFT enables policy optimization through explicitly designed reward functions that align learning objectives with driving performance metrics \cite{e2e-vlm:autovla, e2e-vlm:alphadrive, e2e-vlm:autodrive-r2, causal:e2e-vlm:alpamayo}. In this work, we adopt GRPO \cite{rlvr:grpo} and its variant GDPO \cite{rlvr:gdpo}, both of which generate $G$ outputs per query and update the model based on relative performance within each group. Their shared objective function is defined as follows:
$$\mathcal{J}(\theta) = \mathbb{E}_{q, \{o_i\}} \left[ \frac{1}{G} \sum_{i=1}^{G} \min \left( w_i A_i, \text{clip}(w_i, 1-\epsilon, 1+\epsilon) A_i \right) - \beta \mathbb{D}_{\mathrm{KL}}(\pi_\theta \| \pi_{\text{ref}}) \right]$$
where $w_i$ denotes the importance sampling ratio and $A_i$ represents the group-relative advantage as defined in Appendix \ref{appendix:implementations}. Importantly, the KL divergence constraint prevents the updated policy $\pi_\theta$ from deviating excessively from the reference policy $\pi_{\text{ref}}$, thereby stabilizing the optimization process and preserving the fundamental characteristics of the original policy \cite{limit-rlvr:does, limit-rlvr:echo_chamber, limit-rlvr:rethinking-reflection}. Thus, RFT can be a surrogate for probing the latent behavioral properties of the SFT reference model. In particular, if the behavioral tendencies induced by CF reasoning persist throughout RFT optimization, this suggests that the observed effects are unlikely to be incidental artifacts of the data distribution, but rather reflect more structured learning signals.

\paragraph{Policy Entropy as a Real-Time Exploration Probe.}
correct English
Policy entropy, defined in Appendix \ref{eq:policy_entropy} and recorded throughout RFT, measures token-level decision diversity during generation and has recently been adopted as a real-time indicator of exploration capacity. \cite{policy-entropy:cui2025entropy, policy-entropy:liu2025prorl}. Higher entropy indicates that the model explores a broader decision space, whereas a sharp decrease in entropy suggests convergence toward specific action patterns, i.e., reduced exploration. In this work, we leverage policy entropy as an additional diagnostic signal for a more structural analysis of the effect of CF reasoning. Specifically, we aim to examine whether CF reasoning structurally alters the model’s exploratory behavior, or whether the previously observed effects are merely incidental artifacts of the SFT training setup. If CF reasoning indeed acts as an inductive bias that expands the model’s decision-making exploration, it should consistently manifest as higher entropy in CF models, in alignment with the observed action distribution shifts and perplexity changes \cite{policy-entropy:reasoning_w_exploration}.

\section{Problem Formulation and Experimental Setup}
We define each sample in the dataset as $Y = (a, e, d, r, y)$, where $a \in \{0,1\}$ denotes accident occurrence, $e \in \{0,1\}$ indicates ego-vehicle involvement, $d$ represents the scene description extracted from the video, $r$ denotes the CF reasoning component, and $y$ corresponds to the final meta-action decision. 
Furthermore, we define three longitudinal meta-actions: \texttt{Stop}, \texttt{Decelerate}, and \texttt{Keep}. These span the full range of longitudinal responses from immediate braking to speed maintenance and are mutually exclusive and collectively exhaustive with respect to longitudinal behavior in accident scenarios, making them appropriate for analyzing how CF reasoning affects the action distribution. To analyze the effect of CF reasoning, we fine-tune VLMs to generate responses in the following structured format: accident occurrence, ego involvement, scene descriptions, CF reasoning, and final decision. This ordering reflects the causal structure of retrospective reasoning: the model first reconstructs what happened and why, then determines the appropriate response, encouraging it to condition its final action on CF reasoning. As a baseline, we fine-tune a model with the same structure but without the CF reasoning component, enabling a controlled comparison of the effects of CF reasoning on decision making.

\section{Analysis of Counterfactual Reasoning}
\label{sec:analysis}

\subsection{Counterfactual Reasoning Induces Behavioral Shifts}
\label{subsec:sft_analysis}


\begin{table*}[h]
\centering
\small
\resizebox{\textwidth}{!}{
\begin{tabular}{l c | cc cc | ccc}
\toprule
\multirow{2}{*}{\textbf{Model}} & \multirow{2}{*}{\textbf{Metric}}
  & \multicolumn{2}{c}{\textbf{Accident Recog.}}
  & \multicolumn{2}{c|}{\textbf{Ego Inclusion}}
  & \multicolumn{3}{c}{\textbf{Decision}} \\
 & & Yes & No & Involv. & Not In. & Stop & Decel. & Keep \\
\midrule

\multirow{6}{*}{\textbf{InternVL2-1B}}
 & P@1 w/o & 47.5 & 64.4 & 45.8 & 64.4 & 20.4 & 29.2 & 78.0 \\
 & P@1 w/  & 60.3 & 50.6 & 60.3 & 50.6 & 26.2 & 24.9 & 67.2 \\
 & $\Delta$
   & \cellcolor{blue!15}+12.8 & \cellcolor{red!15}-13.8
   & \cellcolor{blue!15}+14.5 & \cellcolor{red!15}-13.8
   & \cellcolor{blue!15}+5.8 & \cellcolor{red!15}-4.3
   & \cellcolor{red!15}-10.8 \\
\cmidrule{2-9}
 & P@16 w/o & 99.4 & 99.7 & 99.4 & 99.7 & 87.2 & 90.3 & 99.8 \\
 & P@16 w/  & 99.9 & 99.9 & 99.9 & 99.9 & 93.3 & 93.6 & 100.0 \\
 & $\Delta$
   & \cellcolor{red!15}+0.5  & \cellcolor{blue!15}+0.2
   & \cellcolor{red!15}+0.5  & \cellcolor{blue!15}+0.2
   & \cellcolor{blue!15}+6.1 & \cellcolor{blue!15}+3.3
   & \cellcolor{blue!15}+0.2 \\
\midrule

\multirow{6}{*}{\textbf{InternVL2-2B}}
 & P@1 w/o & 38.8 & 74.1 & 38.8 & 74.1 & 17.3 & 22.8 & 83.2 \\
 & P@1 w/  & 54.3 & 66.1 & 54.3 & 66.1 & 26.5 & 26.2 & 78.8 \\
 & $\Delta$
   & \cellcolor{blue!15}+15.5 & \cellcolor{red!15}-8.0
   & \cellcolor{blue!15}+15.5 & \cellcolor{red!15}-8.0
   & \cellcolor{blue!15}+9.2 & \cellcolor{blue!15}+3.4
   & \cellcolor{red!15}-4.4 \\
\cmidrule{2-9}
 & P@16 w/o & 92.9 & 99.9 & 92.9 & 99.9 & 76.0 & 80.3 & 100.0 \\
 & P@16 w/  & 95.8 & 99.0 & 95.8 & 99.0 & 78.0 & 84.1 & 99.7 \\
 & $\Delta$
   & \cellcolor{blue!15}+2.9  & \cellcolor{red!15}-0.9
   & \cellcolor{blue!15}+2.9  & \cellcolor{red!15}-0.9
   & \cellcolor{blue!15}+2.0 & \cellcolor{blue!15}+3.8
   & \cellcolor{red!15}-0.3 \\
\midrule

\multirow{6}{*}{\textbf{InternVL2.5-1B}}
 & P@1 w/o & 53.1 & 65.9 & 53.1 & 65.9 & 20.1 & 25.4 & 74.3 \\
 & P@1 w/  & 56.1 & 63.2 & 56.1 & 63.2 & 27.8 & 27.4 & 73.8 \\
 & $\Delta$
   & \cellcolor{blue!15}+3.0  & \cellcolor{red!15}-2.7
   & \cellcolor{blue!15}+3.0  & \cellcolor{red!15}-2.7
   & \cellcolor{blue!15}+7.7 & \cellcolor{blue!15}+2.0
   & \cellcolor{red!15}-0.5 \\
\cmidrule{2-9}
 & P@16 w/o & 99.1 & 99.4 & 99.1 & 99.4 & 85.0 & 87.4 & 99.8 \\
 & P@16 w/  & 99.1 & 98.8 & 99.1 & 98.8 & 87.2 & 88.7 & 99.4 \\
 & $\Delta$
   & \cellcolor{blue!15}+0.0  & \cellcolor{red!15}-0.6
   & \cellcolor{blue!15}+0.0  & \cellcolor{red!15}-0.6
   & \cellcolor{blue!15}+2.2 & \cellcolor{blue!15}+1.3
   & \cellcolor{red!15}-0.4 \\
\midrule

\multirow{6}{*}{\textbf{Qwen2.5-VL-3B}}
 & P@1 w/o & 60.9 & 49.4 & 54.3 & 52.7 & 21.7 & 38.2 & 59.3 \\
 & P@1 w/  & 75.6 & 32.6 & 74.6 & 33.0 & 45.7 & 36.7 & 45.1 \\
 & $\Delta$
   & \cellcolor{blue!15}+14.7 & \cellcolor{red!15}-16.8
   & \cellcolor{blue!15}+20.3 & \cellcolor{red!15}-19.7
   & \cellcolor{blue!15}+24.0 & \cellcolor{red!15}-1.5
   & \cellcolor{red!15}-14.2 \\
\cmidrule{2-9}
 & P@16 w/o & 98.5 & 90.3 & 98.1 & 93.3 & 76.7 & 94.9 & 97.5 \\
 & P@16 w/  & 100.0 & 86.0 & 100.0 & 86.5 & 97.1 & 94.6 & 92.5 \\
 & $\Delta$
   & \cellcolor{blue!15}+1.5  & \cellcolor{red!15}-4.3
   & \cellcolor{blue!15}+1.9  & \cellcolor{red!15}-6.8
   & \cellcolor{blue!15}+20.4 & \cellcolor{red!15}-0.3
   & \cellcolor{red!15}-5.0 \\
\midrule

\multirow{6}{*}{\textbf{VideoLLaMA2-7B}}
 & P@1 w/o & 47.9 & 53.3 & 47.9 & 53.3 & 16.9 & 26.7 & 69.7 \\
 & P@1 w/  & 72.3 & 28.7 & 71.7 & 30.1 & 30.0 & 30.0 & 53.8 \\
 & $\Delta$
   & \cellcolor{blue!15}+24.4 & \cellcolor{red!15}-24.6
   & \cellcolor{blue!15}+23.8 & \cellcolor{red!15}-23.2
   & \cellcolor{blue!15}+13.1 & \cellcolor{blue!15}+3.3
   & \cellcolor{red!15}-15.9 \\
\cmidrule{2-9}
 & P@16 w/o & 99.3 & 100.0 & 99.3 & 100.0 & 74.8 & 95.4 & 100.0 \\
 & P@16 w/  & 100.0 & 91.8 & 100.0 & 92.9 & 86.6 & 97.2 & 98.6 \\
 & $\Delta$
   & \cellcolor{blue!15}+0.7  & \cellcolor{red!15}-8.2
   & \cellcolor{blue!15}+0.7  & \cellcolor{red!15}-7.1
   & \cellcolor{blue!15}+11.8 & \cellcolor{blue!15}+1.8
   & \cellcolor{red!15}-1.4 \\
\bottomrule
\end{tabular}
}
\caption{
  Pass@1 and Pass@16 comparison w/ and w/o counterfactual reasoning in SFT.
  \textbf{Conservative} decisions (Stop / Decel.) are grouped under a shared header.
  In the $\Delta$ rows, \colorbox{blue!15}{\textbf{bold blue}} highlights gain
  and \colorbox{red!15}{\textbf{bold red}} highlights reduction,
  together indicating a consistent shift toward conservative decision-making
  when counterfactual reasoning is included.
}
\label{tab:passk_sft}
\end{table*}

In this section, we evaluate the impact of CF reasoning on VLMs during the supervised fine-tuning (SFT). To verify whether the observed findings consistently generalize across different VLMs, we extend our evaluation to a diverse set of models, including InternVL2-1B, InternVL2-2B, InternVL2.5-1B, Qwen2.5-VL-3B, and VideoLLaMA2-7B, with detailed selection criteria provided in Appendix \ref{appendix:model_selection_criteria}.

\paragraph{Action Distribution Shift.} Pass@1 reflects the model’s single-shot decision behavior, capturing how the most confident action changes depending on the presence of CF reasoning. Across most models, CF reasoning consistently shifts predictions toward more conservative actions such as \texttt{Stop} and \texttt{Decelerate}, while reducing the proportion of \texttt{Keep}, suggesting a directional bias in decision preference rather than a simple increase in output variability. Extending this observation, Pass@16 approximates the broader set of reachable decisions under multi-sampling and similarly exhibits a shift toward conservative actions. This indicates that the effect is not limited to stochastic single-generation behavior, but instead reflects a systematic reorganization of the model’s underlying action distribution. Overall, these results suggest that CF reasoning induces a structurally embedded and persistent change in the model’s decision space, which becomes more pronounced in larger models, implying stronger internalization of the induced reasoning structure.

\paragraph{Does CF Reasoning Intervene in Decision Generation?} 

\begin{figure}[h]
    \centering
    \includegraphics[width=\textwidth]{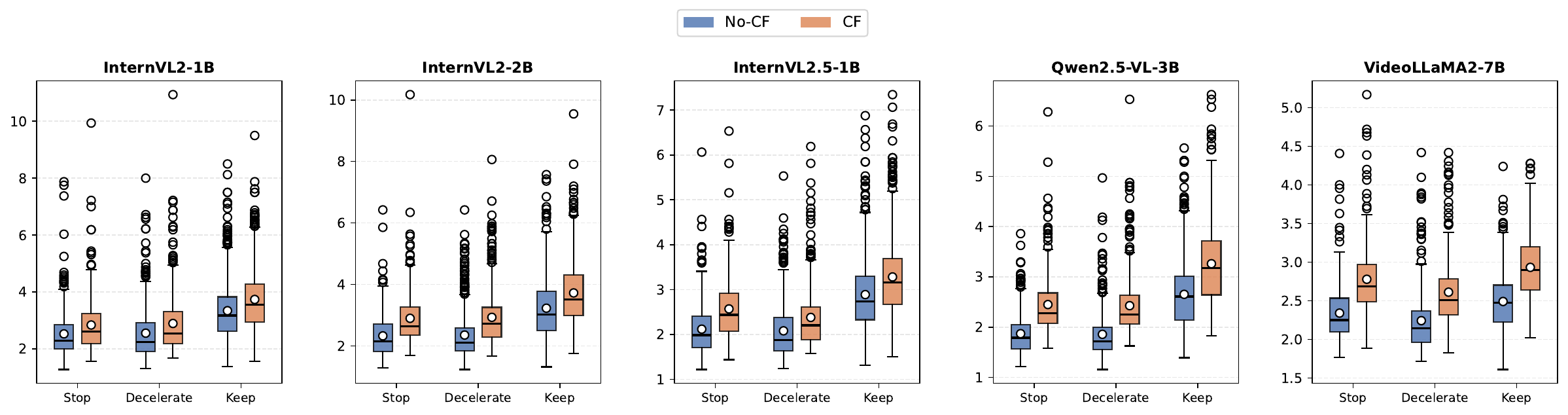}
    \caption{Class-wise perplexity comparison between the models w/o \textbf{(blue)} and w/ CF \textbf{(orange)} reasoning on the shared output $Y^{(-r)}_{\text{GT}}$, corresponding to $\text{PPL}_{\text{base}}(Y^{(-r)}_{\text{GT}})$ and $\text{PPL}_{\text{CF}}(Y^{(-r)}_{\text{GT}})$, respectively.}
    \label{fig:ppl_sft}
\end{figure}

However, it remains unclear whether the observed distribution shift is a surface-level artifact caused by the co-occurrence of CF reasoning tokens or whether it reflects a structural intervention of CF reasoning in the decision generation process that reshapes the model’s action distribution. To investigate this, we evaluate both models on a shared output, namely the common text $Y^{(-r)}_{\text{GT}}$, obtained by excluding the CF reasoning component, and compare their perplexity. If CF reasoning merely serves as an auxiliary output unrelated to decision generation, both models should assign similar likelihoods to $Y^{(-r)}_{\text{GT}}$. In contrast, if CF reasoning functions as a conditioning variable that actively influences decision generation, the CF model is expected to assign lower likelihood due to the removal of a key conditioning component, resulting in higher perplexity. Accordingly, the gap between $\text{PPL}_{\text{base}}(Y^{(-r)}_{\text{GT}})$ and $\text{PPL}_{\text{CF}}(Y^{(-r)}_{\text{GT}})$ serves as a diagnostic signal for structural intervention. As shown in Figure \ref{fig:ppl_sft}, the CF model consistently exhibits higher perplexity across all action classes, suggesting that CF reasoning acts as a conditioning factor that reshapes the decision-generation distribution rather than serving as a purely auxiliary output.

\subsection{Persistence of Behavioral Shifts under Reinforcement Fine-Tuning}
\label{subsec:rl_analysis}

However, it remains unclear whether these changes are merely artifacts or reflect a structural bias that persists throughout training. If CF reasoning serves only as a superficial signal for correct answers, its effect should diminish under reinforcement learning as reward optimization proceeds. In contrast, if it fundamentally reshapes the policy space, the induced bias should persist after reinforcement learning. To investigate this, we conduct RFT using the SFT models as reference models. We evaluate InternVL2.5-1B, which exhibits the smallest bias based on the combined change in \texttt{stop} and \texttt{decel.} under Pass@16, and Qwen2.5-VL-3B, which shows the largest bias. Details of the reward functions are provided in Appendix \ref{appendix:reward_functions}.
\begin{table*}[h]
\centering
\small
\resizebox{\textwidth}{!}{
\begin{tabular}{l c | cc cc | ccc}
\toprule
\multirow{2}{*}{\textbf{Model}} & \multirow{2}{*}{\textbf{Metric}}
  & \multicolumn{2}{c}{\textbf{Accident Recog.}}
  & \multicolumn{2}{c|}{\textbf{Ego Inclusion}}
  & \multicolumn{3}{c}{\textbf{Decision}} \\
 & & Yes & No & Involv. & Not In. & Stop & Decel. & Keep \\
\midrule

\multirow{6}{*}{\textbf{InternVL2.5-1B (GRPO)}}
 & P@1 w/o & 36.6 & 79.9 & 36.0 & 80.4 & 16.9 & 16.7 & 87.2 \\
 & P@1 w/  & 39.6 & 73.5 & 39.6 & 73.5 & 24.6 & 17.9 & 80.3 \\
 & $\Delta$
   & \cellcolor{blue!15}+3.0  & \cellcolor{red!15}-6.4
   & \cellcolor{blue!15}+3.6  & \cellcolor{red!15}-6.9
   & \cellcolor{blue!15}+7.7  & \cellcolor{blue!15}+1.2
   & \cellcolor{red!15}-6.9 \\
\cmidrule{2-9}
 & P@16 w/o & 92.9 & 99.7 & 92.9 & 99.7 & 73.8 & 75.1 & 100.0 \\
 & P@16 w/  & 96.3 & 99.9 & 96.3 & 99.9 & 82.7 & 84.4 &  99.8 \\
 & $\Delta$
   & \cellcolor{blue!15}+3.4  & \cellcolor{blue!15}+0.2
   & \cellcolor{blue!15}+3.4  & \cellcolor{blue!15}+0.2
   & \cellcolor{blue!15}+8.9  & \cellcolor{blue!15}+9.3
   & \cellcolor{red!15}-0.2 \\
\midrule

\multirow{6}{*}{\textbf{Qwen2.5-VL-3B (GRPO)}}
 & P@1 w/o & 49.9 & 56.8 & 47.5 & 58.5 & 17.9 & 35.9 & 65.8 \\
 & P@1 w/  & 71.0 & 32.6 & 67.3 & 35.3 & 41.9 & 32.3 & 42.4 \\
 & $\Delta$
   & \cellcolor{blue!15}+21.1 & \cellcolor{red!15}-24.2
   & \cellcolor{blue!15}+19.8 & \cellcolor{red!15}-23.2
   & \cellcolor{blue!15}+24.0 & \cellcolor{red!15}-3.6
   & \cellcolor{red!15}-23.4 \\
\cmidrule{2-9}
 & P@16 w/o & 95.7 & 94.0 & 95.7 & 94.8 & 69.6 & 92.3 & 96.7 \\
 & P@16 w/  & 100.0 & 89.0 & 100.0 & 92.9 & 94.6 & 92.6 & 92.8 \\
 & $\Delta$
   & \cellcolor{blue!15}+4.3  & \cellcolor{red!15}-5.0
   & \cellcolor{blue!15}+4.3  & \cellcolor{red!15}-1.9
   & \cellcolor{blue!15}+25.0 & \cellcolor{blue!15}+0.3
   & \cellcolor{red!15}-3.9 \\
\midrule

\multirow{6}{*}{\textbf{Qwen2.5-VL-3B (GDPO)}}
 & P@1 w/o & 45.4 & 62.5 & 42.7 & 64.9 & 13.4 & 29.0 & 70.5 \\
 & P@1 w/  & 74.7 & 38.2 & 68.8 & 41.7 & 44.4 & 68.8 & 44.4 \\
 & $\Delta$
   & \cellcolor{blue!15}+29.3 & \cellcolor{red!15}-24.3
   & \cellcolor{blue!15}+26.1 & \cellcolor{red!15}-23.2
   & \cellcolor{blue!15}+31.0 & \cellcolor{blue!15}+39.8
   & \cellcolor{red!15}-26.1 \\
\cmidrule{2-9}
 & P@16 w/o & 95.2 & 97.5 & 95.2 & 98.2 & 68.7 & 91.0 & 98.9 \\
 & P@16 w/  & 100.0 & 87.5 & 100.0 & 94.3 & 93.9 & 93.6 & 89.4 \\
 & $\Delta$
   & \cellcolor{blue!15}+4.8  & \cellcolor{red!15}-10.0
   & \cellcolor{blue!15}+4.8  & \cellcolor{red!15}-3.9
   & \cellcolor{blue!15}+25.2 & \cellcolor{blue!15}+2.6
   & \cellcolor{red!15}-9.5 \\
\bottomrule
\end{tabular}
}
\caption{
  Pass@1 and Pass@16 comparison w/ and w/o counterfactual reasoning after RFT (GRPO / GDPO).
  \textbf{Conservative} decisions (Stop / Decel.) are grouped under a shared header.
  In the $\Delta$ rows, \colorbox{blue!15}{\textbf{bold blue}} highlights gain
  and \colorbox{red!15}{\textbf{bold red}} highlights reduction.
  The consistent shift toward conservative decision-making across both GRPO and GDPO
  confirms that the observed phenomenon is not a training artifact.
}
\label{tab:passk_rft}
\end{table*}

\paragraph{Persistence of CF-Induced Behavioral Shift under Reinforcement Learning.} To examine whether the distribution shift could be mitigated through reinforcement learning, we conducted GRPO on InternVL2.5-1B and Qwen2.5-VL-3B, taking into account both algorithmic and model generality. Subsequently, we conducted GDPO on Qwen2.5-VL-3B, which exhibited the strongest bias during SFT, to further validate algorithmic generality. As shown in Table~\ref{tab:passk_rft}, the decision distribution changes induced by CF reasoning were maintained after RFT. If the patterns observed during SFT were mere training artifacts, they should have been corrected during reinforcement learning through reward optimization; however, the results indicate otherwise. In other words, the policy bias formed by CF reasoning persists after RFT. See Appendix \ref{appendix:passk_sft} and \ref{appendix:passk_rft} for detail analysis.

\begin{figure*}[h]
\centering
\begin{minipage}{0.48\textwidth}
Figure \ref{fig:ppl_rft} shows class-wise perplexity after RFT. Across all action classes, $\text{PPL}_{\text{CF}}(Y^{(-r)}_{\text{GT}})$ remains consistently higher than $\text{PPL}_{\text{base}}(Y^{(-r)}_{\text{GT}})$, following the same trend observed in SFT. Taken together, both pass@k and perplexity results indicate that the distributional effects induced by CF reasoning persist after reinforcement learning, suggesting a persistent structural influence on the model’s generative distribution rather than a superficial behavioral change.
\end{minipage}
\hfill
\begin{minipage}{0.48\textwidth}
    \centering
    \includegraphics[width=\linewidth]{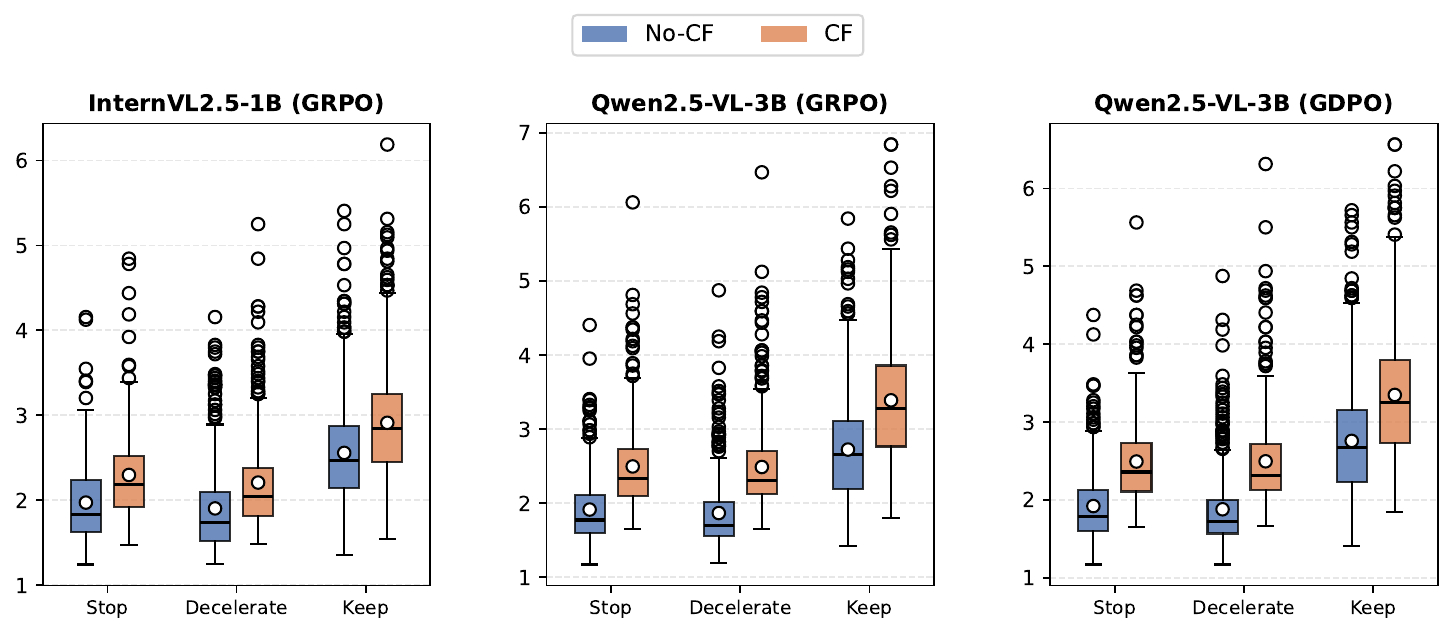}
    \caption{Class-wise perplexity comparison between the models w/o \textbf{(blue)} and w/ CF \textbf{(orange)} reasoning  on the shared output $Y^{(-r)}_{\text{GT}}$ after RFT.}
    \label{fig:ppl_rft}
\end{minipage}
\end{figure*}

\paragraph{Policy Entropy as a Dynamic Probe of Exploration.}
Pass@k and perplexity are metrics that evaluate the resulting action distribution after training convergence; however, they do not directly capture how the policy is formed during learning or how the exploration space evolves over time. To address this limitation, we log and analyze policy entropy throughout the entire RFT process. Since RFT incrementally updates the policy under a KL-divergence constraint with respect to the SFT reference model, it provides a suitable setting for dynamically tracking changes in the action distribution over the full training trajectory. Empirically, models incorporating CF reasoning consistently exhibit higher policy entropy throughout RFT. This indicates that CF reasoning alters the exploration structure of the policy during training. In particular, when considered alongside the shift toward conservative actions such as \texttt{Stop} and \texttt{Decelerate} observed in Pass@k, this increase in entropy does not reflect a mere rise in randomness. Instead, it can be interpreted as a redistribution of probability mass from a previously over-concentrated preference on \texttt{Keep} toward a more balanced allocation over multiple valid actions. Overall, the policy entropy analysis provides complementary evidence that CF reasoning affects not only the relative preference among actions but also the exploration structure of the policy during training, thereby reinforcing the findings from Pass@k and perplexity regarding its structural impact.

\begin{figure*}[h]
    \centering

    \makebox[\textwidth][c]{%
        \begin{subfigure}{0.32\textwidth}
            \centering
            \includegraphics[width=\linewidth]{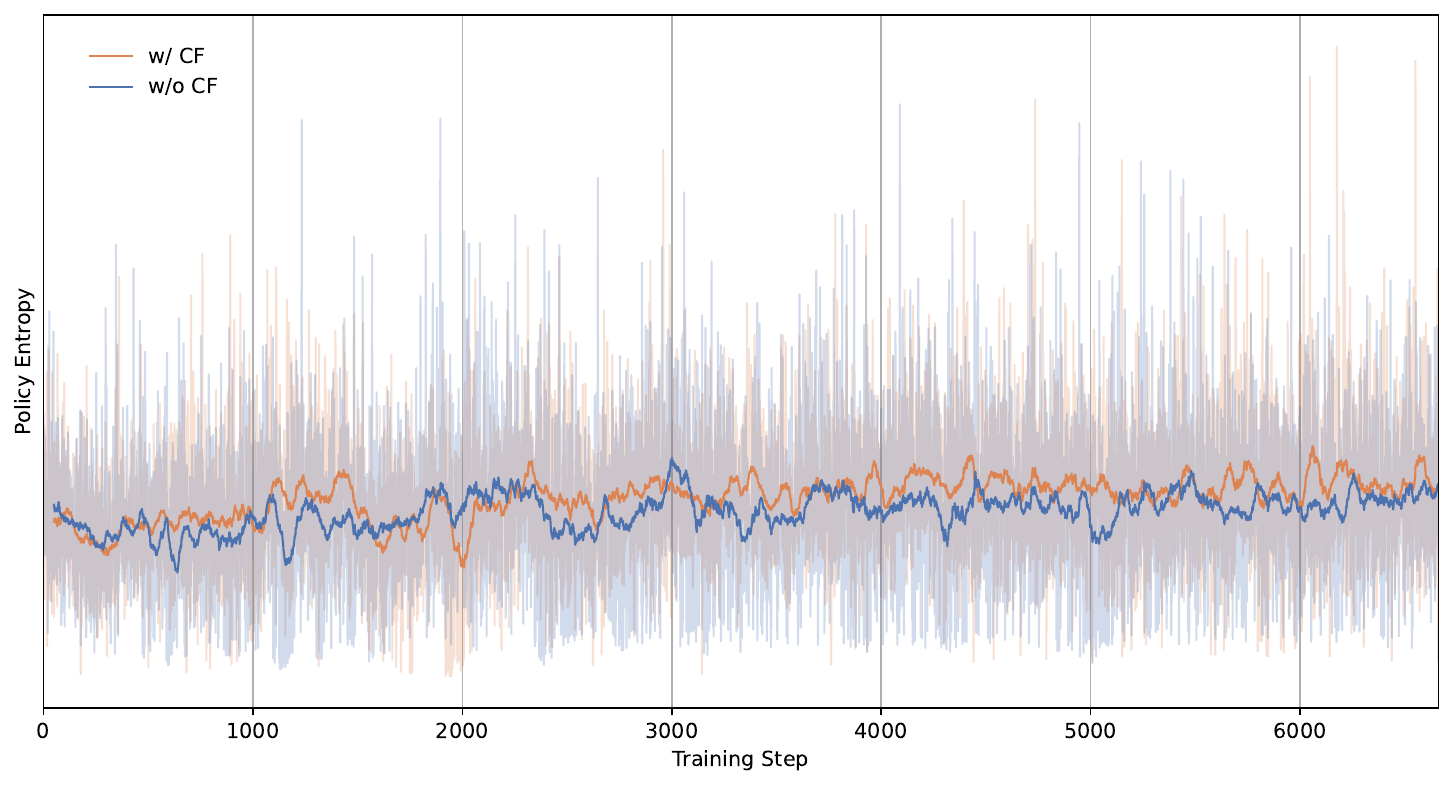}
            \caption{InternVL2.5-1B (GRPO)}
        \end{subfigure}
        \begin{subfigure}{0.32\textwidth}
            \centering
            \includegraphics[width=\linewidth]{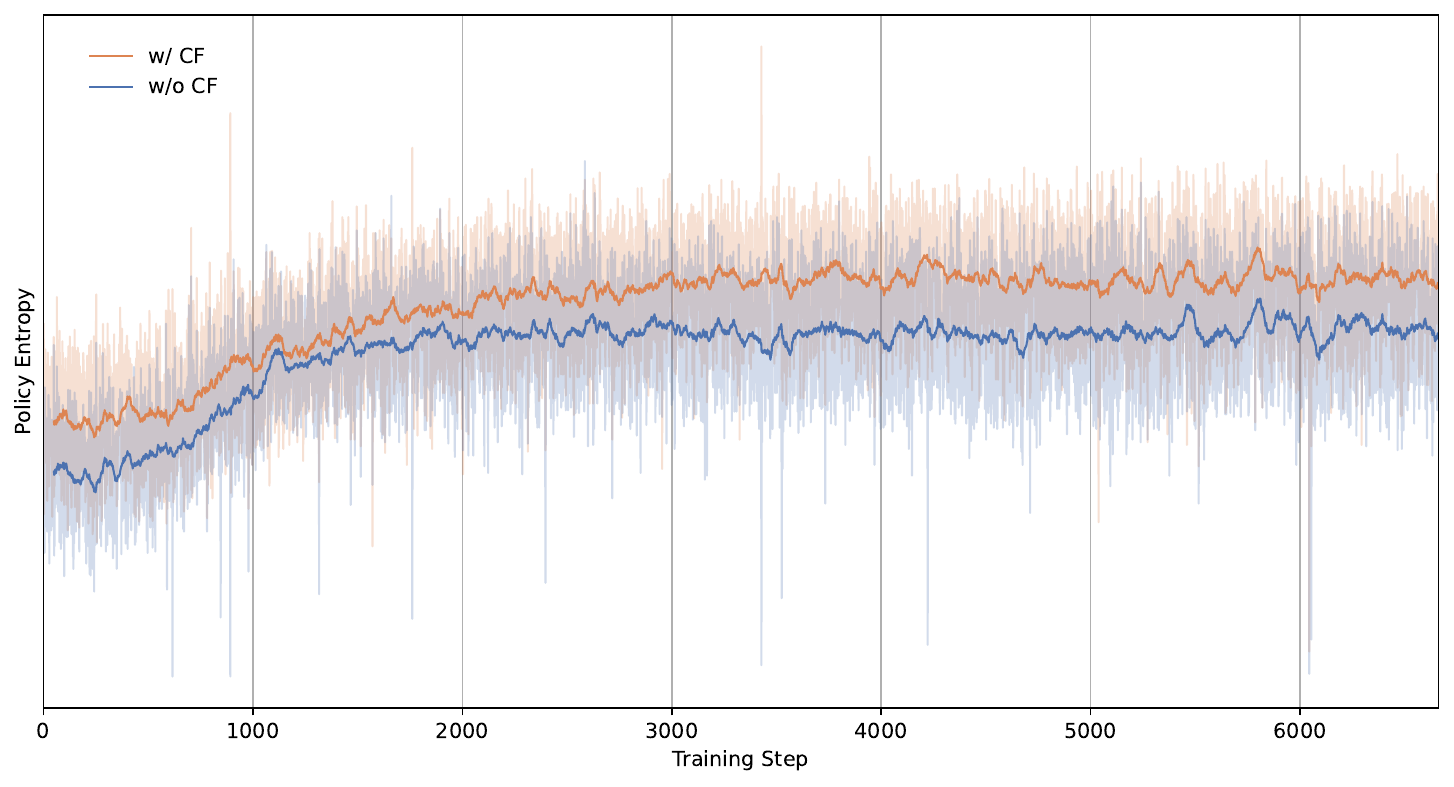}
            \caption{Qwen2.5-VL-3B (GRPO)}
        \end{subfigure}
        \hspace{1em}
        \begin{subfigure}{0.32\textwidth}
            \centering
            \includegraphics[width=\linewidth]{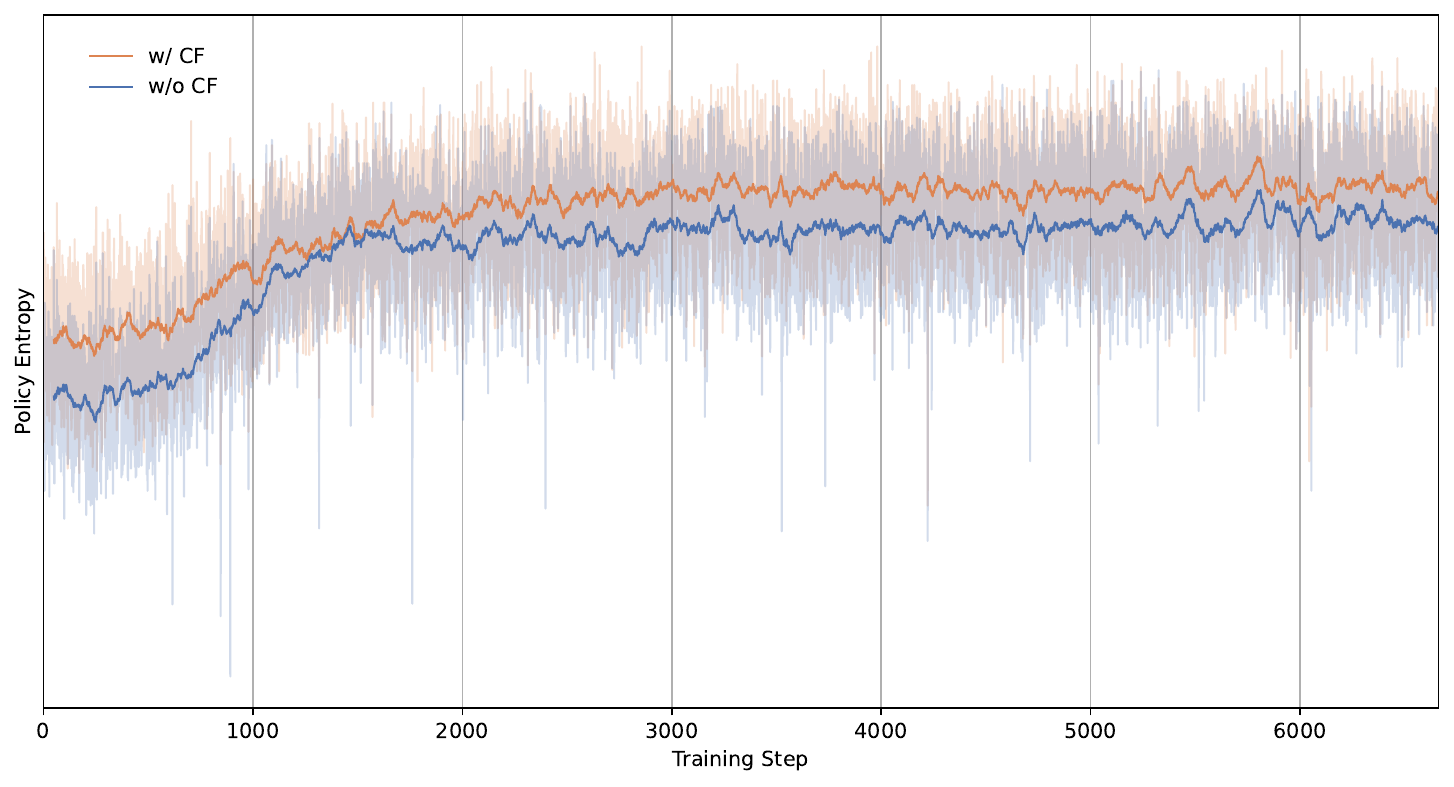}
            \caption{Qwen2.5-VL-3B (GDPO)}
        \end{subfigure}
    }

    \caption{Policy entropy over training steps during RFT.}
    \label{fig:entropy}
\end{figure*}

\subsection{CF Reasoning Decouples from Visual Grounding}
\label{subsec:grounding_analysis}

In summary, we indirectly examined the impact of CF reasoning on decision generation by focusing on the model’s generated text outputs and their probabilistic characteristics. However, the metrics used in the previous analysis are suitable for studying latent behavioral reasoning, but they do not directly evaluate alignment with causal structure. Therefore, we conduct a more direct investigation to determine whether the observed effects arise from genuine causal reasoning induced by CF reasoning or from linguistic bias in this section.

\paragraph{Token Activation Map Analysis.} If CF reasoning couples with perceptual awareness, the model’s decisions should rely more strongly on the visual causes of accidents, with attention correspondingly focusing on relevant risk-related regions. To examine whether CF reasoning induces decision-making bias through genuine visual risk perception, we conduct a visual attribute analysis using Token Activation Maps (TAM) \cite{attention:tam}, which isolate token-level activations to estimate the visual regions on which each output token depends. Unlike methods affected by inter-token interactions, TAM provides a more stable measure of the visual cues used by specific reasoning tokens. Based on this, we compute three metrics: (i) token-level IoU ($IoU_{tok}$), which measures how precisely individual causal keyword tokens align with accident-causing objects; (ii) aggregated IoU ($IoU_{agg}$), which evaluates whether attention remains consistently focused on the target object over the entire reasoning process; and (iii) grounding F1 ($F1_{gnd}$), which jointly accounts for precision and recall to balance the completeness and accuracy of attention localization described in Appendix ~\ref{appendix:implementations_iou_metrics}. As shown in Table~\ref{tab:iou}, IoU values vary across models but remain largely unaffected by prompt variations, indicating consistent spatial grounding patterns. However, no clear alignment between attention and accident-causing objects is observed, suggesting that the CF-induced action distribution shifts may occur independently of visual evidence as shown in Figure \ref{fig:attention_map}.

\begin{table*}[h] 
\centering
\resizebox{\textwidth}{!}{%
\begin{tabular}{@{}lcccccccccccc@{}}
\toprule
\multirow{3}{*}{} & \multicolumn{3}{c}{\textbf{Ablation}} & \multicolumn{3}{c}{\textbf{perception}} & \multicolumn{3}{c}{\textbf{decision}} & \multicolumn{3}{c}{\textbf{counterfactual}} \\
\cmidrule(lr){2-4} \cmidrule(lr){5-7} \cmidrule(lr){8-10} \cmidrule(lr){11-13}
 & \textbf{Type} & \textbf{CF} & \textbf{RFT} & \textbf{$IoU_{tok}$} & \textbf{$IoU_{agg}$} & \textbf{$F1_{gnd}$} & \textbf{$IoU_{tok}$} & \textbf{$IoU_{agg}$} & \textbf{$F1_{gnd}$} & \textbf{$IoU_{tok}$} & \textbf{$IoU_{agg}$} & \textbf{$F1_{gnd}$} \\
\midrule
\multirow{4}{*}{InternVL2.5-1B} 
 & SFT & \xmark & ---  & 9.56 & 10.73 & 18.67 & 7.81 & 8.60 & 14.91 & 7.93 & 8.83 & 15.14 \\
 & SFT & \cmark & ---  & 9.54 & 10.91 & 19.04 & 8.62 & 9.67 & 16.54 & 8.73 & 9.71 & 16.72 \\
 & RFT & \xmark & GRPO & 9.39 & 10.85 & 18.85 & 8.06 & 8.91 & 15.21 & 8.46 & 9.34 & 15.91 \\
 & RFT & \cmark & GRPO & 10.39 & 11.46 & 19.74 & 9.22 & 9.79 & 16.66 & 8.70 & 9.39 & 16.00 \\
\midrule
\multirow{6}{*}{Qwen2.5-VL-3B} 
 & SFT & \xmark & ---  & 16.42 & 20.04 & 32.50 & 17.38 & 19.75 & 32.05 & 16.12 & 18.33 & 29.96 \\
 & SFT & \cmark & ---  & 15.97 & 19.73 & 31.92 & 17.46 & 19.86 & 32.14 & 16.64 & 18.84 & 30.63 \\
 & RFT & \xmark & GRPO & 16.21 & 18.84 & 30.81 & 16.81 & 18.66 & 30.48 & 15.27 & 17.15 & 28.29 \\
 & RFT & \cmark & GRPO & 16.57 & 19.63 & 31.79 & 17.81 & 19.74 & 31.89 & 17.30 & 19.24 & 31.10 \\
 & RFT & \xmark & GDPO & 16.68 & 19.23 & 31.39 & 17.43 & 19.38 & 31.51 & 15.75 & 17.56 & 28.84 \\
 & RFT & \cmark & GDPO & 16.78 & 19.72 & 31.89 & 18.18 & 20.26 & 32.61 & 18.04 & 20.16 & 32.41 \\
\bottomrule
\end{tabular}%
}
\caption{Quantitative evaluation of visual grounding across different prompt types.}
\label{tab:iou}
\end{table*}

\paragraph{Causal Intervention Analysis.}
Previous results suggest that the action distribution shift induced by CF reasoning may be independent of visual evidence. To test this, we conduct a causal intervention study on pre-collision frames, where we remove different visual elements and measure changes in model decisions. Specifically, we consider three masking settings: causal masking (removing the accident-causing object), non-causal masking (removing irrelevant objects), and random masking (removing background regions), and compute the decision change rate by comparing predictions with the original video. As shown in Table~\ref{tab:causal_masking}, human evaluation exhibits the highest sensitivity under causal masking, while non-causal and random masking yield much lower changes, indicating strong reliance on true causal factors. In contrast, most VLMs show no such selectivity, with similar change rates across all masking conditions and this behavior persists regardless of CF reasoning or reinforcement learning. Overall, these results suggest that the action distribution shifts are unlikely to reflect genuine causal reasoning, but are more consistent with visual-agnostic linguistic biases.

\begin{table*}[h]
\centering
\footnotesize
\resizebox{\textwidth}{!}{%
\begin{tabular}{@{}l cc cc cccc cccccc cc@{}}
\toprule
\multirow{3}{*}{}
  & \multicolumn{2}{c}{\textbf{InternVL2-1B}}
  & \multicolumn{2}{c}{\textbf{InternVL2-2B}}
  & \multicolumn{4}{c}{\textbf{InternVL2.5-1B}}
  & \multicolumn{6}{c}{\textbf{Qwen2.5-VL-3B}}
  & \multicolumn{2}{c}{\textbf{VideoLLaMA2-7B}} \\
\cmidrule(lr){2-3}
\cmidrule(lr){4-5}
\cmidrule(lr){6-9}
\cmidrule(lr){10-15}
\cmidrule(lr){16-17}
  & \multicolumn{2}{c}{SFT}
  & \multicolumn{2}{c}{SFT}
  & \multicolumn{2}{c}{SFT}
  & \multicolumn{2}{c}{RFT (GRPO)}
  & \multicolumn{2}{c}{SFT}
  & \multicolumn{2}{c}{RFT (GRPO)}
  & \multicolumn{2}{c}{RFT (GDPO)}
  & \multicolumn{2}{c}{SFT} \\
\cmidrule(lr){2-3}
\cmidrule(lr){4-5}
\cmidrule(lr){6-7}
\cmidrule(lr){8-9}
\cmidrule(lr){10-11}
\cmidrule(lr){12-13}
\cmidrule(lr){14-15}
\cmidrule(lr){16-17}
  & \textbf{\xmark} & \textbf{\cmark}
  & \textbf{\xmark} & \textbf{\cmark}
  & \textbf{\xmark} & \textbf{\cmark}
  & \textbf{\xmark} & \textbf{\cmark}
  & \textbf{\xmark} & \textbf{\cmark}
  & \textbf{\xmark} & \textbf{\cmark}
  & \textbf{\xmark} & \textbf{\cmark}
  & \textbf{\xmark} & \textbf{\cmark} \\
\midrule
Causal
  & 0.455 & 0.494
  & 0.408 & 0.451
  & 0.425 & 0.382
  & 0.307 & 0.340
  & 0.457 & 0.524
  & 0.404 & 0.520
  & 0.417 & 0.484
  & 0.434 & 0.508 \\
Non-causal
  & 0.463 & 0.498
  & 0.367 & 0.404
  & 0.416 & 0.373
  & 0.310 & 0.346
  & 0.457 & 0.546
  & 0.416 & 0.504
  & 0.399 & 0.548
  & 0.396 & 0.519 \\
Random
  & 0.475 & 0.508
  & 0.385 & 0.416
  & 0.452 & 0.363
  & 0.295 & 0.355
  & 0.458 & 0.519
  & 0.411 & 0.524
  & 0.388 & 0.531
  & 0.414 & 0.513 \\
\bottomrule
\end{tabular}%
}
\caption{Comparison of causal masking effects across different models and training settings.}
\label{tab:causal_masking}
\end{table*}

\section{Extended Analysis}
\label{sec:more_analysis}
\paragraph{Generalization to Trajectory-Level Behavior.} To examine whether the conservative shift persists beyond crash scenarios and under fine-grained trajectory-level evaluation, we conducted an additional study using the existing counterfactual dataset \cite{causal:e2e-vlm:omnidrive}. We compared models trained with and without CF reasoning. For each sample $i$, we compute the maximum lateral displacement of the predicted trajectory, $d_i=\max_t |\ell_{i,t}|$, where $\ell_{i,t}$ is the ego-frame lateral coordinate at time $t$. A trajectory is counted as a lateral maneuver if $d_i \ge \tau$. The lateral maneuver rate over an evaluation subset $\mathcal{S}$ is
\begin{equation}
\rho_\tau(\mathcal{S}) = \frac{1}{|\mathcal{S}|}\sum_{i\in\mathcal{S}} \mathbb{1}\!\left[d_i \ge \tau\right],
\end{equation}
and we report the gap $\Delta_\rho=\rho_\tau^{\text{CF}}-\rho_\tau^{\text{No-CF}}$ for $\tau\in\{0.3, 0.5, 0.8\}$\,m. We evaluate on all samples ($\mathcal{S}_{\text{all}}$) and on the subsets whose ground-truth trajectories contain a lateral maneuver, pooled ($\mathcal{S}_{\text{GT}}$) and split into left-turn ($\mathcal{S}_{\text{L}}$) and right-turn ($\mathcal{S}_{\text{R}}$). As shown in Table~\ref{tab:discussion:omnidrive}, the CF model consistently exhibits lower maneuver rates ($\Delta_\rho<0$) at every threshold, both on $\mathcal{S}_{\text{all}}$ and on $\mathcal{S}_{\text{GT}}$. The same trend holds for both $\mathcal{S}_{\text{L}}$ and $\mathcal{S}_{\text{R}}$, indicating that the conservative shift is not merely attributable to the prevalence of straight-driving scenarios. These results show that the conservative behavioral bias induced by CF reasoning is also observed in general driving scenarios and under low-level trajectory evaluation. Moreover, performance improvements on nuScenes do not necessarily reflect improved reasoning: since straight trajectories make up most of the ground truth in the dataset~\cite{causal:e2e:ad-mlp,ad:unitraj}, conservative predictions can achieve low average error, which is consistent with the model trained with CF reasoning.

\begin{table*}[h]
\centering
\setlength{\tabcolsep}{5pt}
\begin{tabular}{@{}clrrrr@{}}
\toprule
$\tau$ (m) & Subset & $N$ & $\rho_\tau^{\text{No-CF}}$ & $\rho_\tau^{\text{CF}}$ & $\Delta_\rho$  \\
\midrule
\multirow{4}{*}{0.3m}
& $\mathcal{S}_{\text{all}}$& 5,119 & 8.85\%  & 3.07\% & $-5.78$ pp  \\
& $\mathcal{S}_{\text{GT}}$ & 2,286 & 17.94\% & 5.95\% & $-11.99$ pp \\
& $\mathcal{S}_{\text{L}}$  & 874   & 29.52\% & 11.21\% & $-18.31$ pp \\
& $\mathcal{S}_{\text{R}}$  & 1,412 & 10.76\% & 2.69\% & $-8.07$ pp \\
\midrule
\multirow{4}{*}{0.5m}
& $\mathcal{S}_{\text{all}}$& 5,119 & 6.58\%  & 1.93\% & $-4.65$ pp \\
& $\mathcal{S}_{\text{GT}}$ & 1,676 & 18.85\% & 5.55\% & $-13.31$ \\
& $\mathcal{S}_{\text{L}}$  & 697   & 32.14\% & 11.19\% & $-20.95$ pp \\
& $\mathcal{S}_{\text{R}}$  & 979   & 9.40\%  & 1.53\% & $-7.87$ pp \\
\midrule
\multirow{4}{*}{0.8m}
& $\mathcal{S}_{\text{all}}$& 5,119 & 3.48\%  & 0.84\% & $-2.64$ pp \\
& $\mathcal{S}_{\text{GT}}$ & 1,244 & 13.34\% & 3.05\% & $-10.29$ pp \\
& $\mathcal{S}_{\text{L}}$  & 512   & 25.39\% & 6.64\% & $-18.75$ pp \\
& $\mathcal{S}_{\text{R}}$  & 732   & 4.92\%  & 0.55\% & $-4.37$ pp \\
\bottomrule
\end{tabular}
\caption{Lateral maneuver rate $\rho_\tau$ of No-CF and CF models on OmniDrive, where a trajectory counts as a maneuver if its maximum lateral displacement is at least $\tau$. $\Delta_\rho=\rho_\tau^{\text{CF}}-\rho_\tau^{\text{No-CF}}$. $\mathcal{S}_{\text{all}}$: all samples; $\mathcal{S}_{\text{GT}}$: samples whose ground-truth trajectory contains a lateral maneuver; $\mathcal{S}_{\text{L}}$, $\mathcal{S}_{\text{R}}$: its left- and right-turn subsets ($\mathcal{S}_{\text{GT}}=\mathcal{S}_{\text{L}}\cup\mathcal{S}_{\text{R}}$).}
\label{tab:discussion:omnidrive}
\end{table*}

\paragraph{Investigating Semantic Priors as a Cause of the Action Shift.} Furthermore, to investigate whether the conservative shift arises from hazard-specific linguistic priors rather than visual and causal grounding, we conducted a conflict test. We injected either adversarial hazard-related text or safety-neutral text into normal driving videos without accidents and measured the false positive rate (FPR) of \texttt{Stop} and \texttt{Decel.} predictions. We define *hazard specificity* as the difference in FPR between hazard-related and safety-neutral text injections, where a positive value indicates a stronger response to hazard-related text. As shown in Table~\ref{tab:discussion:semantic}, models with CF reasoning already predict \texttt{Stop} and \texttt{Decel.} in 61--69\% of cases, compared with 12--25\% for models without CF reasoning, confirming their conservative bias. Under text injection, No-CF models exhibit near-zero or weakly positive hazard specificity ($+0.71$ to $+5.00$\,pp), whereas CF models exhibit negative hazard specificity across all three training settings ($-6.50$ to $-14.21$\,pp), indicating that they respond more strongly to safety-neutral text than to hazard-related text. This pattern runs counter to what would be expected if hazard-specific semantic priors were driving the conservative shift. Nevertheless, the elevated FPR of CF models under both types of injection suggests that their predictions are sensitive to injected text regardless of its safety-related content. Taken together, these results suggest that hazard-related semantic priors alone cannot account for the observed conservative shift.

\begin{table*}[h]
\centering
\footnotesize
\resizebox{\textwidth}{!}{%
\begin{tabular}{@{}l cc cc cc@{}}
\toprule
\multirow{2}{*}{\textbf{Metric}}
  & \multicolumn{2}{c}{\textbf{Qwen2.5-VL-3B (SFT)}}
  & \multicolumn{2}{c}{\textbf{Qwen2.5-VL-3B (GRPO)}}
  & \multicolumn{2}{c}{\textbf{Qwen2.5-VL-3B (GDPO)}} \\
\cmidrule(lr){2-3}
\cmidrule(lr){4-5}
\cmidrule(lr){6-7}
  & \textbf{No-CF} & \textbf{CF}
  & \textbf{No-CF} & \textbf{CF}
  & \textbf{No-CF} & \textbf{CF} \\
\midrule
Baseline FPR
  & 25.00\% & 69.00\%
  & 21.58\% & 66.96\%
  & 12.35\% & 61.01\% \\

Adversarial FPR
  & 33.25\% & 72.25\%
  & 31.21\% & 63.24\%
  & 22.92\% & 60.20\% \\

Neutral FPR
  & 28.25\% & 78.75\%
  & 29.65\% & 77.46\%
  & 22.21\% & 70.32\% \\

Hazard Specificity (Adv. - Neu.)
  & +5.00 pp & $-$6.50 pp
  & +1.56 pp & $-$14.21 pp
  & +0.71 pp & $-$10.12 pp \\
\bottomrule
\end{tabular}%
}
\caption{False positive rates and hazard specificity under baseline,
adversarial, and neutral conditions across different model architectures
and training methods. No-CF and CF denote training without and with
CF reasoning, respectively. Hazard specificity is defined as
$\mathrm{FPR}_{\mathrm{Adv}}-\mathrm{FPR}_{\mathrm{Neu}}$.}
\label{tab:discussion:semantic}
\end{table*}

\section{Related Work}

\paragraph{Shortcut Learning and Causal Confusion.}
Ensuring robust performance under distribution shift is a central challenge in autonomous driving. Thus, causal reasoning is regarded as a particularly promising solution due to its invariance to environmental changes compared to spurious correlations. However, existing works based on imitation learning, which learn by mimicking expert demonstrations, are inherently vulnerable to shortcut learning \cite{causal:e2e:causal_confusion, causal:e2e:driveadapter, causal:e2e:exploring_causality_in_e2e}. This phenomenon, also referred to as causal confusion, arises when models rely on statistical correlations in the data rather than underlying causal structure. As a result, models tend to depend on superficial patterns instead of causal reasoning. A representative example is the observation that competitive driving performance can be achieved even using only ego-status and navigation information \cite{causal:e2e:ad-mlp, e2e-ad:is-ego-status}. To mitigate such issues, several studies have proposed explicitly disentangling causal factors \cite{causal:e2e:cadet, causal:e2e:causalvad, causal:e2e:decoupling_perception_and_ego_status}. However, these approaches still struggle in complex scene understanding and contextual reasoning scenarios, indicating that shortcut learning remains an open challenge in end-to-end autonomous driving.

\paragraph{Illusion of Causal Reasoning in VLM for Autonomous Driving.}

To address limitations of conventional end-to-end approaches, recent studies have explored VLMs for autonomous driving, leveraging language as a coarse semantic representation to better capture complex scene contexts and achieving improved performance in closed-loop settings \cite{causal:e2e-vlm:simlingo, e2e-vlm:orion, e2e-vlm:autovla}. Building on this, prior work has introduced intermediate linguistic reasoning steps such as chain-of-thought \cite{e2e-vlm:drivelm}, chain-of-causation \cite{causal:e2e-vlm:coc-vla}, and counterfactual reasoning \cite{causal:cf_vla}, leading to consistent performance gains and an implicit assumption that such reasoning reflects genuine causal understanding. However, it remains unclear whether these improvements stem from causal reasoning or from shortcut learning induced by the language modality, as standard driving metrics are confounded and insufficient for analyzing internal reasoning processes \cite{causal:e2e-vlm:more_than_meets_the_eye}. Recent evidence from CausalDriveBench \cite{causal:e2e-vlm:causaldrivebench} further highlights this limitation, showing that causal QA accuracy is uncorrelated with trajectory prediction accuracy across driving VLAs, indicating that conventional driving metrics alone cannot assess whether a model's linguistic reasoning reflects genuine causal understanding. To address this, we introduce a framework to evaluate the behavioral effects of CF reasoning by quantifying the induced shifts in a discrete meta-action space and validating whether these shifts persist under reinforcement learning, thereby ruling out training artifacts. This enables a systematic analysis of the gap between linguistic shortcut and genuine causal understanding.

\section{Conclusion and Limitations}
\label{sec:conclusion}
In this study, we investigate the effect of counterfactual (CF) reasoning in VLMs for autonomous driving. Our results indicate that CF reasoning consistently induces an action distribution shift, and that these changes primarily originate from language-level reasoning rather than genuine causal reasoning grounded in visual observations. This finding suggests that CF reasoning can guide the model in determining what to say, but does not necessarily teach the model how to reason. In other words, the behavior shift is more plausibly attributed to the learning of a language-induced spurious correlation, rather than the internalization of genuine causal reasoning.

However, this study focuses on open-source VLMs that allow access to internal probability distributions and learning dynamics. Consequently, the generalizability of our findings to closed-source or specialized VLMs \cite{causal:e2e-vlm:alpamayo, causal:cf_vla} remains uncertain, as our analyses based on perplexity and policy entropy require access to internal model probabilities. Therefore, our conclusions are limited to settings where such internal signals are observable, highlighting the need for future work to extend CF reasoning analysis beyond these constraints. We further discuss potential causes of the observed action distribution shift and outline future directions in Section \ref{appendix:discussion}.

\section{Acknowledgement}
This work was supported by the National Research Foundation of Korea (NRF) grant funded by the Korea government (MSIT) (RS-2026-25510069), the Institute of Information and Communications Technology Planning and Evaluation (IITP) grant funded by the Korea government (No. 2022-0-00469), and, in part, by the BK21 FOUR (Connected AI Education \& Research Program for Industry and Society Innovation, KAIST EE, No. 4120200113769).

\clearpage
\bibliographystyle{unsrt}
\bibliography{ref}

@book{causal:ladder_of_causation,
  title={The book of why: the new science of cause and effect},
  author={Pearl, Judea and Mackenzie, Dana},
  year={2018},
  publisher={Basic books}
}

@article{causal:cf_vla,
  title={Counterfactual VLA: Self-Reflective Vision-Language-Action Model with Adaptive Reasoning},
  author={Peng, Zhenghao and Ding, Wenhao and You, Yurong and Chen, Yuxiao and Luo, Wenjie and Tian, Thomas and Cao, Yulong and Sharma, Apoorva and Xu, Danfei and Ivanovic, Boris and others},
  journal={arXiv preprint arXiv:2512.24426},
  year={2025}
}

@article{policy-entropy:cui2025entropy,
  title={The entropy mechanism of reinforcement learning for reasoning language models},
  author={Cui, Ganqu and Zhang, Yuchen and Chen, Jiacheng and Yuan, Lifan and Wang, Zhi and Zuo, Yuxin and Li, Haozhan and Fan, Yuchen and Chen, Huayu and Chen, Weize and others},
  journal={arXiv preprint arXiv:2505.22617},
  year={2025}
}

@article{policy-entropy:liu2025prorl,
  title={Prorl: Prolonged reinforcement learning expands reasoning boundaries in large language models},
  author={Liu, Mingjie and Diao, Shizhe and Lu, Ximing and Hu, Jian and Dong, Xin and Choi, Yejin and Kautz, Jan and Dong, Yi},
  journal={arXiv preprint arXiv:2505.24864},
  year={2025}
}

@article{policy-entropy:reasoning_w_exploration, 
    title={Reasoning with Exploration: An Entropy Perspective}, volume={40},
    DOI={10.1609/aaai.v40i36.40290}, number={36}, 
    journal={Proceedings of the AAAI Conference on Artificial Intelligence}, 
    author={Cheng, Daixuan and Huang, Shaohan and Zhu, Xuekai and Dai, Bo and Zhao, Xin and Zhang, Zhenliang and Wei, Furu}, year={2026}, 
    pages={30377–30385} 
}

@article{limit-rlvr:rethinking-reflection,
  title={Rethinking reflection in pre-training},
  author={Shah, Darsh J and Rushton, Peter and Singla, Somanshu and Parmar, Mohit and Smith, Kurt and Vanjani, Yash and Vaswani, Ashish and Chaluvaraju, Adarsh and Hojel, Andrew and Ma, Andrew and others},
  journal={arXiv preprint arXiv:2504.04022},
  year={2025}
}

@article{limit-rlvr:echo_chamber,
  title={Echo chamber: Rl post-training amplifies behaviors learned in pretraining},
  author={Zhao, Rosie and Meterez, Alexandru and Kakade, Sham and Pehlevan, Cengiz and Jelassi, Samy and Malach, Eran},
  journal={arXiv preprint arXiv:2504.07912},
  year={2025}
}

@article{limit-rlvr:does,
  title={Does reinforcement learning really incentivize reasoning capacity in llms beyond the base model?},
  author={Yue, Yang and Chen, Zhiqi and Lu, Rui and Zhao, Andrew and Wang, Zhaokai and Song, Shiji and Huang, Gao},
  journal={arXiv preprint arXiv:2504.13837},
  year={2025}
}

@inproceedings{causal:whatif_tv_off?,
  title={What if the tv was off? examining counterfactual reasoning abilities of multi-modal language models},
  author={Zhang, Letian and Zhai, Xiaotong and Zhao, Zhongkai and Zong, Yongshuo and Wen, Xin and Zhao, Bingchen},
  booktitle={Proceedings of the IEEE/CVF Conference on Computer Vision and Pattern Recognition},
  pages={21853--21862},
  year={2024}
}

@article{causal:defacto,
  title={DeFacto: Counterfactual Thinking with Images for Enforcing Evidence-Grounded and Faithful Reasoning},
  author={Xu, Tianrun and Jing, Haoda and Li, Ye and Wei, Yuquan and Feng, Jun and Chen, Guanyu and Gao, Haichuan and Zhang, Tianren and Chen, Feng},
  journal={arXiv preprint arXiv:2509.20912},
  year={2025}
}

@inproceedings{causal:bridging_llms_and_causal_world_model,
title={Language Agents Meet Causality -- Bridging {LLM}s and Causal World Models},
author={John Gkountouras and Matthias Lindemann and Phillip Lippe and Efstratios Gavves and Ivan Titov},
booktitle={The Thirteenth International Conference on Learning Representations},
year={2025},
url={https://openreview.net/forum?id=y9A2TpaGsE}
}

@article{causal:llm_cannot_discover_causality,
  title={LLM cannot discover causality, and should be restricted to non-decisional support in causal discovery},
  author={Wu, Xingyu and Yu, Kui and Wu, Jibin and Tan, Kay Chen},
  journal={arXiv preprint arXiv:2506.00844},
  year={2025}
}

@article{e2e-vlm:drivemlm,
  title={Drivemlm: Aligning multi-modal large language models with behavioral planning states for autonomous driving},
  author={Wang, Wenhai and Xie, Jiangwei and Hu, ChuanYang and Zou, Haoming and Fan, Jianan and Tong, Wenwen and Wen, Yang and Wu, Silei and Deng, Hanming and Li, Zhiqi and others},
  journal={arXiv preprint arXiv:2312.09245},
  year={2023}
}

@inproceedings{e2e-vlm:rda-driver,
  title={Making large language models better planners with reasoning-decision alignment},
  author={Huang, Zhijian and Tang, Tao and Chen, Shaoxiang and Lin, Sihao and Jie, Zequn and Ma, Lin and Wang, Guangrun and Liang, Xiaodan},
  booktitle={European Conference on Computer Vision},
  pages={73--90},
  year={2024},
  organization={Springer}
}

@article{e2e-vlm:driving_with_internvl_challange,
  title={Driving with internvl: Oustanding champion in the track on driving with language of the autonomous grand challenge at cvpr 2024},
  author={Li, Jiahan and Li, Zhiqi and Lu, Tong},
  journal={arXiv preprint arXiv:2412.07247},
  year={2024}
}

@inproceedings{e2e-vlm:drivelm,
  title={Drivelm: Driving with graph visual question answering},
  author={Sima, Chonghao and Renz, Katrin and Chitta, Kashyap and Chen, Li and Zhang, Hanxue and Xie, Chengen and Bei{\ss}wenger, Jens and Luo, Ping and Geiger, Andreas and Li, Hongyang},
  booktitle={European conference on computer vision},
  pages={256--274},
  year={2024},
  organization={Springer}
}

@article{e2e-vlm:drivegpt4,
  title={Drivegpt4: Interpretable end-to-end autonomous driving via large language model},
  author={Xu, Zhenhua and Zhang, Yujia and Xie, Enze and Zhao, Zhen and Guo, Yong and Wong, Kwan-Yee K and Li, Zhenguo and Zhao, Hengshuang},
  journal={IEEE Robotics and Automation Letters},
  year={2024},
  publisher={IEEE}
}

@inproceedings{e2e-vlm:vlaad,
  title={Vlaad: Vision and language assistant for autonomous driving},
  author={Park, SungYeon and Lee, MinJae and Kang, JiHyuk and Choi, Hahyeon and Park, Yoonah and Cho, Juhwan and Lee, Adam and Kim, DongKyu},
  booktitle={Proceedings of the IEEE/CVF Winter Conference on Applications of Computer Vision},
  pages={980--987},
  year={2024}
}

@article{e2e-vlm:occllama,
  title={Occllama: An occupancy-language-action generative world model for autonomous driving},
  author={Wei, Julong and Yuan, Shanshuai and Li, Pengfei and Hu, Qingda and Gan, Zhongxue and Ding, Wenchao},
  journal={arXiv preprint arXiv:2409.03272},
  year={2024}
}

@inproceedings{e2e-ad:is-ego-status,
  title={Is ego status all you need for open-loop end-to-end autonomous driving?},
  author={Li, Zhiqi and Yu, Zhiding and Lan, Shiyi and Li, Jiahan and Kautz, Jan and Lu, Tong and Alvarez, Jose M},
  booktitle={Proceedings of the IEEE/CVF Conference on Computer Vision and Pattern Recognition},
  pages={14864--14873},
  year={2024}
}

@article{ad:unitraj,
  title={UniTraj: A Unified Framework for Scalable Vehicle Trajectory Prediction},
  author={Feng, Lan and Bahari, Mohammadhossein and Amor, Kaouther Messaoud Ben and Zablocki, {\'E}loi and Cord, Matthieu and Alahi, Alexandre},
  journal={arXiv preprint arXiv:2403.15098},
  year={2024}
}

@article{e2e-vlm:senna,
  title={Senna: Bridging large vision-language models and end-to-end autonomous driving},
  author={Jiang, Bo and Chen, Shaoyu and Liao, Bencheng and Zhang, Xingyu and Yin, Wei and Zhang, Qian and Huang, Chang and Liu, Wenyu and Wang, Xinggang},
  journal={arXiv preprint arXiv:2410.22313},
  year={2024}
}

@inproceedings{e2e-vlm:lmdrive,
  title={Lmdrive: Closed-loop end-to-end driving with large language models},
  author={Shao, Hao and Hu, Yuxuan and Wang, Letian and Song, Guanglu and Waslander, Steven L and Liu, Yu and Li, Hongsheng},
  booktitle={Proceedings of the IEEE/CVF Conference on Computer Vision and Pattern Recognition},
  pages={15120--15130},
  year={2024}
}

@article{e2e-vlm:alphadrive,
  title={Alphadrive: Unleashing the power of vlms in autonomous driving via reinforcement learning and reasoning},
  author={Jiang, Bo and Chen, Shaoyu and Zhang, Qian and Liu, Wenyu and Wang, Xinggang},
  journal={arXiv preprint arXiv:2503.07608},
  year={2025}
}

@article{e2e-vlm:driveagent-r1,
  title={DriveAgent-R1: Advancing VLM-based Autonomous Driving with Active Perception and Hybrid Thinking},
  author={Zheng, Weicheng and Mao, Xiaofei and Ye, Nanfei and Li, Pengxiang and Zhan, Kun and Lang, Xianpeng and Zhao, Hang},
  journal={arXiv preprint arXiv:2507.20879},
  year={2025}
}

@article{e2e-vlm:fastdrivevla,
  title={Fastdrivevla: Efficient end-to-end driving via plug-and-play reconstruction-based token pruning},
  author={Cao, Jiajun and Zhang, Qizhe and Jia, Peidong and Zhao, Xuhui and Lan, Bo and Zhang, Xiaoan and Li, Zhuo and Wei, Xiaobao and Chen, Sixiang and Li, Liyun and others},
  journal={arXiv preprint arXiv:2507.23318},
  year={2025}
}

@article{e2e-vlm:autodrive-r2,
  title={AutoDrive-R$^2$: Incentivizing Reasoning and Self-Reflection Capacity for VLA Model in Autonomous Driving},
  author={Yuan, Zhenlong and Qian, Chengxuan and Tang, Jing and Chen, Rui and Song, Zijian and Sun, Lei and Chu, Xiangxiang and Cai, Yujun and Zhang, Dapeng and Li, Shuo},
  journal={arXiv preprint arXiv:2509.01944},
  year={2025}
}

@article{e2e-vlm:corevla,
  title={CoReVLA: A dual-stage end-to-end autonomous driving framework for long-tail scenarios via collect-and-refine},
  author={Fang, Shiyu and Cui, Yiming and Liang, Haoyang and Lv, Chen and Hang, Peng and Sun, Jian},
  journal={arXiv preprint arXiv:2509.15968},
  year={2025}
}

@article{e2e-vlm:vdrive,
  title={VDRive: Leveraging reinforced VLA and diffusion policy for end-to-end autonomous driving},
  author={Guo, Ziang and Zhang, Zufeng},
  journal={arXiv preprint arXiv:2510.15446},
  year={2025}
}

@article{e2e-vlm:drivlm,
  title={DriVLM: Domain Adaptation of Vision-Language Models in Autonomous Driving},
  author={Zheng, Xuran and Yoo, Chang D},
  journal={arXiv preprint arXiv:2501.05081},
  year={2025}
}

@article{e2e-vlm:drive-r1,
  title={Drive-R1: Bridging Reasoning and Planning in VLMs for Autonomous Driving with Reinforcement Learning},
  author={Li, Yue and Tian, Meng and Zhu, Dechang and Zhu, Jiangtong and Lin, Zhenyu and Xiong, Zhiwei and Zhao, Xinhai},
  journal={arXiv preprint arXiv:2506.18234},
  year={2025}
}

@article{e2e-vlm:reasoning-vla,
  title={Reasoning-VLA: A fast and general vision-language-action reasoning model for autonomous driving},
  author={Zhang, Dapeng and Yuan, Zhenlong and Chen, Zhangquan and Liao, Chih-Ting and Chen, Yinda and Shen, Fei and Zhou, Qingguo and Chua, Tat-Seng},
  journal={arXiv preprint arXiv:2511.19912},
  year={2025}
}

@article{e2e-vlm:uniugp,
  title={UniUGP: Unifying understanding, generation, and planing for end-to-end autonomous driving},
  author={Lu, Hao and Liu, Ziyang and Jiang, Guangfeng and Luo, Yuanfei and Chen, Sheng and Zhang, Yangang and Chen, Ying-Cong},
  journal={arXiv preprint arXiv:2512.09864},
  year={2025}
}

@article{e2e-vlm:e3ad,
  title={E3AD: An emotion-aware vision-language-action model for human-centric end-to-end autonomous driving},
  author={Tang, Yihong and Liao, Haicheng and Nie, Tong and He, Junlin and Qu, Ao and Chen, Kehua and Ma, Wei and Li, Zhenning and Sun, Lijun and Xu, Chengzhong},
  journal={arXiv preprint arXiv:2512.04733},
  year={2025}
}

@article{e2e-vlm:autovla,
  title={AutoVLA: A Vision-Language-Action Model for End-to-End Autonomous Driving with Adaptive Reasoning and Reinforcement Fine-Tuning},
  author={Zhou, Zewei and Cai, Tianhui and Zhao, Seth Z and Zhang, Yun and Huang, Zhiyu and Zhou, Bolei and Ma, Jiaqi},
  journal={arXiv preprint arXiv:2506.13757},
  year={2025}
}

@article{e2e-vlm:impromptuvla,
  title={Impromptu VLA: Open Weights and Open Data for Driving Vision-Language-Action Models},
  author={Chi, Haohan and Gao, Huan-ang and Liu, Ziming and Liu, Jianing and Liu, Chenyu and Li, Jinwei and Yang, Kaisen and Yu, Yangcheng and Wang, Zeda and Li, Wenyi and others},
  journal={arXiv preprint arXiv:2505.23757},
  year={2025}
}

@article{e2e-vlm:bevdriver,
  title={BEVDriver: Leveraging BEV Maps in LLMs for Robust Closed-Loop Driving},
  author={Winter, Katharina and Azer, Mark and Flohr, Fabian B},
  journal={arXiv preprint arXiv:2503.03074},
  year={2025}
}

@article{e2e-vlm:dsdrive,
  title={DSDrive: Distilling Large Language Model for Lightweight End-to-End Autonomous Driving with Unified Reasoning and Planning},
  author={Liu, Wenru and Liu, Pei and Ma, Jun},
  journal={arXiv preprint arXiv:2505.05360},
  year={2025}
}

@inproceedings{e2e-vlm:covla-agent,
  title={Covla: Comprehensive vision-language-action dataset for autonomous driving},
  author={Arai, Hidehisa and Miwa, Keita and Sasaki, Kento and Watanabe, Kohei and Yamaguchi, Yu and Aoki, Shunsuke and Yamamoto, Issei},
  booktitle={2025 IEEE/CVF Winter Conference on Applications of Computer Vision (WACV)},
  pages={1933--1943},
  year={2025},
  organization={IEEE}
}

@inproceedings{e2e-vlm:orion,
  title={Orion: A holistic end-to-end autonomous driving framework by vision-language instructed action generation},
  author={Fu, Haoyu and Zhang, Diankun and Zhao, Zongchuang and Cui, Jianfeng and Liang, Dingkang and Zhang, Chong and Zhang, Dingyuan and Xie, Hongwei and Wang, Bing and Bai, Xiang},
  booktitle={Proceedings of the IEEE/CVF International Conference on Computer Vision},
  pages={24823--24834},
  year={2025}
}

@article{e2e-vlm:senna-2,
  title={Senna-2: Aligning VLM and End-to-End Driving Policy for Consistent Decision Making and Planning},
  author={Song, Yuehao and Chen, Shaoyu and Gao, Hao and Zhu, Yifan and Yue, Weixiang and Zou, Jialv and Jiang, Bo and Lu, Zihao and Wang, Yu and Zhang, Qian and others},
  journal={arXiv preprint arXiv:2603.11219},
  year={2026}
}

@article{causal:e2e:causal_confusion,
  title={Causal confusion in imitation learning},
  author={De Haan, Pim and Jayaraman, Dinesh and Levine, Sergey},
  journal={Advances in neural information processing systems},
  volume={32},
  year={2019}
}

@article{causal:e2e:ad-mlp,
  title={Rethinking the open-loop evaluation of end-to-end autonomous driving in nuScenes (2023)},
  author={Zhai, JT and Feng, Z and Du, J and Mao, Y and Liu, JJ and Tan, Z and Zhang, Y and Ye, X and Wang, J},
  journal={arXiv preprint arXiv:2305.10430},
  year={2023}
}

@inproceedings{causal:e2e:driveadapter,
  title={Driveadapter: Breaking the coupling barrier of perception and planning in end-to-end autonomous driving},
  author={Jia, Xiaosong and Gao, Yulu and Chen, Li and Yan, Junchi and Liu, Patrick Langechuan and Li, Hongyang},
  booktitle={Proceedings of the IEEE/CVF International Conference on Computer Vision},
  pages={7953--7963},
  year={2023}
}

@article{causal:e2e:exploring_causality_in_e2e,
  title={Exploring the causality of end-to-end autonomous driving},
  author={Li, Jiankun and Li, Hao and Liu, Jiangjiang and Zou, Zhikang and Ye, Xiaoqing and Wang, Fan and Huang, Jizhou and Wu, Hua and Wang, Haifeng},
  journal={arXiv preprint arXiv:2407.06546},
  year={2024}
}

@inproceedings{causal:e2e:cadet,
  title={Cadet: a causal disentanglement approach for robust trajectory prediction in autonomous driving},
  author={Pourkeshavarz, Mozhgan and Zhang, Junrui and Rasouli, Amir},
  booktitle={Proceedings of the IEEE/CVF Conference on Computer Vision and Pattern Recognition},
  pages={14874--14884},
  year={2024}
}

@inproceedings{causal:e2e:decoupling_perception_and_ego_status,
  title={Decoupling scene perception and ego status: A multi-context fusion approach for enhanced generalization in end-to-end autonomous driving},
  author={Tang, Jiacheng and Feng, Mingyue and Liu, Jiachao and Wang, Yaonong and Pu, Jian},
  booktitle={Proceedings of the AAAI Conference on Artificial Intelligence},
  volume={40},
  number={11},
  pages={9413--9420},
  year={2026}
}

@article{causal:e2e:causalvad,
  title={CausalVAD: De-confounding End-to-End Autonomous Driving via Causal Intervention},
  author={Tang, Jiacheng and Zhou, Zhiyuan and He, Zhuolin and Zhang, Jia and Zhang, Kai and Pu, Jian},
  journal={arXiv preprint arXiv:2603.18561},
  year={2026}
}

@inproceedings{causal:e2e-vlm:omnidrive,
  title={Omnidrive: A holistic vision-language dataset for autonomous driving with counterfactual reasoning},
  author={Wang, Shihao and Yu, Zhiding and Jiang, Xiaohui and Lan, Shiyi and Shi, Min and Chang, Nadine and Kautz, Jan and Li, Ying and Alvarez, Jose M},
  booktitle={Proceedings of the Computer Vision and Pattern Recognition Conference},
  pages={22442--22452},
  year={2025}
}

@inproceedings{causal:e2e-vlm:simlingo,
  title={Simlingo: Vision-only closed-loop autonomous driving with language-action alignment},
  author={Renz, Katrin and Chen, Long and Arani, Elahe and Sinavski, Oleg},
  booktitle={Proceedings of the Computer Vision and Pattern Recognition Conference},
  pages={11993--12003},
  year={2025}
}

@article{causal:e2e-vlm:alpamayo,
  title={Alpamayo-r1: Bridging reasoning and action prediction for generalizable autonomous driving in the long tail},
  author={Wang, Yan and Luo, Wenjie and Bai, Junjie and Cao, Yulong and Che, Tong and Chen, Ke and Chen, Yuxiao and Diamond, Jenna and Ding, Yifan and Ding, Wenhao and others},
  journal={arXiv preprint arXiv:2511.00088},
  year={2025}
}

@article{causal:e2e-vlm:coc-vla,
  title={CoC-VLA: Delving into Adversarial Domain Transfer for Explainable Autonomous Driving via Chain-of-Causality Visual-Language-Action Model},
  author={Zhang, Dapeng and Shen, Fei and Zhao, Rui and Chen, Yinda and Zhi, Peng and Li, Chenyang and Zhou, Rui and Zhou, Qingguo},
  journal={arXiv preprint arXiv:2511.19914},
  year={2025}
}

@article{causal:e2e-vlm:more_than_meets_the_eye,
  title={More Than Meets the Eye? Uncovering the Reasoning-Planning Disconnect in Training Vision-Language Driving Models},
  author={Song, Xurui and Huai, Shuo and Jiang, JingJing and Kong, Jiayi and Luo, Jun},
  journal={arXiv preprint arXiv:2510.04532},
  year={2025}
}

@article{causal:e2e-vlm:from_representational,
  title={From Representational Complementarity to Dual Systems: Synergizing VLM and Vision-Only Backbones for End-to-End Driving},
  author={Ang, Sining and Yang, Yuguang and Dang, Chenxu and Chen, Canyu and Chi, Cheng and Liu, Haiyan and Mao, Xuanyao and Bao, Jason and Sun, Bingchuan and Wang, Yan and others},
  journal={arXiv e-prints},
  pages={arXiv--2602},
  year={2026}
}

@misc{causal:e2e-vlm:grounding_shapes_plan,
      title={Grounding What Shapes the Plan: Rethinking Groundedness for Physical Intelligence in Autonomous Driving}, 
      author={Minkyoung Cho and Zewei Zhou and Wenhao Ding and Shuhan Tan and Boyi Li and Yuxiao Chen and Yan Wang and Zheng Lian and Min-Hung Chen and Chaowei Xiao and Zhuoqing Mao and Boris Ivanovic and Marco Pavone and Yulong Cao},
      year={2026},
      eprint={2610.07521},
      archivePrefix={arXiv},
      primaryClass={cs.AI},
      url={https://arxiv.org/abs/2610.07521}, 
}

@misc{causal:e2e-vlm:causaldrivebench,
      title={CausalDriveBench: Evaluating Causal Reasoning in Vision-Language-Action Models for Autonomous Driving}, 
      author={Narendiran Chembu and Navvrat Rao and Shreedhar Shreeshail Kodate and Gayatri Srujana Banda and Arko Sarkar and Abhinav Khanna and Rajarshee Das and Umesh Kanala and Siddarth Khandelwal and Kumar Aman and Aish Dubey and Kaustubh Beedkar and Arjun Jain},
      year={2026},
      eprint={2609.32157},
      archivePrefix={arXiv},
      primaryClass={cs.CV},
      url={https://arxiv.org/abs/2609.32157}, 
}

@article{robotics:cast,
  title={CAST: Counterfactual labels improve instruction following in vision-language-action models},
  author={Glossop, Catherine and Chen, William and Bhorkar, Arjun and Shah, Dhruv and Levine, Sergey},
  journal={arXiv preprint arXiv:2508.13446},
  year={2025}
}

@article{robotics:chop,
  title={CHOP: Counterfactual Human Preference Labels Improve Obstacle Avoidance in Visuomotor Navigation Policies},
  author={Seneviratne, Gershom and An, Jianyu and Shende, Vaibhav and Ellahy, Sahire and Amin, Yaxita and Manasanjani, Kondapi and Chopra, Samarth and Kannan, Jonathan Deepak and Manocha, Dinesh},
  journal={arXiv preprint arXiv:2603.02004},
  year={2026}
}

@InProceedings{robotics:linking_causal_to_gen_in_vla,
  title = 	 {Embodied Interpretability: Linking Causal Understanding to Generalization in Vision-Language-Action Models},
  author =       {Zhang, Hanxin and Xu, Mingshuo and Dhafer, Abdulqader and Yue, Shigang and Dong, Hongbiao and Hao, Zhou Daniel},
  booktitle = 	 {Proceedings of the 43rd International Conference on Machine Learning},
  pages = 	 {159526--159549},
  year = 	 {2026},
  editor = 	 {Zhang, Tong and Dudik, Miroslav and Jaggi, Martin and Agarwal, Alekh and Li, Sharon and Schuurmans, Dale and Zhu, Jerry and Berkenkamp, Felix and Dong, Hanze and Bietti, Alberto},
  volume = 	 {306},
  series = 	 {Proceedings of Machine Learning Research},
  month = 	 {06--11 Jul},
  publisher =    {PMLR}
}

@article{rlvr:grpo,
  title={Deepseekmath: Pushing the limits of mathematical reasoning in open language models},
  author={Shao, Zhihong and Wang, Peiyi and Zhu, Qihao and Xu, Runxin and Song, Junxiao and Bi, Xiao and Zhang, Haowei and Zhang, Mingchuan and Li, YK and Wu, Yang and others},
  journal={arXiv preprint arXiv:2402.03300},
  year={2024}
}

@article{rlvr:gdpo,
  title={Gdpo: Group reward-decoupled normalization policy optimization for multi-reward rl optimization},
  author={Liu, Shih-Yang and Dong, Xin and Lu, Ximing and Diao, Shizhe and Belcak, Peter and Liu, Mingjie and Chen, Min-Hung and Yin, Hongxu and Wang, Yu-Chiang Frank and Cheng, Kwang-Ting and others},
  journal={arXiv preprint arXiv:2601.05242},
  year={2026}
}

@article{metric:passk,
  title={Large language monkeys: Scaling inference compute with repeated sampling},
  author={Brown, Bradley and Juravsky, Jordan and Ehrlich, Ryan and Clark, Ronald and Le, Quoc V and R{\'e}, Christopher and Mirhoseini, Azalia},
  journal={arXiv preprint arXiv:2407.21787},
  year={2024}
}

@article{metric:passk:example:atlas,
  title={Atlas: Orchestrating Heterogeneous Models and Tools for Multi-Domain Complex Reasoning},
  author={Wu, Jinyang and Zhai, Guocheng and Jin, Ruihan and Yuan, Jiahao and Shen, Yuhao and Zhang, Shuai and Wen, Zhengqi and Tao, Jianhua},
  journal={arXiv preprint arXiv:2601.03872},
  year={2026}
}

@inproceedings{dataset:nusc,
  title={nuscenes: A multimodal dataset for autonomous driving},
  author={Caesar, Holger and Bankiti, Varun and Lang, Alex H and Vora, Sourabh and Liong, Venice Erin and Xu, Qiang and Krishnan, Anush and Pan, Yu and Baldan, Giancarlo and Beijbom, Oscar},
  booktitle={Proceedings of the IEEE/CVF conference on computer vision and pattern recognition},
  pages={11621--11631},
  year={2020}
}

@inproceedings{dataset:ccd,
  title={Uncertainty-based traffic accident anticipation with spatio-temporal relational learning},
  author={Bao, Wentao and Yu, Qi and Kong, Yu},
  booktitle={Proceedings of the 28th ACM International Conference on Multimedia},
  pages={2682--2690},
  year={2020}
}

@inproceedings{dataset:VIENA,
  title={VIENA: A driving anticipation dataset},
  author={Aliakbarian, Mohammad Sadegh and Saleh, Fatemeh Sadat and Salzmann, Mathieu and Fernando, Basura and Petersson, Lars and Andersson, Lars},
  booktitle={Asian Conference on Computer Vision},
  pages={449--466},
  year={2018},
  organization={Springer}
}

@inproceedings{dataset:GTACrash,
  title={Crash to not crash: Learn to identify dangerous vehicles using a simulator},
  author={Kim, Hoon and Lee, Kangwook and Hwang, Gyeongjo and Suh, Changho},
  booktitle={Proceedings of the AAAI Conference on Artificial Intelligence},
  volume={33},
  number={01},
  pages={978--985},
  year={2019}
}

@inproceedings{dataset:Argus,
  title={Road traffic accidents detection based on crash estimation},
  author={Sabry, Khaled and Emad, Mohamed},
  booktitle={2021 17th International Computer Engineering Conference (ICENCO)},
  pages={63--68},
  year={2021},
  organization={IEEE}
}

@article{dataset:YoutubeCrash,
  title={Predicting vehicle collisions using data collected from video games},
  author={Kim, Hoon and Lee, Kangwook and Hwang, Gyeongjo and Suh, Changho},
  journal={Machine Vision and Applications},
  volume={32},
  number={4},
  pages={93},
  year={2021},
  publisher={Springer}
}

@article{dataset:IITH_Road,
  title={Deep learning based detection and localization of road accidents from traffic surveillance videos},
  author={Pawar, Karishma and Attar, Vahida},
  journal={ICT Express},
  volume={8},
  number={3},
  pages={379--387},
  year={2022},
  publisher={Elsevier}
}

@article{dataset:TAD,
  title={TAD: A large-scale benchmark for traffic accidents detection from video surveillance},
  author={Xu, Yajun and Hu, Huan and Huang, Chuwen and Nan, Yibing and Liu, Yuyao and Wang, Kai and Liu, Zhaoxiang and Lian, Shiguo},
  journal={IEEE Access},
  volume={13},
  pages={2018--2033},
  year={2024},
  publisher={IEEE}
}

@article{dataset:MP-RAD,
  title={Detection of road accidents using synthetically generated multi-perspective accident videos},
  author={Vijay, Thakare Kamalakar and Dogra, Debi Prosad and Choi, Heeseung and Nam, Gipyo and Kim, Ig-Jae},
  journal={IEEE Transactions on Intelligent Transportation Systems},
  volume={24},
  number={2},
  pages={1926--1935},
  year={2022},
  publisher={IEEE}
}

@inproceedings{dataset:riskbench,
  title={Riskbench: A scenario-based benchmark for risk identification},
  author={Kung, Chi-Hsi and Yang, Chieh-Chi and Pao, Pang-Yuan and Lu, Shu-Wei and Chen, Pin-Lun and Lu, Hsin-Cheng and Chen, Yi-Ting},
  booktitle={2024 IEEE International Conference on Robotics and Automation (ICRA)},
  pages={14800--14807},
  year={2024},
  organization={IEEE}
}

@inproceedings{dataset:DeepAccident,
  title={Deepaccident: A motion and accident prediction benchmark for v2x autonomous driving},
  author={Wang, Tianqi and Kim, Sukmin and Wenxuan, Ji and Xie, Enze and Ge, Chongjian and Chen, Junsong and Li, Zhenguo and Luo, Ping},
  booktitle={Proceedings of the AAAI Conference on Artificial Intelligence},
  volume={38},
  number={6},
  pages={5599--5606},
  year={2024}
}

@inproceedings{dataset:mm_au,
  title={Abductive ego-view accident video understanding for safe driving perception},
  author={Fang, Jianwu and Li, Lei-lei and Zhou, Junfei and Xiao, Junbin and Yu, Hongkai and Lv, Chen and Xue, Jianru and Chua, Tat-Seng},
  booktitle={Proceedings of the IEEE/CVF Conference on Computer Vision and Pattern Recognition},
  pages={22030--22040},
  year={2024}
}

@article{dataset:coool,
  title={Coool: Challenge of out-of-label a novel benchmark for autonomous driving},
  author={AlShami, Ali K and Kalita, Ananya and Rabinowitz, Ryan and Lam, Khang and Bezbarua, Rishabh and Boult, Terrance and Kalita, Jugal},
  journal={arXiv preprint arXiv:2412.05462},
  year={2024}
}

@inproceedings{dataset:nexar,
  title={Nexar Dashcam Collision Prediction Dataset and Challenge},
  author={Moura, Daniel and Zhu, Shizhan and Zvitia, Orly},
  booktitle={Proceedings of the Computer Vision and Pattern Recognition Conference},
  pages={2583--2591},
  year={2025}
}

@inproceedings{dataset:TUMTraf_Accid3nD,
  title={Towards Vision Zero: The TUM Traffic Accid3nD Dataset},
  author={Zimmer, Walter and Greer, Ross and Zhou, Xingcheng and Song, Rui and Cao, Hu and Lehmberg, Daniel and Pavel, Marc and Ghita, Ahmed Alaaeldin and Gopalkrishnan, Akshay and Caesar, Holger and others},
  booktitle={Proceedings of the IEEE/CVF International Conference on Computer Vision},
  pages={841--851},
  year={2025}
}

@inproceedings{dataset:lingoqa,
  title={Lingoqa: Visual question answering for autonomous driving},
  author={Marcu, Ana-Maria and Chen, Long and H{\"u}nermann, Jan and Karnsund, Alice and Hanotte, Benoit and Chidananda, Prajwal and Nair, Saurabh and Badrinarayanan, Vijay and Kendall, Alex and Shotton, Jamie and others},
  booktitle={European Conference on Computer Vision},
  pages={252--269},
  year={2024},
  organization={Springer}
}

@article{dataset:bench2drive,
  title={Bench2drive: Towards multi-ability benchmarking of closed-loop end-to-end autonomous driving},
  author={Jia, Xiaosong and Yang, Zhenjie and Li, Qifeng and Zhang, Zhiyuan and Yan, Junchi},
  journal={Advances in Neural Information Processing Systems},
  volume={37},
  pages={819--844},
  year={2024}
}

@article{dataset:navsim,
  title={Navsim: Data-driven non-reactive autonomous vehicle simulation and benchmarking},
  author={Dauner, Daniel and Hallgarten, Marcel and Li, Tianyu and Weng, Xinshuo and Huang, Zhiyu and Yang, Zetong and Li, Hongyang and Gilitschenski, Igor and Ivanovic, Boris and Pavone, Marco and others},
  journal={Advances in Neural Information Processing Systems},
  volume={37},
  pages={28706--28719},
  year={2024}
}

@article{dataset:fail2drive,
  title={Fail2Drive: Benchmarking Closed-Loop Driving Generalization},
  author={Gerstenecker, Simon and Geiger, Andreas and Renz, Katrin},
  journal={arXiv preprint arXiv:2604.08535},
  year={2026}
}

@inproceedings{attention:grad-cam,
  title={Grad-cam: Visual explanations from deep networks via gradient-based localization},
  author={Selvaraju, Ramprasaath R and Cogswell, Michael and Das, Abhishek and Vedantam, Ramakrishna and Parikh, Devi and Batra, Dhruv},
  booktitle={Proceedings of the IEEE international conference on computer vision},
  pages={618--626},
  year={2017}
}

@InProceedings{attention:tam,
    author    = {Li, Yi and Wang, Hualiang and Ding, Xinpeng and Wang, Haonan and Li, Xiaomeng},
    title     = {Token Activation Map to Visually Explain Multimodal LLMs},
    booktitle = {Proceedings of the IEEE/CVF International Conference on Computer Vision (ICCV)},
    month     = {October},
    year      = {2025},
    pages     = {48-58}
}


\clearpage
\appendix

\renewcommand{\thesection}{\Alph{section}} 

\startcontents[appendix] 
\section*{Appendix} 
\printcontents[appendix] 
               {}{1} 
               {\setcounter{tocdepth}{2}} 

\clearpage

\section{Discussion}
\label{appendix:discussion}
\paragraph{Why Does CF Reasoning Induce Behavioral Shifts Without Visual Grounding?} A key factor is that CF reasoning in current VLM-based autonomous driving systems is primarily learned at the language level. During supervised fine-tuning, supervision is provided only in the form of textual reasoning outputs, without explicit alignment between these texts and the corresponding visual evidence. As a result, CF reasoning is optimized as a language modeling objective rather than a visually grounded causal reasoning, leading to a natural decoupling between linguistic reasoning and visual causal structure. This interpretation is consistent with our observations of invariant attention patterns and non-selective responses under causal intervention. These findings further challenge the implicit assumption that performance improvements from language-based reasoning necessarily reflect genuine causal understanding. In practice, both LLMs and VLMs exhibit fundamental limitations in stable causal inference, particularly under distribution shift in long-tail scenarios \cite{causal:whatif_tv_off?, causal:llm_cannot_discover_causality, dataset:fail2drive}. 

\paragraph{Is Visual Grounding Sufficient for Causal Decision-Making?} To address the limitation, prior work has proposed leveraging counterfactual consistency to encourage reliance on visual evidence rather than purely linguistic patterns \cite{causal:defacto} and further explored integrating LLM-based agents with causal world models, extending decision-making beyond next-token prediction toward structured environments that support intervention and counterfactual reasoning \cite{causal:bridging_llms_and_causal_world_model}. These directions highlight the importance of developing CF reasoning as a learning signal that is aligned with physical grounding, rather than relying solely on linguistic supervision. However, visual grounding alone does not necessarily ensure that the resulting reasoning is effectively reflected in downstream decision-making. In other words, establishing a correspondence between reasoning and visual evidence is distinct from ensuring that such reasoning appropriately guides the model's actions \cite{causal:e2e-vlm:grounding_shapes_plan,robotics:linking_causal_to_gen_in_vla}.

\paragraph{Future Directions.}
Despite their ability to capture rich correlational structures and derive coarse semantic representations of scenes in autonomous driving through large-scale pretraining \cite{causal:e2e-vlm:from_representational}, VLMs still lack reliable causal reasoning capabilities. Therefore, the implicit assumption that language-based reasoning enables VLMs to perform causal reasoning in autonomous driving, and that improvements in driving performance necessarily reflect genuine causal understanding, warrants reconsideration. This calls for evaluation paradigms that go beyond conventional driving performance metrics to explicitly examine the gap between causal understanding and downstream driving behavior. In particular, evaluating the alignment between reasoning and action could provide insights into whether the model's reasoning genuinely informs its decisions, rather than merely accompanying behavioral improvements.

Furthermore, given the substantial challenges involved in endowing VLMs with explicit causal reasoning capabilities, an alternative research direction is to pursue more tractable objectives that mitigate the influence of causal confounders, such as linguistic shortcuts. Rather than directly solving the broader problem of causal reasoning, identifying and suppressing such shortcuts may offer a practical step toward more reliable and causally grounded decision-making \cite{causal:e2e:causalvad, causal:e2e:cadet, causal:e2e:decoupling_perception_and_ego_status}.

\section{Detailed Implementation}
\label{appendix:implementations}
\subsection{Group Relative Policy Optimization (GRPO)}
\label{appendix:implementations_grpo}
For each query $q$ (comprising an input image and prompt), GRPO samples a group of $G$ outputs $\{o_1, o_2, \dots, o_G\}$ from the old policy $\pi_{\theta_{\text{old}}}$. It then calculates the reward $r_i$ for each output and updates the policy using relative advantages computed based on the group mean. The objective function is defined as:

$$\begin{aligned}
\mathcal{J}_{\mathrm{GRPO}}(\theta) &= \mathbb{E}_{q \sim P(Q), \{o_i\}_{i=1}^G \sim \pi_{\theta_{\text{old}}}} \left[ \frac{1}{G} \sum_{i=1}^G \mathcal{J}_i - \beta \mathbb{D}_{\mathrm{KL}}(\pi_\theta \| \pi_{\text{ref}}) \right], \\
\mathcal{J}_i &= \min \left( w_i \hat{A}_i, \text{clip}(w_i, 1-\epsilon, 1+\epsilon) \hat{A}_i \right),
\end{aligned}$$

where $w_i = \frac{\pi_\theta(o_i|q)}{\pi_{\theta_{\text{old}}}(o_i|q)}$ is the importance sampling ratio, and $\hat{A}_i$ denotes the group-relative advantage.

\subsection{Group reward-Decoupled Normalization Policy Optimization (GDPO)}
\label{appendix:implementations_gdpo}
Unlike GRPO, which directly incorporates group-relative advantages into the policy objective, GDPO decouples group-wise reward normalization from policy optimization. Specifically, the normalized rewards $\tilde{r}_i$ are treated as fixed advantages, and no gradients are propagated through the normalization statistics $\mu_q$ and $\sigma_q$. The GDPO objective is formulated in a PPO-style manner as:

$$\begin{aligned}
\mathcal{J}_{\mathrm{GDPO}}(\theta) &= \mathbb{E}_{q \sim P(Q), \{o_i\}_{i=1}^G \sim \pi_{\theta_{\text{old}}}} \left[ \frac{1}{G} \sum_{i=1}^G \mathcal{L}_i^{\mathrm{GDPO}} - \beta \mathbb{D}_{\mathrm{KL}}(\pi_\theta \| \pi_{\text{ref}}) \right], \\
\mathcal{L}_i^{\mathrm{GDPO}} &= \min \left( w_i \tilde{r}_i, \text{clip}(w_i, 1-\epsilon, 1+\epsilon)\tilde{r}_i \right),
\end{aligned}$$

where $w_i = \frac{\pi_\theta(o_i|q)}{\pi_{\theta_{\text{old}}}(o_i|q)}$ denotes the same importance sampling ratio used in GRPO.

\subsection{Pass@k}
\label{appendix:implementations_passk}
\begin{equation}
\text{pass}@k := \mathbb{E} \left[ 1 - \frac{\binom{n-c}{k}}{\binom{n}{k}} \right]
\label{eq:pass_at_k}
\end{equation}
\noindent where $n$ is the total number of generated samples, and $c$ denotes the count of correct (safe) outputs. This metric provides a more robust evaluation of the model's potential than a single-shot prediction, as it accounts for the probabilistic nature of the model's reasoning.

\subsection{Perplexity}
\label{appendix:implementations_perplexity}
Given a model $m$, a prompt $x$, and a sequence of tokens $Y = (y_1, \dots, y_T)$, the perplexity is defined as the exponentiated average negative log-likelihood:

$$
\mathrm{PPL}_m(Y \mid x) = \exp \left( -\frac{1}{T} \sum_{t=1}^{T} \log P_m(y_t \mid x, y_1, \ldots, y_{t-1}) \right)
$$

\subsection{Policy Entropy}
\label{appendix:implementations_policy_entropy}
\begin{equation}
    H_t(\pi_\theta) = -\sum_{v \in \mathcal{V}} \pi_\theta(v \mid \mathbf{x}, \mathbf{y}_{<t}) \log \pi_\theta(v \mid \mathbf{x}, \mathbf{y}_{<t})
\end{equation}

where $\mathcal{V}$ is the vocabulary. To ensure a fair comparison across models with varying response lengths (e.g., concise standard SFT vs. verbose Counterfactual models), we employ a \textit{length-normalized} average entropy:

\begin{equation}
    \mathcal{H}_{\text{policy}}(\pi_\theta, \mathbf{x}, \mathbf{y}) = \frac{1}{T} \sum_{t=1}^{T} H_t(\pi_\theta)
    \label{eq:policy_entropy}
\end{equation}

By normalizing for sequence length $T$, $\mathcal{H}_{\text{policy}}$ reduces the influence of verbosity on entropy comparisons. We note, however, that this normalization may not fully eliminate length-related effects, as per-token entropy tends to decrease toward the end of longer sequences (e.g., during repetitive closing patterns). Consequently, if CF models produce longer responses on average, the length-normalized estimator is likely to \textit{underestimate} their true entropy relative to shorter non-CF responses. This structural bias works \textit{against} our hypothesis, making the observed entropy elevation in CF models a conservative estimate of the underlying distributional shift. 

To further validate that the entropy elevation is not primarily driven by response length, we compute the Pearson correlation coefficient between mean completion length and policy entropy during RFT. As shown in Table~\ref{tab:entropy_corr}, the correlation is consistently lower for CF models (0.365--0.392) than for non-CF models (0.439--0.499) across all settings, suggesting that response length explains less of the entropy variance in CF models than in their non-CF counterparts.
\begin{table*}[h]
\centering
\begin{tabular}{lcc}
\toprule
\textbf{Model} & \textbf{w/ CF} & \textbf{w/o CF} \\
\midrule
InternVL2.5-1B (GRPO) & 0.365 & 0.499 \\
Qwen2.5-VL-3B (GRPO)  & 0.392 & 0.461 \\
Qwen2.5-VL-3B (GDPO)  & 0.370 & 0.439 \\
\bottomrule
\end{tabular}
\caption{Pearson correlation coefficient between mean completion length and policy entropy during RFT. CF models consistently exhibit lower correlation than their non-CF counterparts across all settings, suggesting that the entropy elevation in CF models is not primarily driven by response length.}
\label{tab:entropy_corr}
\end{table*}

\subsection{Metrics for Token Activation Map (TAM) Evaluation}
\label{appendix:implementations_iou_metrics}

To quantitatively evaluate whether counterfactual reasoning 
enhances visual grounding on accident-causing objects, we 
employ Token Activation Map (TAM)~\cite{attention:tam}, 
which attributes each output token's generation to specific 
visual regions. Unlike standard attention visualization that 
conflates multi-head and multi-layer interactions, TAM 
isolates the contribution of vision tokens to each decoded 
token through targeted perturbation of hidden states.

For a generated token $t_i$ at decoding step $i$, TAM 
computes the activation map $\mathbf{A}_i \in \mathbb{R}^{H \times W}$ 
by measuring the sensitivity of the token's logit to each 
vision token embedding. Specifically, given vision tokens 
$\{\mathbf{v}_j\}_{j=1}^{H \times W}$ from the vision encoder 
and the LLM hidden state $\mathbf{h}_i$ before the language 
head, TAM perturbs each vision token and observes the change 
in output probability:

\begin{equation}
\mathbf{A}_i(x, y) = \left| \log p(t_i | \mathbf{h}_i) - 
\log p(t_i | \mathbf{h}_i^{(x,y)\to 0}) \right|
\end{equation}

where $\mathbf{h}_i^{(x,y)\to 0}$ denotes the hidden state 
when vision token at spatial position $(x, y)$ is zeroed out. 
The resulting map $\mathbf{A}_i$ is then normalized to 
$[0, 1]$ and resized to the original image resolution for 
bounding box comparison. Based on it, we define three complementary metrics to measure visual grounding at different granularities:

\paragraph{Token-level IoU ($IoU_{tok}$)}
For each causal keyword token (e.g., ``car'', ``pedestrian'') 
in the generated response, we compute the Intersection over 
Union (IoU) between its activation map $\mathbf{A}_i$ and 
the ground-truth bounding box $\mathcal{B}_{causal}$ of the 
accident-causing object:

\begin{equation}
IoU_i = \frac{|\mathbf{M}_i \cap \mathcal{B}_{causal}|}
{|\mathbf{M}_i \cup \mathcal{B}_{causal}|}
\end{equation}

where $\mathbf{M}_i = \{\mathbf{A}_i \geq 0.2 \cdot \max(\mathbf{A}_i)\}$ 
is the binary mask obtained by thresholding at 20\% of the 
peak activation, following standard practice in CAM-based 
methods~\cite{attention:grad-cam}. Token-level IoU is then 
averaged over all causal tokens:

\begin{equation}
IoU_{tok} = \frac{1}{N_{causal}} \sum_{i \in \mathcal{T}_{causal}} IoU_i
\end{equation}

where $\mathcal{T}_{causal}$ is the set of causal keyword 
token indices and $N_{causal} = |\mathcal{T}_{causal}|$.

\paragraph{Aggregated IoU ($IoU_{agg}$)}
To capture the cumulative visual focus across the entire 
reasoning process, we aggregate activation maps from all 
causal tokens by summing their normalized maps:

\begin{equation}
\mathbf{A}_{agg} = \sum_{i \in \mathcal{T}_{causal}} 
\frac{\mathbf{A}_i - \min(\mathbf{A}_i)}
{\max(\mathbf{A}_i) - \min(\mathbf{A}_i)}
\end{equation}

The aggregated IoU is then computed on the combined map:

\begin{equation}
IoU_{agg} = \frac{|\mathbf{M}_{agg} \cap \mathcal{B}_{causal}|}
{|\mathbf{M}_{agg} \cup \mathcal{B}_{causal}|}
\end{equation}

where $\mathbf{M}_{agg} = \{\mathbf{A}_{agg} \geq 0.2 \cdot 
\max(\mathbf{A}_{agg})\}$. This metric reflects whether the 
model's attention, when viewed holistically across all 
reasoning tokens, aligns with the causal object.

\paragraph{Grounding F1 ($F1_{gnd}$)}
IoU alone may underweight scenarios where the model attends 
to the causal region but also includes spurious areas, or 
conversely, misses parts of the object. We therefore compute 
precision and recall on the aggregated binary mask:

\begin{align}
\text{Precision} &= \frac{|\mathbf{M}_{agg} \cap \mathcal{B}_{causal}|}
{|\mathbf{M}_{agg}|} \\
\text{Recall} &= \frac{|\mathbf{M}_{agg} \cap \mathcal{B}_{causal}|}
{|\mathcal{B}_{causal}|}
\end{align}

The Grounding F1 score balances these two aspects:

\begin{equation}
F1_{gnd} = \frac{2 \cdot \text{Precision} \cdot \text{Recall}}
{\text{Precision} + \text{Recall}}
\end{equation}

A high $F1_{gnd}$ indicates that the model's visual attention 
is both concentrated on the causal object (high precision) 
and covers it comprehensively (high recall), suggesting 
genuine visual grounding rather than diffuse or misaligned 
activation.

In summary, if counterfactual reasoning improves causal understanding, CF-enabled models are expected to exhibit higher $IoU_{tok}$, indicating that each reasoning token attends more precisely to the accident cause; higher $IoU_{agg}$, reflecting that the cumulative reasoning process focuses more consistently on the causal region; and higher $F1_{gnd}$, suggesting that the attention is both precise (i.e., low false positives) and comprehensive (i.e., low false negatives).

However, as shown in Table~\ref{tab:iou}, no such improvements 
are observed after instruction fine-tuning with counterfactual reasoning, indicating that the decision shifts via counterfactual reasoning operate independently of visual evidence.

\clearpage
\section{Dataset}
\label{appendix:dataset}

\subsection{Overview of the Dataset}

\begin{table*}[h]
\centering
\resizebox{\textwidth}{!}{%
\begin{tabular}{@{}l c c l r c c c c@{}}
\toprule
\textbf{Dataset} & \textbf{Year} & \textbf{Type} & \textbf{Perspect.} & \textbf{\#Frames} & \textbf{Acc. Recog.} & \textbf{Desc.} & \textbf{CF QA} & \textbf{Decision} \\
\midrule
VIENA$^2$ \cite{dataset:VIENA}  & 2019 & synth & vehicle & 2,250,000 & \cmark & \xmark & \xmark & \xmark \\
GTACrash \cite{dataset:GTACrash} & 2019 & synth & vehicle & 751,610 & \cmark & \xmark & \xmark & \xmark \\
Argus \cite{dataset:Argus}  & 2021 & real & roadside & 120,000 & \cmark & \xmark & \xmark & \xmark \\
YoutubeCrash \cite{dataset:YoutubeCrash} & 2021 & real & vehicle & 7,720 & \cmark & \xmark & \xmark & \xmark \\
IITH Road \cite{dataset:IITH_Road} & 2022 & real & roadside & 127,138 & \cmark & \xmark & \xmark & \xmark \\
TAD \cite{dataset:TAD} & 2022 & real & roadside & 24,810 & \cmark & \xmark & \xmark & \xmark \\
MP-RAD \cite{dataset:MP-RAD} & 2023 & synth & roadside & 366,600 & \cmark & \xmark & \xmark & \xmark \\
RiskBench \cite{dataset:riskbench} & 2024 & synth & vehicle & N/A & \cmark & \xmark & \xmark & \xmark \\
DeepAccident \cite{dataset:DeepAccident} & 2024 & synth & V2V \& V2I & 57,000 & \cmark & \xmark & \xmark & \xmark \\
MM-AU \cite{dataset:mm_au} & 2024 & real & vehicle & 2,190,000 & \cmark & \cmark & \xmark & \xmark \\
TUMTraf-Accid3nD \cite{dataset:TUMTraf_Accid3nD} & 2025 & real & roadside & 111,656 & \cmark & \cmark & \xmark & \xmark \\
\textbf{Ours} & 2026 & real & vehicle & \textbf{2,620,356} & \cmark & \cmark & \cmark & \cmark \\
\bottomrule
\end{tabular}%
}
\caption{Comparison of our testbed dataset with existing accident and traffic datasets. Unlike previous works limited to recognition, our dataset supports high-level reasoning tasks including description, counterfactual QA, and decision making.}
\label{tab:dataset_comparison}
\end{table*}

The testbed dataset used in this study was constructed by integrating \cite{dataset:ccd, dataset:coool, dataset:nexar} and applying captioning to each video. Importantly, each video in these datasets has been annotated by domain experts with labels indicating whether an accident occurs. Each dataset provides predefined labels indicating whether a given video corresponds to an accident scenario or a normal driving scenario. However, these existing driving accident datasets primarily focus on recognition of the scene and do not include driving actions or counterfactual supervision. As a result, they lack counterfactual captions and decision required to address the central question of this study: whether counterfactual reasoning truly affects safety-relevant performance.

\begin{table}[h]
\centering
\small
\begin{tabular}{lccc}
\toprule
\textbf{Decision Class} & \textbf{Train} & \textbf{Test} & \textbf{Total} \\
\midrule
stop & 1,026 (15.4\%) & 336 (25.0\%) & 1,362 \\
decelerate     & 1,423 (21.4\%) & 336 (25.0\%) & 1,759 \\
keep      & 4,212 (63.2\%) & 672 (50.0\%) & 4,884 \\
\midrule
\textbf{Total} & \textbf{6,661} & \textbf{1,344} & \textbf{8,005} \\
\bottomrule
\end{tabular}
\caption{Dataset composition. Test set is intentionally 
balanced for robust evaluation of safety-critical decisions.}
\end{table}

\subsection{Automated Annotation Pipeline}

\begin{figure}[h]
    \centering
    \includegraphics[width=\columnwidth]{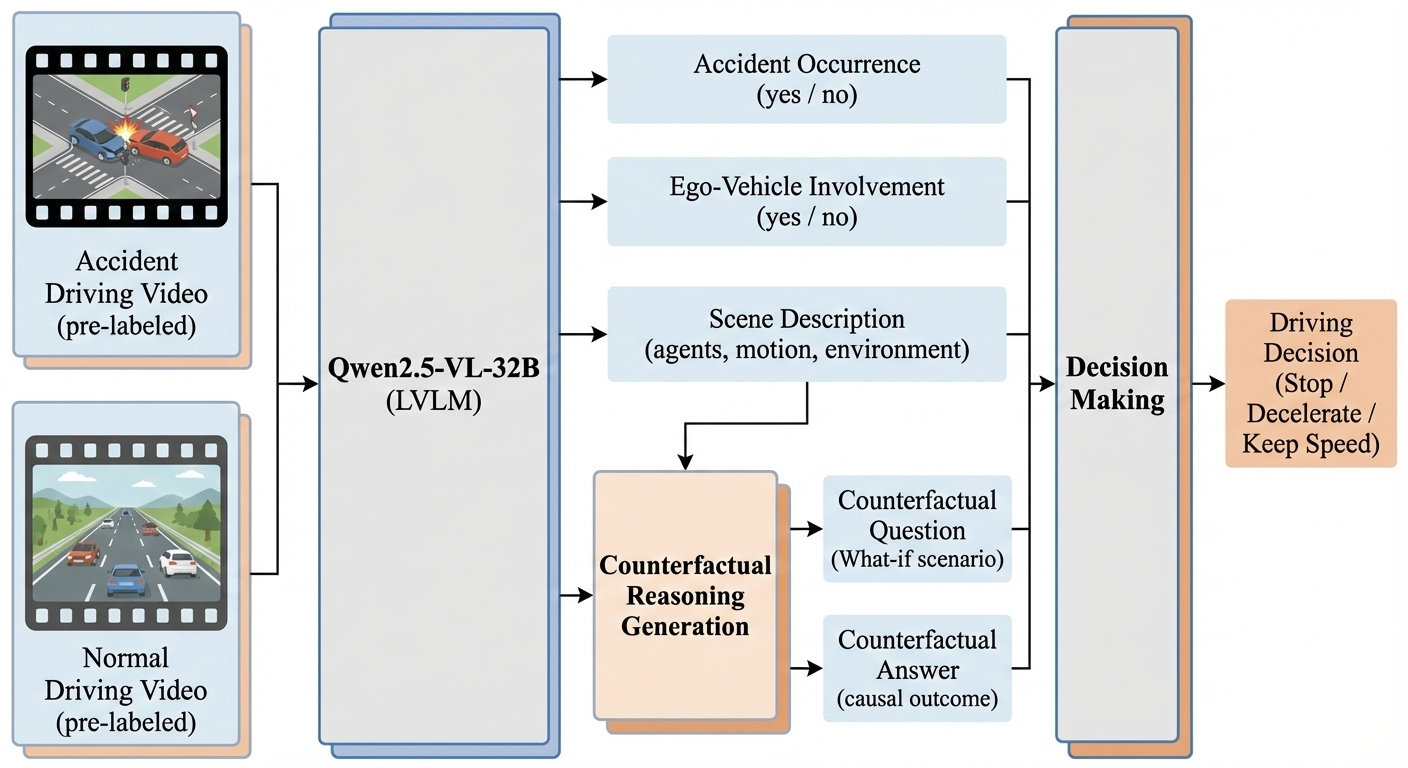}
    \caption{The automated annotation pipeline for the dataset.}
    \label{fig:automated_pipeline}
\end{figure}

To address these limitations, we introduce a multi-stage automated annotation pipeline to construct a dataset for analyzing whether counterfactual reasoning truly affects safety-relevant performance from existing accident driving videos. Drawing inspiration from the visual question answering methodology proposed in \cite{dataset:lingoqa}, we leverage the advanced multimodal capabilities of \texttt{Qwen2.5-VL-32B-Instruct} to generate structured captions \cite{e2e-vlm:orion, causal:cf_vla}. Our annotation process produces six key components for each video: \texttt{accident occurrence, ego-vehicle involvement, scene description, counterfactual question, counterfactual answer}, and \texttt{driving decision}. The pipeline consists of three sequential steps:

\begin{itemize}[leftmargin=*, noitemsep, topsep=0pt]
    \item \textbf{Step 1: Scene Recognition and Description.} 
    First, we feed the raw accident video into \texttt{Qwen2.5-VL-32B-Instruct} to extract fundamental scene information. The model is prompted to identify whether an accident occurred (\texttt{accident occurrence}) and whether the ego-vehicle was involved (\texttt{ego involvement}). Simultaneously, it generates a detailed textual explanation of the traffic scenario (\texttt{description}), focusing on the behavior of road agents and environmental conditions.

    \item \textbf{Step 2: Counterfactual Reasoning Generation.} 
    Based on the \texttt{description} obtained in Step 1, we guide the model to generate a \texttt{counterfactual question} and its corresponding \texttt{counterfactual answer}. This step aims to augment the dataset with reasoning capabilities by asking "what-if" scenarios (e.g., \textit{"what would have happened if the ego-vehicle had not braked?"}), thereby enhancing the model's understanding of causal relationships in safety-critical situations.

    \item \textbf{Step 3: Decision Making.} 
    Finally, we synthesize the \texttt{decision} for the original video. The model aggregates all prior information—accident status, ego-involvement, scene description, and counterfactual reasoning—to deduce the optimal driving maneuver (e.g., stop, decelerate, keep). This ensures that the final decision is logically grounded in both the observed reality and potential alternative outcomes.
\end{itemize}

\subsection{Quality Assurance}

\begin{figure}[h]
    \centering
    \includegraphics[width=\columnwidth]{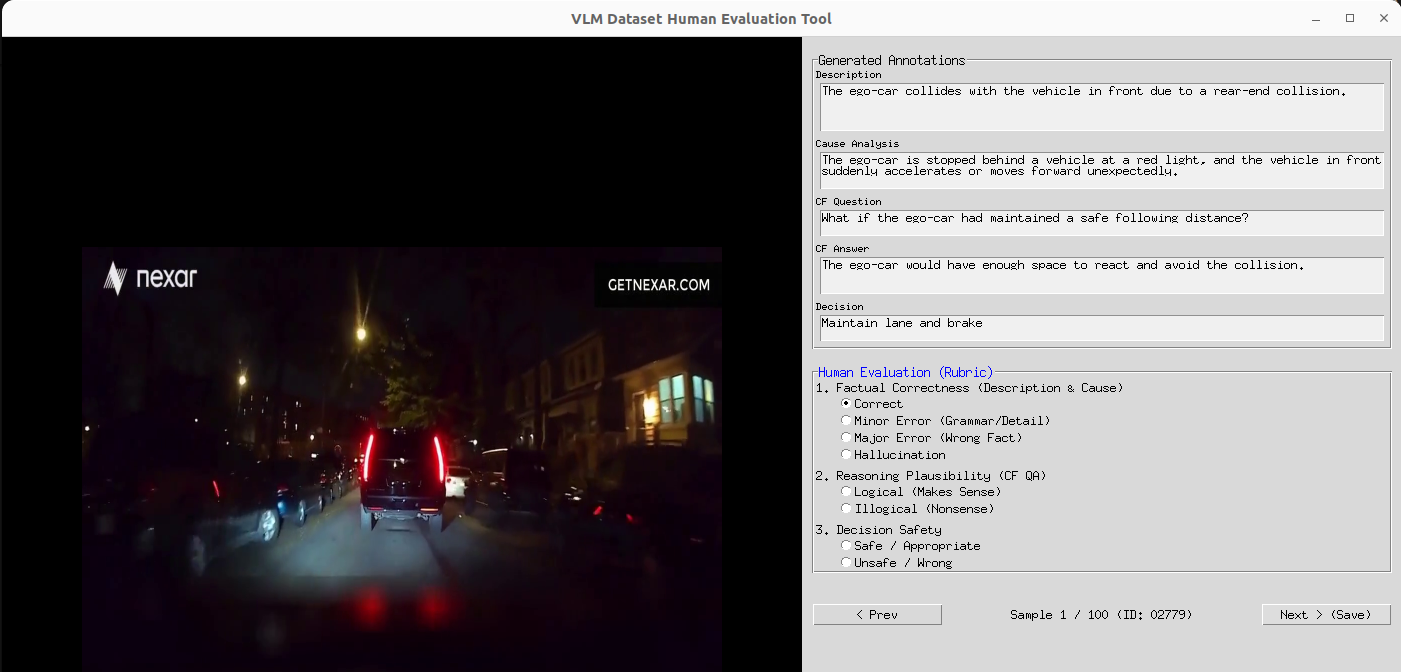}
    \caption{Diagnosis tool for quality assessment.}
    \label{fig:human_eval}
\end{figure}

To ensure the reliability of our dataset while maintaining the scalability of the construction process, we implemented a rigorous quality assurance protocol. Unlike traditional datasets that rely on labor-intensive manual labeling, we leverage the zero-shot reasoning capabilities of \texttt{Qwen2.5-VL-32B-Instruct} to establish an automated pipeline, which we then statistically validate through a detailed human audit.

\paragraph{Verification Protocol.} 
We randomly sampled 100 videos from our dataset for a formal verification. Independent annotators evaluated each sample in terms of factual correctness (i.e., the accuracy of the scene description relative to the visual content), counterfactual plausibility (i.e., the logical and physical soundness of the hypothetical scenarios), and decision appropriateness (i.e., the safety and optimality of the predicted driving actions).

\paragraph{Quantitative Validation.} 
The automated pipeline achieved an 86.0\% usability rate. The remaining 14.0\% exhibited major errors or hallucinations. While we acknowledge these imperfections, an 86\% reliability rate in complex, multi-modal reasoning tasks is highly competitive and justifies the use of raw model-generated data for large-scale training and benchmarking \cite{dataset:lingoqa}.

\begin{table*}[h]
\centering
\begin{tabular}{lcc}
\toprule
\textbf{Category} & \textbf{Count} & \textbf{Percentage (\%)} \\
\midrule
Correct & 77 & 77.0\% \\
Minor Errors (e.g., phrasing) & 9 & 9.0\% \\
Major Errors (e.g., logic flaws) & 8 & 8.0\% \\
Hallucinations (e.g., phantom objects) & 6 & 6.0\% \\
\midrule
\textbf{Usable (Correct + Minor)} & \textbf{86} & \textbf{86.0\%} \\
\bottomrule
\end{tabular}
\caption{Human audit results ($N=100$). Samples categorized as 'Correct' or 'Minor Error' are considered usable for robust training and evaluation.}
\label{tab:human_eval_results}
\end{table*}

\begin{table*}[h]
\centering
\small
\resizebox{\textwidth}{!}{
\begin{tabular}{l c | cc cc ccc}
\toprule
\multirow{2}{*}{\textbf{Model}} & \multirow{2}{*}{\textbf{Metric}} & \multicolumn{2}{c}{\textbf{Accident Recog.}} & \multicolumn{2}{c}{\textbf{Ego Inclusion}} & \multicolumn{3}{c}{\textbf{Decision}} \\
 & & Yes & No & Involv. & Not In. & Stop & Decel. & Keep \\
\midrule

\multirow{3}{*}{\textbf{InternVL2-1B}}
 & P@1 w/o & 69.4 & 58.0 & 58.0 & 69.4 & 33.3 & 32.1 & 71.4 \\
 & P@1 w/  & 66.7 & 30.0 & 66.7 & 30.0 & 22.2 & 32.1 & 59.2 \\
 & \cellcolor{gray!10} $\Delta$ & -2.7 & -28.0 & +8.7 & -39.4 & -11.1 & 0.0 & -12.2 \\
\midrule

\multirow{3}{*}{\textbf{InternVL2-2B}}
 & P@1 w/o & 41.7 & 66.6 & 41.7 & 66.6 & 11.1 & 25.0 & 83.7 \\
 & P@1 w/  & 50.0 & 56.0 & 50.0 & 56.0 & 22.2 & 32.1 & 83.7 \\
 & \cellcolor{gray!10} $\Delta$ & +8.3 & -10.6 & +8.3 & -10.6 & +11.1 & +7.1 & 0.0 \\
\midrule

\multirow{3}{*}{\textbf{InternVL2.5-1B}}
 & P@1 w/o & 44.4 & 60.0 & 44.4 & 60.0 & 22.2 & 28.6 & 69.4 \\
 & P@1 w/  & 52.8 & 60.0 & 52.8 & 60.0 & 44.4 & 25.0 & 73.5 \\
 & \cellcolor{gray!10} $\Delta$ & +8.4 & 0.0 & +8.4 & 0.0 & +22.2 & -3.6 & +4.1 \\
\midrule

\multirow{3}{*}{\textbf{Qwen2.5-VL-3B}}
 & P@1 w/o & 63.9 & 42.0 & 55.6 & 44.0 & 33.3 & 46.4 & 55.1 \\
 & P@1 w/  & 75.0 & 38.0 & 75.0 & 38.0 & 55.6 & 21.4 & 44.9 \\
 & \cellcolor{gray!10} $\Delta$ & +11.1 & -4.0 & +19.4 & -6.0 & +22.3 & -25.0 & -10.2 \\
\midrule

\multirow{3}{*}{\textbf{VideoLLaMA2-7B}}
 & P@1 w/o & 47.2 & 68.0 & 47.2 & 68.0 & 11.1 & 17.9 & 77.6 \\
 & P@1 w/  & 77.8 & 34.0 & 75.0 & 34.0 & 33.3 & 28.6 & 46.9 \\
 & \cellcolor{gray!10} $\Delta$ & +30.6 & -34.0 & +27.8 & -34.0 & +22.2 & +10.7 & -30.7 \\
\midrule

\end{tabular}
}
\caption{Pass@1 comparison w/ and w/o counterfactual in SFT for random samples.}
\label{tab:passk_sft_quality_validation}
\end{table*}

To verify robustness to annotation noise, we re-evaluated on the 86 high-quality samples confirmed through human audit. The Pass@1 results demonstrate identical directional trends to those observed in the full dataset, confirming that the 14\% noise does not systematically bias our conclusions.

\clearpage
\subsection{Statistical Analysis of the Dataset}

\begin{figure}[h]
    \centering
    \includegraphics[width=\columnwidth]{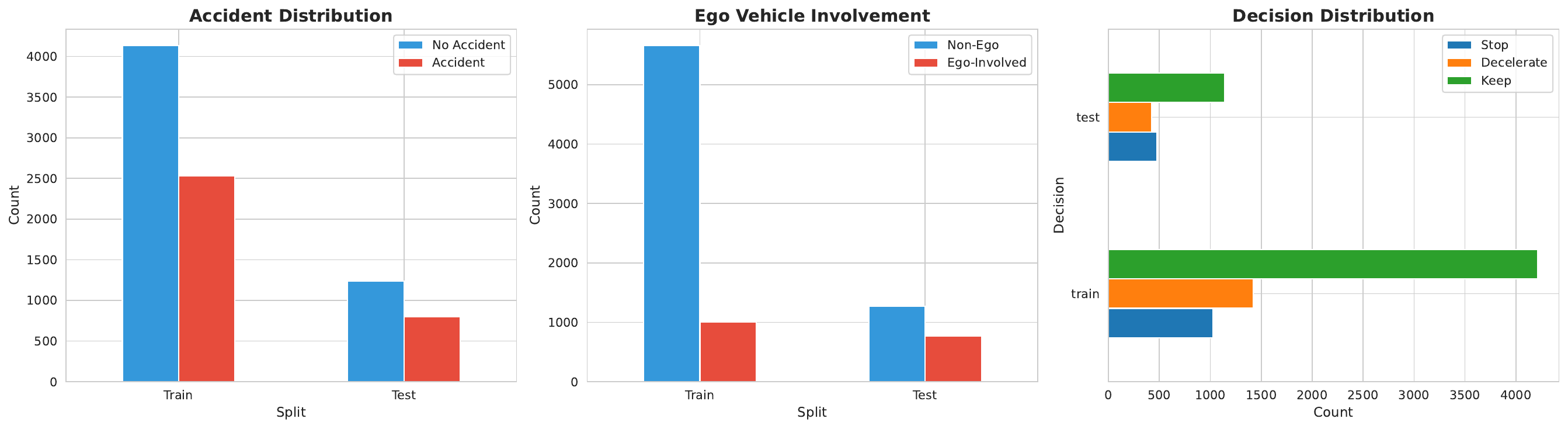}
    \caption{\textbf{Overview of dataset statistics and label distributions.} The plots illustrate the imbalance in accident occurrence (top-left) and ego-vehicle involvement (bottom-left). The horizontal bar chart (right) compares the distribution of driving decisions (stop, decelerate, keep) across the training (blue) and testing (red) splits, demonstrating a consistent class distribution between the two sets.}
    \label{fig:dataset_static_fig1_clss}
\end{figure}

\begin{figure}[h]
    \centering
    \includegraphics[width=\columnwidth]{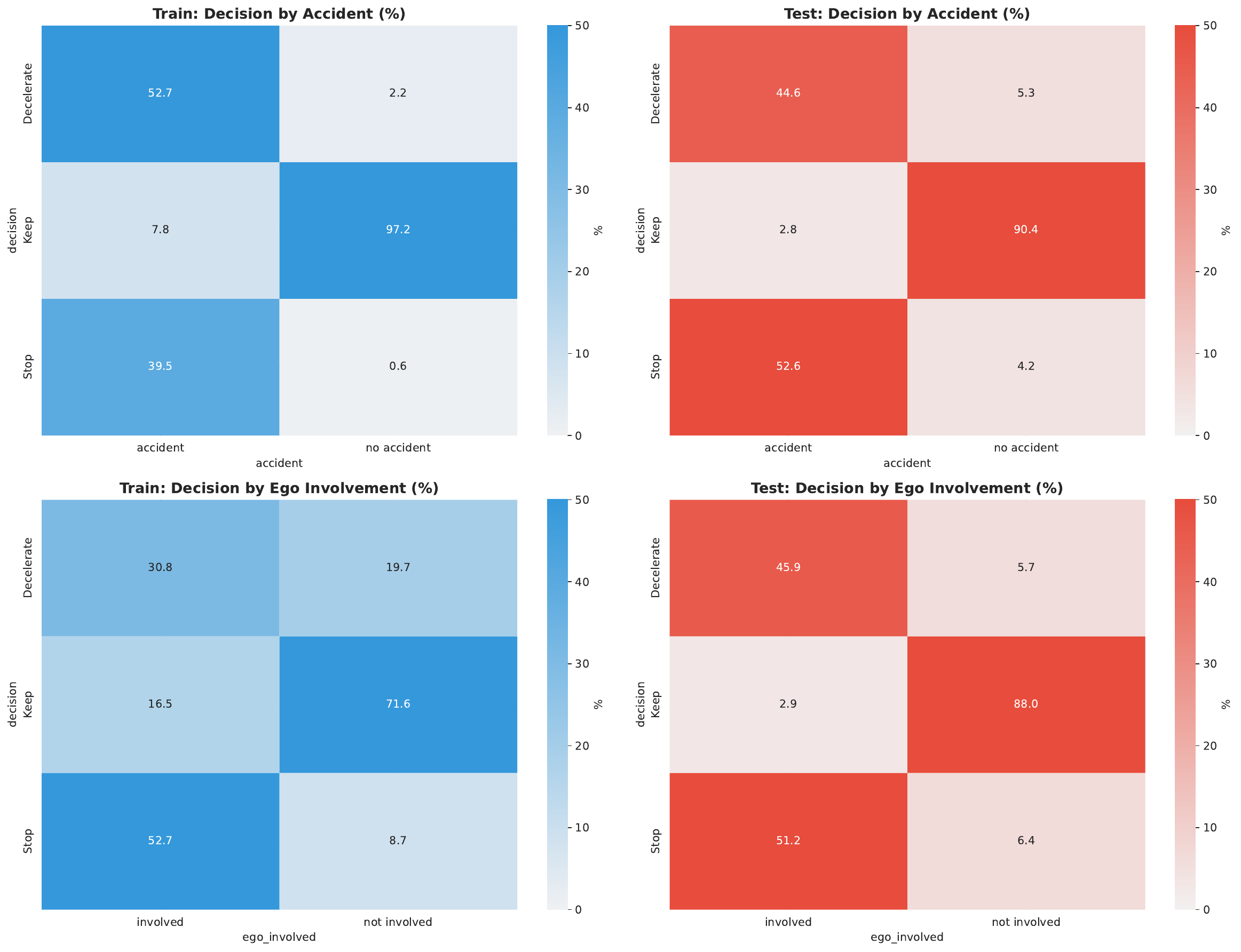}
    \caption{\textbf{Conditional distribution of driving decisions with respect to accident occurrence and ego-vehicle involvement.} The heatmaps display column-normalized percentages to analyze the correlation between decisions and scenario attributes. The top row shows the relationship between decisions and accidents, while the bottom row depicts ego-involvement. Left columns represent the training set (blue scale), and right columns represent the test set (red scale).}
    \label{fig:dataset_static_fig2_cmap}
\end{figure}

\begin{figure}[h]
    \centering
    \includegraphics[width=\columnwidth]{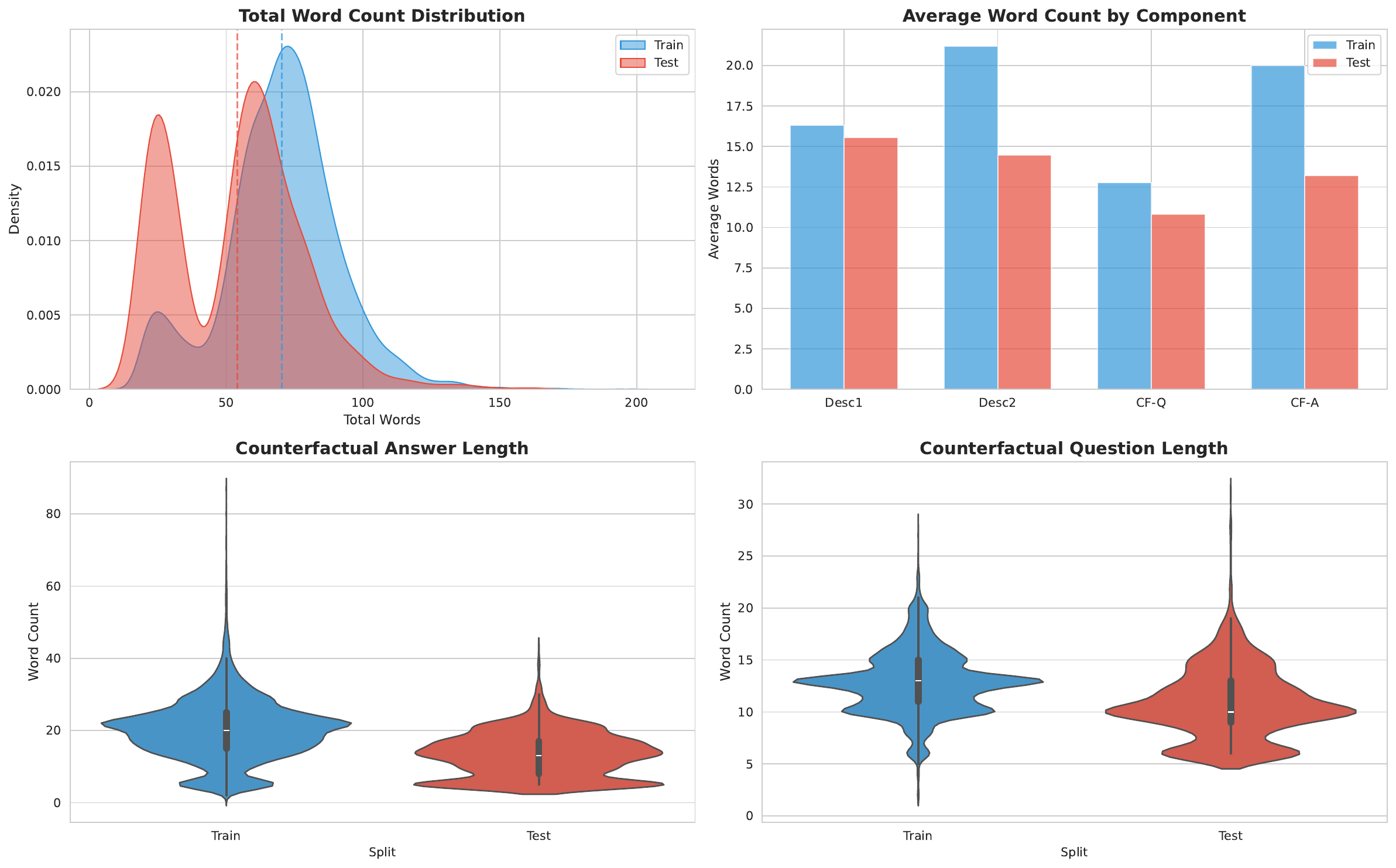}
    \caption{\textbf{Analysis of linguistic characteristics and token distributions.} (top-left) Kernel Density Estimation (KDE) of total word counts showing alignment between training and testing splits. (top-right) Average word counts broken down by descriptions, counterfactual Q\&A. (bottom) Violin plots illustrating the distribution of word counts for counterfactual answers (bottom-left) and questions (bottom-right). Blue indicates the training set, and red indicates the test set.}
    \label{fig:dataset_static_fig3_word.dist}
\end{figure}

\begin{figure}[h]
    \centering
    \includegraphics[width=\columnwidth]{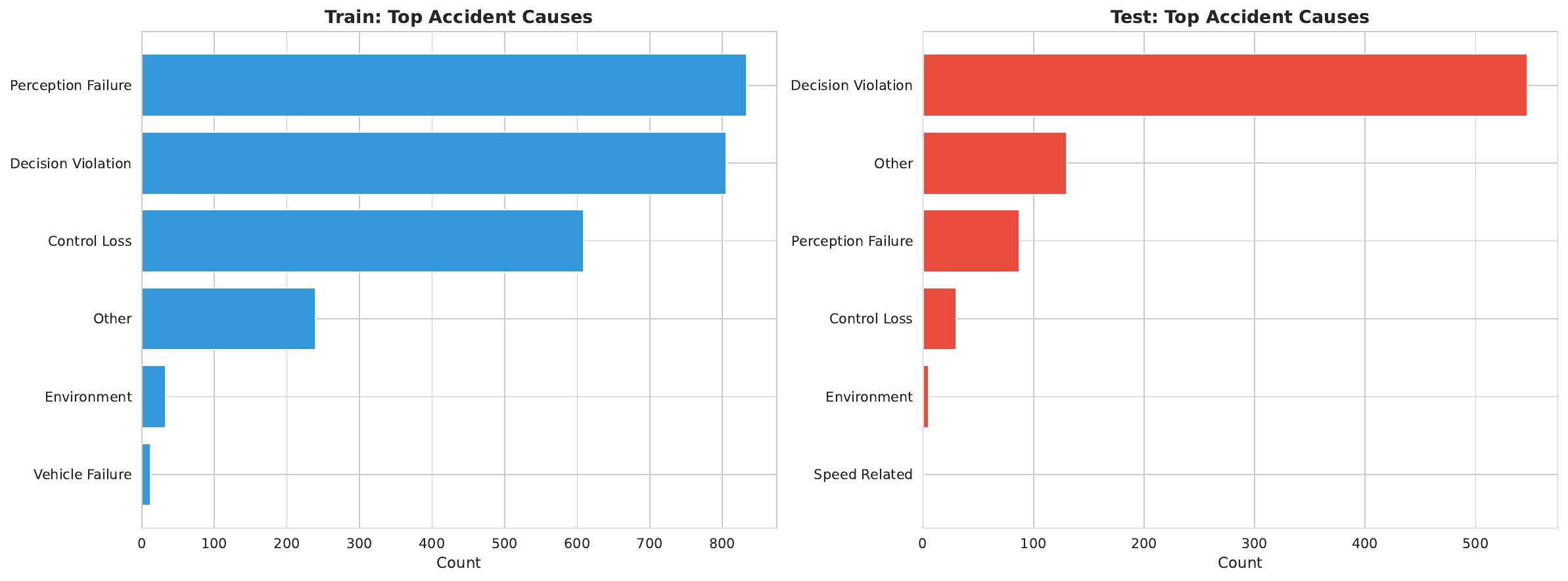}
    \caption{\textbf{Distribution of primary accident causes.} The horizontal bar charts display the frequency of the top annotated accident causes (e.g., control loss, perception failure) for the training (left, blue) and testing (right, red) sets. The semantic distribution of accident scenarios is preserved across the splits.}
    \label{fig:dataset_static_fig4_scene}
\end{figure}

\clearpage 
\section{Prompt Templates}
\label{appendix:prompts}
\begin{tcolorbox}[title=description and counterfactual Q\&A generation for normal video, breakable]
\small
\textbf{System prompt:}
\begin{verbatim}

You are given a video of driving scene with no accident. 
For this video, **there is no accident**. 
Describe the following information in natural language sentences 
(each item should be 10-20 words):

1. Accident occurrence (always 0)
2. Whether the ego-car is involved (always 0)
3. Scene description 
(describe what is actually happening in the video)
4. Counterfactual question 
(what-if scenario)
5. Counterfactual answer 
(what would happen in the what-if scenario)

Use complete sentences to report all six items.

Video Description:
<scene summary or keyframe captions here>
Ego-car involvement: <yes/no>

Strictly follow this output format: 
1, 1, <accident>...</accident>, <accident>...</accident>, 
<no_accident>...</no_accident>, <no_accident>...</no_accident>

For counterfactual question and answer, 
create diverse what-if scenarios beyond vehicle swerving. 
Consider weather, pedestrians, traffic signals, road conditions, 
or mechanical failures.

Now, generate the annotation for the given video 
as a single comma-separated line. 
Do not add any extra text.
\end{verbatim}
\end{tcolorbox}

\begin{tcolorbox}[title=description and counterfactual Q\&A generation  for accident video, breakable]
\small
\textbf{System prompt:}
\begin{verbatim}
You are given a video of an accident. 
For this video, **an accident occurs or harzard scene exists**. 
Describe the following information in natural language sentences 
(each item should be 10-20 words):

1. Accident occurrence (always 1)
2. Whether the ego-car is involved (0 for no, 1 for yes)
3. Accident result
4. Accident cause
5. Counterfactual question (what-if scenario)
6. Counterfactual answer 
(what would happen if no accident occurred)

Use complete sentences to report all six items.

Video Description:
<scene summary or keyframe captions here>
Ego-car involvement: <yes/no>

Strictly follow this output format:
Output: 1, (0 for no, 1 for yes), 
<accident>...</accident>, <accident>...</accident>, 
<no_accident>...</no_accident>, <no_accident>...</no_accident>

Now, generate the annotation for the given video 
as a single comma-separated line. 
Do not add any extra text.
\end{verbatim}
\end{tcolorbox}

\begin{tcolorbox}[title=decision generation, breakable]
\small
\textbf{System prompt:}
\begin{verbatim}
You are an expert driving policy annotator. 
You are given a video and its captions describing a driving scene. 
The video may show either normal driving or the moments just 
before a potential accident. 

The caption always follows this format:

1. Accident occurrence (0 for no accident, 1 for accident)
2. Whether the ego-car is involved (0 for no, 1 for yes)
3. Accident result
4. Accident cause
5. Counterfactual question (what-if scenario)
6. Counterfactual answer 
(what would happen if no accident occurred)

Examples:

# Accident occurred
Accident occurrence: 1
Ego involved: 0 or 1
Accident result: 
    a car in front of ego-car collides with a pedestrian
Accident cause: 
    the pedestrian violates traffic signal
Counterfactual question: 
    what if the pedestrian does not violate the signal?
Counterfactual answer: 
    there would be no collision

# No accident
Accident occurrence: 0
Ego involved: 0
Accident result: 
    no accident
Accident cause: 
    the car in front of ego-car turns right
Counterfactual question: 
    what if the car suddenly stopped?
Counterfactual answer: 
    there would be a collision

---

Caption for the given video:
{data['conversations'][1]['value']}

Your task:
Return the **driving decision** as a single action class.  
The possible action classes are:

- STOP
- DECELERATE
- KEEP

Rules:
1. If an accident occurs and the ego-car is involved: 
return the action the ego-car should have taken before the accident 
2. If an accident occurs but the ego-car is not involved: 
return the safe action to avoid the accident 
3. If no accident occurs or no action is required: 
return <decision>KEEP</decision>.
4. You may refer to the counterfactual question and answer 
if it helps decide the correct action.

Output format:
<decision>ACTION</decision>

Return only a single line in this format. 
Do not add any explanation or extra text.
\end{verbatim}
\end{tcolorbox}


\begin{tcolorbox}[title=Fine-Tuning and Test-Time Prompts for model w/o CF reasoning, breakable]
\small
\textbf{System prompt:}
\begin{verbatim}


You are given a video of a driving scene. 
The video may show normal driving 
or the moments just before a potential accident. 

Describe the following information in natural language sentences 
(each item should be 10-20 words):
1. Accident occurrence (0 for no accident, 1 for accident)
2. Whether the ego-car is involved (0 for no, 1 for yes)
3. Accident result
4. Accident cause
5. Decision for given scene

Use complete sentences to report all six items.

Video Description:
<scene summary or keyframe captions here>

Examples:

# Accident occurred
Accident occurrence: 1
Ego involved: 0 or 1
Accident result: 
    a car in front of ego-car collides with a pedestrian
Accident cause: 
    the pedestrian violates traffic signal

# No accident
Accident occurrence: 0
Ego involved: 0
Accident result: 
    no accident
Accident cause: 
    the car in front of ego-car turns right
Now, describe the following video in the same sentence-style format:
<scene summary or keyframe captions of your target video>

Format of Accident occurred:
Output: 1, (0 for no, 1 for yes), 
<accident>Accident result</accident>, 
<accident>Accident cause</accident>, 
<decision>DECISION</decision>

Format of No accident:
Output: 0, 0, 
<no_accident>Accident result</no_accident>, 
<no_accident>Accident cause</no_accident>, 
<decision>DECISION</decision>

Now, generate the response for the given video 
as a **single comma-separated line**. 
Do not add any extra text.
\end{verbatim}
\end{tcolorbox}

\begin{tcolorbox}[title=Fine-Tuning and Test-Time Prompts for model w/ CF reasoning, breakable]
\small
\textbf{System prompt:}
\begin{verbatim}


You are given a video of a driving scene. 
The video may show normal driving 
or the moments just before a potential accident. 

Describe the following information in natural language sentences 
(each item should be 10-20 words):
1. Accident occurrence (0 for no accident, 1 for accident)
2. Whether the ego-car is involved (0 for no, 1 for yes)
3. Accident result
4. Accident cause
5. Counterfactual question 
(what-if scenario)
6. Counterfactual answer 
(what would happen if no accident occurred)
7. Decision for given scene

Use complete sentences to report all six items.

Video Description:
<scene summary or keyframe captions here>

Examples:

# Accident occurred
Accident occurrence: 1
Ego involved: 0 or 1
Accident result: 
    a car in front of ego-car collides with a pedestrian
Accident cause: 
    the pedestrian violates traffic signal
Counterfactual question: 
    what if the pedestrian does not violate the signal?
Counterfactual answer: 
    there would be no collision

# No accident
Accident occurrence: 0
Ego involved: 0
Accident result: 
    no accident
Accident cause: 
    the car in front of ego-car turns right
Counterfactual question: 
    What if the car suddenly stopped?
Counterfactual answer: 
    there would be a collision
Now, describe the following video in the same sentence-style format:
<scene summary or keyframe captions of your target video>

Format of Accident occurred:
Output: 1, (0 for no, 1 for yes), 
<accident>Accident result</accident>, 
<accident>Accident cause</accident>, 
<no_accident>Counterfactual question</no_accident>, 
<no_accident>Counterfactual Answer</no_accident>
<decision>DECISION</decision>

Format of No accident:
Output: 0, 0, 
<no_accident>Accident result</no_accident>, 
<no_accident>Accident cause</no_accident>, 
<accident>Counterfactual Question</accident>, 
<accident>Counterfactual Answer</accident>
<decision>DECISION</decision>

Now, generate the response for the given video 
as a **single comma-separated line**. 
Do not add any extra text.
\end{verbatim}
\end{tcolorbox}

\begin{tcolorbox}[title=Token Activation Map Analysis Prompts, breakable]
\small
\textbf{System prompt:}
\begin{verbatim}
# perceptual prompt
What is the most critical object the ego-car should be cautious of 
in this driving scene? 

# decision prompt 
Based on the current visual scene, what is the immediate action 
required for the ego-car?

# counterfactual prompt
(counterfactual question on the main factor based on the GT scene description)
(e.g.) What if the car suddenly stopped?
\end{verbatim}
\end{tcolorbox}

\clearpage
\section{Reward Functions}
\label{appendix:reward_functions}
To align the model's outputs with safety-critical driving standards, we design a comprehensive reward function $R$ composed of six sub-rewards. Let $y$ denote the generated completion and $y^*$ denote the ground truth. The total reward is defined as a weighted sum:

\begin{equation}
    R = w_{\text{acc}} R_{\text{acc}} + w_{\text{ego}} R_{\text{ego}} + w_{\text{dec}} R_{\text{dec}} + w_{\text{safe}} R_{\text{safe}} + w_{\text{corr}} R_{\text{corr}} + w_{\text{fmt}} R_{\text{fmt}}
\label{eq:total_reward}
\end{equation}

\noindent where the weights are set to $w_{\text{acc}}=0.1$, $w_{\text{ego}}=0.1$, $w_{\text{dec}}=0.4$, $w_{\text{safe}}=0.2$, $w_{\text{corr}}=0.1$, and $w_{\text{fmt}}=0.1$, prioritizing the correct driving decision and safety.

\paragraph{Recognition Rewards ($R_{\text{acc}}, R_{\text{ego}}$)}
We verify the fundamental perception capabilities: accident occurrence ($y_{\text{acc}}$) and ego-vehicle involvement ($y_{\text{ego}}$). These rewards utilize the indicator function $\mathbb{I}(\cdot)$, which equals 1 if the condition is met and 0 otherwise:
\begin{equation}
    R_{\text{acc}} = \mathbb{I}(y_{\text{acc}} = y^*_{\text{acc}}), \quad R_{\text{ego}} = \mathbb{I}(y_{\text{ego}} = y^*_{\text{ego}})
\end{equation}

\paragraph{Decision Reward ($R_{\text{dec}}$)}
The core driving decision (e.g., decelerate, stop) is extracted from the \texttt{<decision>} tags.
\begin{equation}
    R_{\text{dec}} = \mathbb{I}(y_{\text{dec}} = y^*_{\text{dec}})
\end{equation}

\paragraph{Safety Critical Reward ($R_{\text{safe}}$)}
To penalize dangerous failures, we introduce a specialized reward for ``stop'' scenarios. If the ground truth requires a stop ($y^*_{\text{dec}} = \text{stop}$), the model receives a penalty for missing it:
\begin{equation}
    R_{\text{safe}} = 
    \begin{cases} 
    1.0 & \text{if } y^*_{\text{dec}} = \text{stop} \land y_{\text{dec}} = \text{stop} \\
    -0.5 & \text{if } y^*_{\text{dec}} = \text{stop} \land y_{\text{dec}} \neq \text{stop} \\
    0.0 & \text{otherwise}
    \end{cases}
\end{equation}

\paragraph{Consistency Reward ($R_{\text{corr}}$)}
To ensure logical consistency between the classification flag and the textual description, $R_{\text{corr}}$ verifies whether the predicted accident flag $y_{\text{acc}}$ matches the ground truth and aligns with the corresponding XML tags. Let $T(y)$ be the set of tags generated in the description. The reward is defined as:

\begin{equation}
    R_{\text{corr}} = \mathbb{I}(y_{\text{acc}} = y^*_{\text{acc}}) \cdot \mathbb{I}\left( 
    \begin{aligned} 
    &(y_{\text{acc}} = 1 \land \texttt{<accident>} \in T(y)) \lor \\ 
    &(y_{\text{acc}} = 0 \land \texttt{<no\_accident>} \in T(y)) 
    \end{aligned} 
    \right)
\end{equation}

\paragraph{Format Rewards ($R_{\text{fmt}}$)} 
This reward ensures structural validity by verifying the number of parsed segments $|S_y|$. The expected length differs depending on the reasoning mode. For models utilizing counterfactual reasoning, we require extended segments for the traces:

\begin{equation}
    R_{\text{fmt}} = 
    \begin{cases} 
    \mathbb{I}(|S_y| = 7) & \text{if w/ CF reasoning)} \\
    \mathbb{I}(|S_y| = 5) & \text{otherwise}
    \end{cases}
\end{equation}

The $R_{\text{safe}}$ assigns additional weight to emergency stop predictions, which may amplify the safety prior observed in SFT. However, since both w/ CF and w/o CF reference models are trained under identical reward structures, the observed difference in RFT action distribution shift is attributable to the CF reasoning effect in the reference model rather than the reward design alone. The persistence of the SFT-RFT gap suggests that CF reasoning establishes a prior that interacts with, rather than is replaced by, the reward signal.

\paragraph{Disentangling CF Reasoning from Reward Design.} To verify whether the observed action distribution shift is attributable to the reward design rather than CF reasoning itself, we conduct an ablation study by removing the safety-critical reward $R_{\text{safe}}$ from the reward function. We select Qwen2.5-VL-3B (GRPO) as the target model, as it exhibited the strongest CF-induced bias in Table~\ref{tab:passk_rft}. To compensate for the removed $R_{\text{safe}}$, we increase the decision reward weight from $w_{\text{dec}}=0.4$ to $w_{\text{dec}}=0.6$, ensuring that the total reward scale remains comparable to the original setting.

\begin{figure}[h]
    \centering
    \includegraphics[width=\columnwidth]{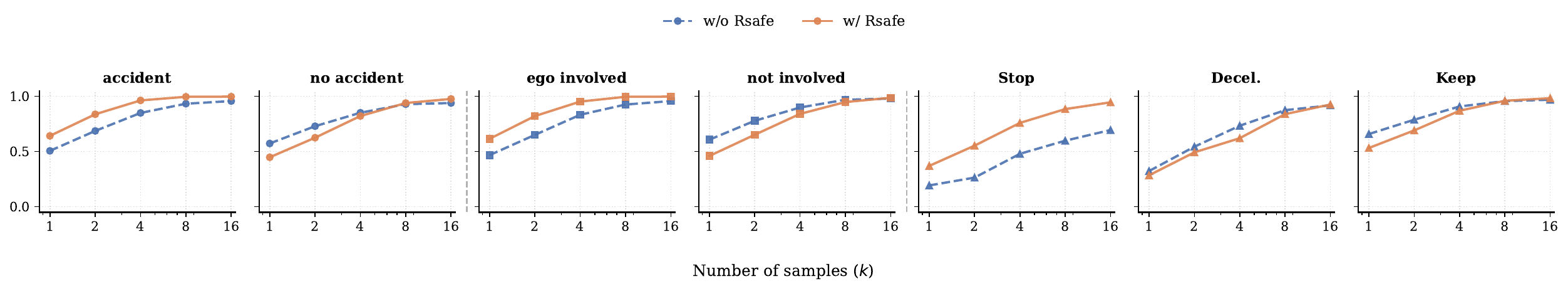}
    \caption{Class-wise pass@$k$ performance across model scales and CF reasoning in ablation on $R_{\text{safe}}$.}
    \label{fig:passk_Rsafe}
\end{figure}

\begin{table*}[h]
\centering
\begin{tabular}{c | cc cc ccc}
\toprule
\multirow{2}{*}{Metric} & \multicolumn{2}{c}{\textbf{Accident Recog.}} & \multicolumn{2}{c}{\textbf{Ego Inclusion}} & \multicolumn{3}{c}{\textbf{Decision}} \\
 & Yes & No & Involv. & Not In. & Stop & Decel. & Keep \\
\midrule
P@1 w/o CF & 50.6 & 57.3 & 46.7 & 60.7 & 19.3 & 32.3 & 65.8 \\
P@1 w/ CF  & 64.2 & 44.8 & 61.5 & 46.0 & 37.0 & 28.5 & 53.2 \\
\cellcolor{gray!10} $\Delta$ & +13.6 & -12.5 & +14.8 & -14.7 & +17.7 & -3.8 & -12.6 \\
\cmidrule{1-8}
P@16 w/o CF & 95.7 & 93.9 & 95.7 & 97.9 & 69.5 & 91.8 & 96.9 \\
P@16 w/ CF  & 99.7 & 97.6 & 99.7 & 98.7 & 94.5 & 92.6 & 98.3 \\
\cellcolor{gray!10} $\Delta$ & +4.0 & +3.7 & +4.0 & +0.8 & +25.0 & +0.8 & +1.4 \\
\bottomrule
\end{tabular}
\caption{
Effect of counterfactual (CF) reasoning \textbf{without $R_{\text{safe}}$} across different $k$ on Qwen2.5-VL-3B (GRPO). While the magnitude of the shift is reduced compared to the full setting, consistent improvements in accident-related and safety-critical decisions (e.g., Stop) persist across both P@1 and P@16, indicating that the effect is not limited to low-$k$ sampling.
}
\label{tab:rsafe_ablation}
\end{table*}

As shown in Table~\ref{tab:rsafe_ablation} and Figure ~\ref{fig:passk_Rsafe}, the distribution shift induced by CF reasoning persists even in the absence of $R_{\text{safe}}$: the Stop decision rate increases by $+17.7$\%p under the CF condition compared to the non-CF condition.
While the magnitude of the shift is reduced compared to the full reward setting ($+31.0$\%p), the direction remains consistent, indicating that $R_{\text{safe}}$ amplifies the bias but does not cause it. These results provide direct evidence that the action distribution shift originates from CF reasoning itself, rather than from the reward design.

\clearpage 
\section{Model Selection Criteria}
\label{appendix:model_selection_criteria}

To ensure the generality of our findings, we select backbone models according to three principles:
(1) prevalence in recent Vision-Language-Action (VLA) literature,
(2) architectural diversity in multimodal fusion strategies, and
(3) scalability across different parameter regimes under practical computational constraints.

\paragraph{Architectural Coverage.}
We adopt models from three representative multimodal families that are widely used in contemporary autonomous driving research:

\begin{itemize}
    \item \textbf{InternVL2-1B, InternVL2-2B and InternVL2.5-1B:}
    These models have recently emerged as strong open-source backbones in driving-related VLA systems. 
    Using multiple scales within the same architectural family enables controlled analysis of parameter scaling effects.
    
    \item \textbf{Qwen2.5-VL-3B:}
    A widely adopted backbone in recent large-scale VLA frameworks, representing a distinct multimodal alignment strategy compared to InternVL.
    
    \item \textbf{VideoLLaMA2-7B:}
    A temporally-aware video-language model, included to evaluate whether the observed trends persist in video-native architectures with larger parameter scales.
\end{itemize}

Collectively, our selection spans diverse multimodal fusion designs and parameter regimes ranging from 1B to 7B parameters, mitigating the risk that the observed effects are backbone-specific artifacts.

\begin{table*}[h]
    \centering
    \footnotesize 
    \renewcommand{\arraystretch}{1.3} 
    \setlength{\tabcolsep}{3pt} 
    
    \begin{tabularx}{\textwidth}{ 
        >{\raggedright\arraybackslash}p{2.2cm} 
        >{\raggedright\arraybackslash}p{4.5cm} 
        >{\raggedright\arraybackslash}X 
    }
    \toprule
    \textbf{Category} & \textbf{Rationale for Selection} & \textbf{Related Works} \\
    \midrule

    \textbf{LLaMA \& Mistral} 
    & Standard baseline in early VLA research. We adopt \textbf{VideoLLaMA2-7B} to represent this lineage.
    & DriveMLM \cite{e2e-vlm:drivemlm}, RDA-Driver \cite{e2e-vlm:rda-driver}, DriveLM \cite{e2e-vlm:drivelm}, DriveGPT4 \cite{e2e-vlm:drivegpt4}, VLAAD \cite{e2e-vlm:vlaad}, OccLLaMA \cite{e2e-vlm:occllama}, OmniDrive \cite{causal:e2e-vlm:omnidrive}, LMDrive \cite{e2e-vlm:lmdrive}, BEVDriver \cite{e2e-vlm:bevdriver}, CoVLA-Agent \cite{e2e-vlm:covla-agent}, DSDrive \cite{e2e-vlm:dsdrive} \\

    \midrule

    \textbf{Qwen Family} 
    & Dominant in 2025 research. We adopt \textbf{Qwen2.5-VL-3B} to align with SOTA trends.
    & ImpromptuVLA \cite{e2e-vlm:impromptuvla}, DriveAgent-R1 \cite{e2e-vlm:driveagent-r1}, FastDriveVLA \cite{e2e-vlm:fastdrivevla}, AutoDrive-R$^2$ \cite{e2e-vlm:autodrive-r2}, CoReVLA \cite{e2e-vlm:corevla}, AutoVLA \cite{e2e-vlm:autovla}, E3AD \cite{e2e-vlm:e3ad}, UniUGP \cite{e2e-vlm:uniugp}, Reasoning-VLA \cite{e2e-vlm:reasoning-vla} \\

    \midrule

    \textbf{InternVL Family} 
    & Vision-Centric alternative. We adopt \textbf{InternVL2} to verify scalability.
    & Drive-R1 \cite{e2e-vlm:drive-r1}, SimLingo \cite{causal:e2e-vlm:simlingo}, VDRive \cite{e2e-vlm:vdrive}, DriVLM \cite{e2e-vlm:drivlm}, Li et al. \cite{e2e-vlm:driving_with_internvl_challange} \\

    \bottomrule
    \end{tabularx}
    \caption{Summary of VLA Models and Related Works. We categorize models based on their architectural lineage.}
    \label{tab:backbone_summary}
\end{table*}

\paragraph{Use of Public Resources.}
We acknowledge the use of the following publicly available resources in this work. :

\begin{itemize}

    \item InternVL2-1B \footnote{\url{https://huggingface.co/OpenGVLab/InternVL2-1B}} \dotfill Apache License 2.0

    \item InternVL2-2B \footnote{\url{https://huggingface.co/OpenGVLab/InternVL2-2B}} \dotfill Apache License 2.0

    \item InternVL2.5-1B \footnote{\url{https://huggingface.co/OpenGVLab/InternVL2_5-1B}} \dotfill Apache License 2.0

    \item Qwen2.5-VL-3B \footnote{\url{https://huggingface.co/Qwen/Qwen2.5-VL-3B-Instruct}} \dotfill Apache License 2.0

    \item VideoLLaMA2-7B \footnote{\url{https://huggingface.co/DAMO-NLP-SG/VideoLLaMA2-7B-Base}} \dotfill Apache License 2.0

\end{itemize}

\clearpage
\section{Training Details}
\label{appendix:hyperparams}

For supervised fine-tuning (SFT), we utilized four NVIDIA A5000 GPUs, whereas reinforcement fine-tuning (RFT) was conducted using a single NVIDIA L40S GPU.

\begin{table*}[h]
\centering
\begin{tabular}{llll}
\toprule
\textbf{Setting} & \textbf{Model} & \textbf{Hyperparameter} & \textbf{Value} \\
\midrule

\multirow{30}{*}{SFT}
& \multirow{6}{*}{InternVL2-1B}
& Batch size & 4 \\
& & Precision & \texttt{bf16} \\
& & Epoch & 5 \\
& & Learning rate & $1\times10^{-5}$ \\
& & Weight decay & 0.05 \\
& & \# of frames & 8 \\ 
\cmidrule(lr){2-4}

& \multirow{6}{*}{InternVL2-2B}
& Batch size & 4 \\
& & Precision & \texttt{bf16} \\
& & Epoch & 5 \\
& & Learning rate & $1\times10^{-5}$ \\
& & Weight decay & 0.05 \\
& & \# of frames & 8 \\ 
\cmidrule(lr){2-4}

& \multirow{6}{*}{InternVL2.5-1B}
& Batch size & 4 \\
& & Precision & \texttt{bf16} \\
& & Epoch & 5 \\
& & Learning rate & $1\times10^{-5}$ \\
& & Weight decay & 0.05 \\
& & \# of frames & 8 \\ 
\cmidrule(lr){2-4}

& \multirow{6}{*}{Qwen2.5-VL-3B}
& Batch size & 2 \\
& & Precision & \texttt{bf16} \\
& & Epoch & 5 \\
& & Learning rate & $1\times10^{-5}$ \\
& & Weight decay & 0.01 \\
& & \# of frames & 8 \\ 
\cmidrule(lr){2-4}

& \multirow{6}{*}{VideoLLaMA2-7B}
& Batch size & 2 \\
& & Precision & \texttt{bf16} \\
& & Epoch & 5 \\
& & Learning rate & $2\times10^{-5}$ \\
& & Weight decay & 0.01 \\
& & \# of frames & 8 \\ 

\midrule

\multirow{10}{*}{RFT}
& \multirow{5}{*}{InternVL2.5-1B (GRPO)}
& Batch size & 2 \\
& & Temperature & 0.9 \\ 
& & KL loss coefficient & $4\times10^{-2}$ \\
& & Group size & 2 \\
& & Learning rate & $1\times10^{-6}$ \\
\cmidrule(lr){2-4}

& \multirow{5}{*}{Qwen2.5-VL-3B (GRPO, GDPO)}
& Batch size & 2 \\
& & Temperature & 1.2 \\
& & KL loss coefficient & $4\times10^{-2}$ \\
& & Group size & 2 \\
& & Learning rate & $1\times10^{-6}$ \\

\bottomrule
\end{tabular}
\caption{Training hyperparameters for SFT and RFT.}
\label{tab:hyperparams_all}
\end{table*}

\clearpage
\section{Pseudo Codes}
\label{appendix:pseudo_codes}

\subsection{Policy Entropy}
\begin{lstlisting}[label={lst:policy_entropy}]
def calculate_policy_entropy(logits, completion_mask, temperature=1.0):
    """
    Computes the length-normalized policy entropy for generated completions.
    
    Args:
        logits (Tensor): Shape (Batch, Seq_Len, Vocab_Size)
        completion_mask (Tensor): Shape (Batch, Seq_Len). 1 for completion, 0 for prompt.
        temperature (float): Scaling factor for logits.
    
    Returns:
        float: Average entropy per token (invariant to sequence length).
    """
    
    # 1. Temperature Scaling
    # Adjusts the sharpness of the distribution.
    logits = logits / temperature

    # 2. Compute Entropy Per Token
    # H(t) = - sum(p(x) * log p(x))
    probs = softmax(logits, dim=-1)
    log_probs = log_softmax(logits, dim=-1)
    
    # Shape: (Batch, Seq_Len)
    per_token_entropy = -sum(probs * log_probs, dim=-1)

    # 3. Apply Completion Mask
    # Filter out prompt tokens (entropy = 0 for prompt positions)
    masked_entropy = per_token_entropy * completion_mask

    # 4. Length Normalization (Critical Step)
    # Calculate effective length of the generated completion
    sequence_length = sum(completion_mask, dim=-1)
    
    # Divide total entropy by sequence length
    # This ensures the metric reflects 'uncertainty density' rather than total uncertainty.
    policy_entropy = sum(masked_entropy, dim=-1) / clamp(sequence_length, min=1.0)

    return mean(policy_entropy) # Average over batch
\end{lstlisting}

\subsection{Perplexity (PPL)}
\begin{lstlisting}[label={lst:perplexity}]
import re
from math import exp

def extract_common_text(full_response: str) -> str:
    """
    Extracts only the shared text segments (Description & Decision) from the model output,
    filtering out Counterfactual (CF) Q&A sections to ensure fair comparison.
    """
    # 1. Extract all tagged sections (e.g., <accident>...</accident>)
    sections = re.findall(r'<[^>]+>.*?</[^>]+>', full_response)
    
    # 2. Extract prefix numbers (Accident Occurrence, Ego Involvement)
    prefix_match = re.match(r'^(\d+,\s*\d+)', full_response)
    prefix_numbers = prefix_match.group(1) if prefix_match else ""

    # 3. Filter common sections
    if len(sections) >= 5: 
        # Case: Model with CF reasoning (Output includes CF Q&A)
        # Format: [Desc1, Desc2, CF_Q, CF_A, Decision]
        # Action: Remove CF_Q (idx 2) and CF_A (idx 3)
        common_sections = [sections[0], sections[1], sections[-1]]
    else:
        # Case: Baseline Model (No CF Q&A)
        # Format: [Desc1, Desc2, Decision]
        # Action: Keep all sections
        common_sections = sections

    # 4. Reconstruct the common text
    return prefix_numbers + ", " + ", ".join(common_sections)


def compute_perplexity(model, video, prompt, target_text):
    """
    Computes the perplexity of the target_text given the video and prompt.
    """
    # 1. Construct full input sequence
    full_input = prompt + target_text
    
    # 2. Forward pass (Video + Text)
    # Inputs are tokenized and fed into the model to obtain logits
    logits = model(video, full_input)
    
    # 3. Set loss calculation range (Shifted Labels)
    # Exclude the prompt portion from loss computation; 
    # focus only on 'target_text'
    start_idx = len(tokenized_prompt)
    target_logits = logits[:, start_idx-1 : -1]
    target_labels = full_input_ids[:, start_idx:]
    
    # 4. Compute Cross Entropy Loss and Perplexity
    loss = CrossEntropyLoss(target_logits, target_labels)
    perplexity = exp(loss)
    
    return perplexity


# Main Evaluation Loop
def main(model_path, dataset_path):
    # Load model and dataset
    model = load_model(model_path)
    dataset = load_json(dataset_path)
    
    # Dictionary to store perplexity results
    ppl_results = {
        'overall': [],
        'stop': [],
        'decelerate': [],
        'keep': []
    }

    print("Starting Fair Perplexity Evaluation (Common Text Only)...")

    for sample in dataset:
        video_path = sample['video']
        # Original Ground Truth
        full_gt_text = sample['ground_truth']
        
        # Extract common text for fair comparison
        # Ensures that both CF and No-CF models 
        # are evaluated on identical text segments.
        common_target_text = extract_common_text(full_gt_text)
        
        # Extract decision class for breakdown analysis 
        # (Stop/Decelerate/Keep)
        decision_class = extract_decision_tag(full_gt_text)

        # Compute PPL based on the common text
        ppl = compute_perplexity(model, video_path, PROMPT, common_target_text)
        
        # Store results
        ppl_results['overall'].append(ppl)
        if decision_class in ppl_results:
            ppl_results[decision_class].append(ppl)

    # Compute and save final statistics
    print_statistics(ppl_results)
    save_results(ppl_results)

if __name__ == "__main__":
    main()
\end{lstlisting}

\clearpage
\section{Detailed Experimental Results}
\subsection{Supervised Fine-Tuning}
\label{appendix:passk_sft}

\begin{figure}[h]
    \centering
    \includegraphics[width=\columnwidth]{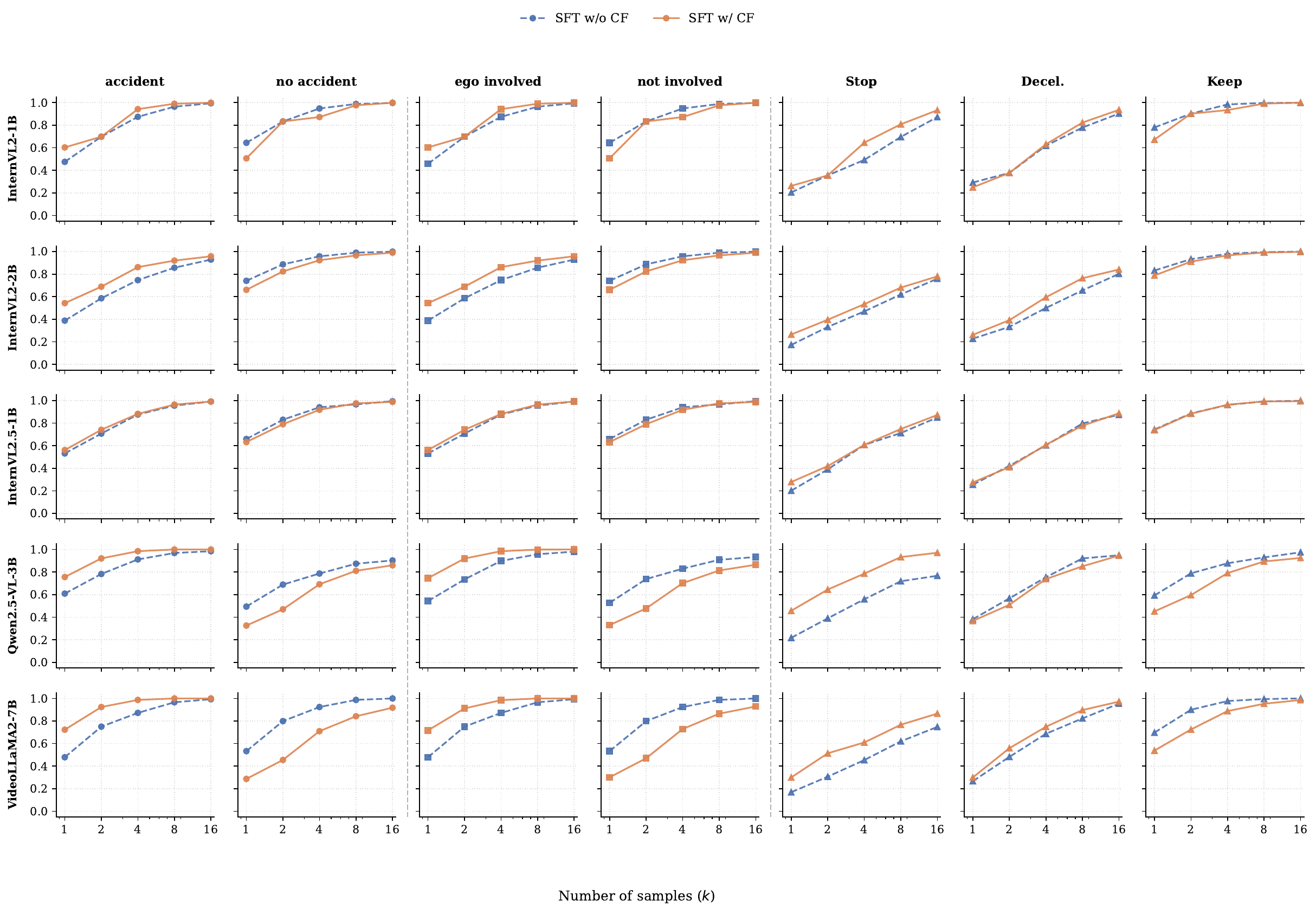}
    \caption{Class-wise pass@$k$ performance across model scales and CF reasoning settings in SFT.}
    \label{fig:passk_sft}
\end{figure}

\paragraph{Analysis of Accident Recognition and Ego Inclusion.} 
As illustrated in Fig. \ref{fig:passk_sft}, CF reasoning consistently enhances the pass@1 performance for accident recognition and ego inclusion across all model scales. In smaller models, such as \texttt{InternVL2-1B} and \texttt{2B}, the performance gap between models with and without CF reasoning progressively diminishes as $k$ approaches 16, eventually converging to a similar baseline. However, a distinct trade-off emerges as the model size increases. In larger models like \texttt{Qwen2.5-VL-3B} and \texttt{VideoLLaMA2-7B}, while CF reasoning significantly boosts performance on accident-related labels ("accident", "ego involved"), it simultaneously leads to a performance degradation on non-accident labels ("no accident", "not involved") compared to the baseline. This suggests that CF reasoning does not merely provide additional information but fundamentally recalibrates the model's perception, making it inherently more sensitive to safety-critical risk factors.

\paragraph{Persistence of Safety-bias in Decision Fields.} 
This trend is further pronounced in the decision field, revealing a clear disparity in "confidence levels" tied to model scale. For the \texttt{Stop}, while smaller models show converging performance curves as k increases, larger models—specifically \texttt{Qwen2.5-VL-3B} and \texttt{VideoLLaMA2-7B}—maintain a persistent $\Delta$pass@$k$ that does not dissipate even at $k$=16. Conversely, for the \texttt{Dec.} and \texttt{Keep}, the baseline models without CF reasoning exhibit superior performance in these larger scales. This indicates that larger models internalize accident-related cues learned via CF reasoning as a robust "world prior." Consequently, in large-scale VLM, CF reasoning induces a deep-seated action distribution shift that remains resistant to stochastic sampling, acting as a dominant mechanism that drives the model toward conservative decision-making.

\clearpage
\subsection{Reinforcement Fine-Tuning}
\label{appendix:passk_rft}
\begin{figure}[t]
    \centering
    \includegraphics[width=\columnwidth]{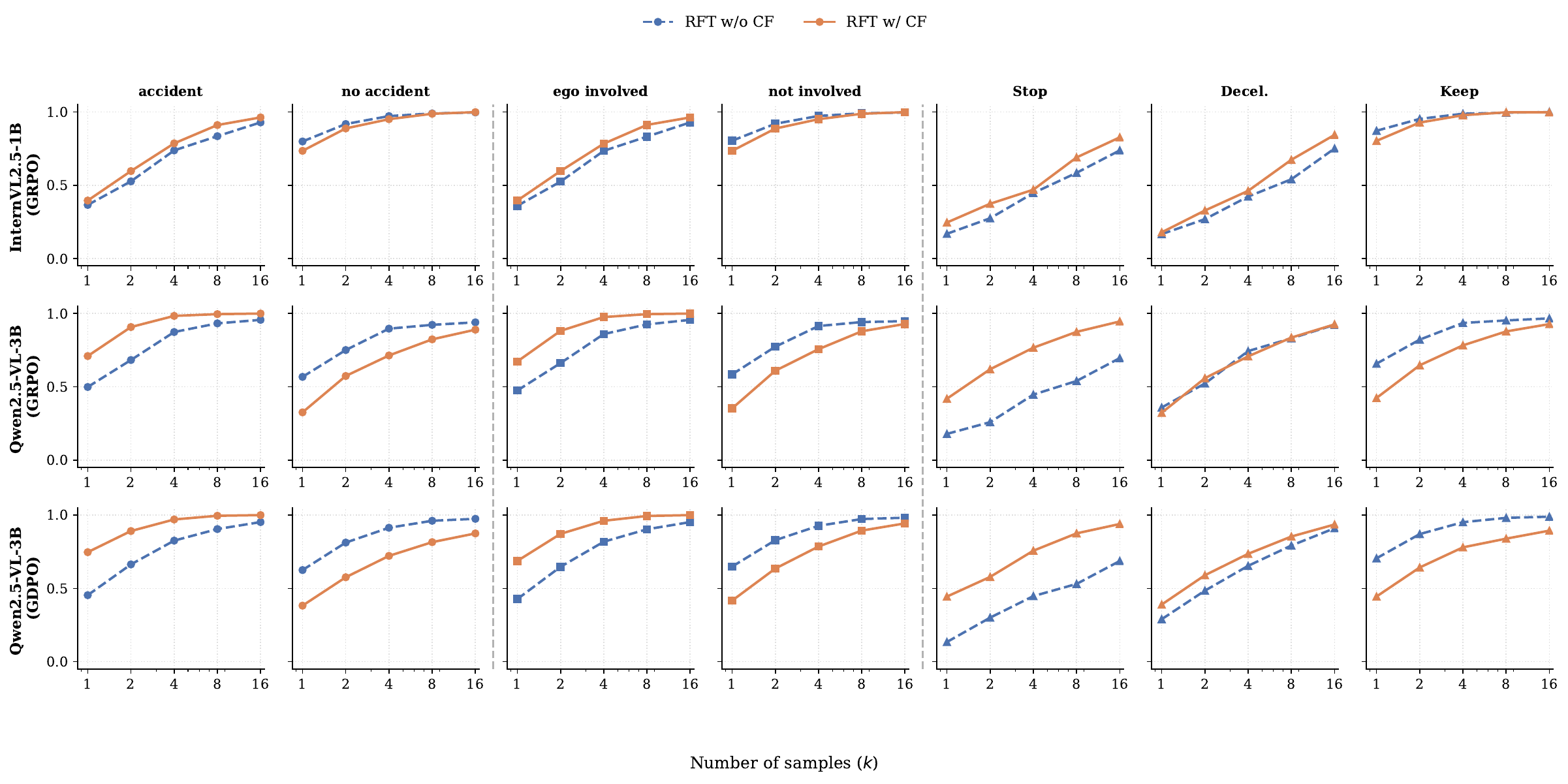}
    \caption{Class-wise pass@$k$ performance across model scales and CF reasoning settings in RFT.}
    \label{fig:passk_rft}
\end{figure}

\begin{figure}[h]
    \centering
    \includegraphics[width=\columnwidth]{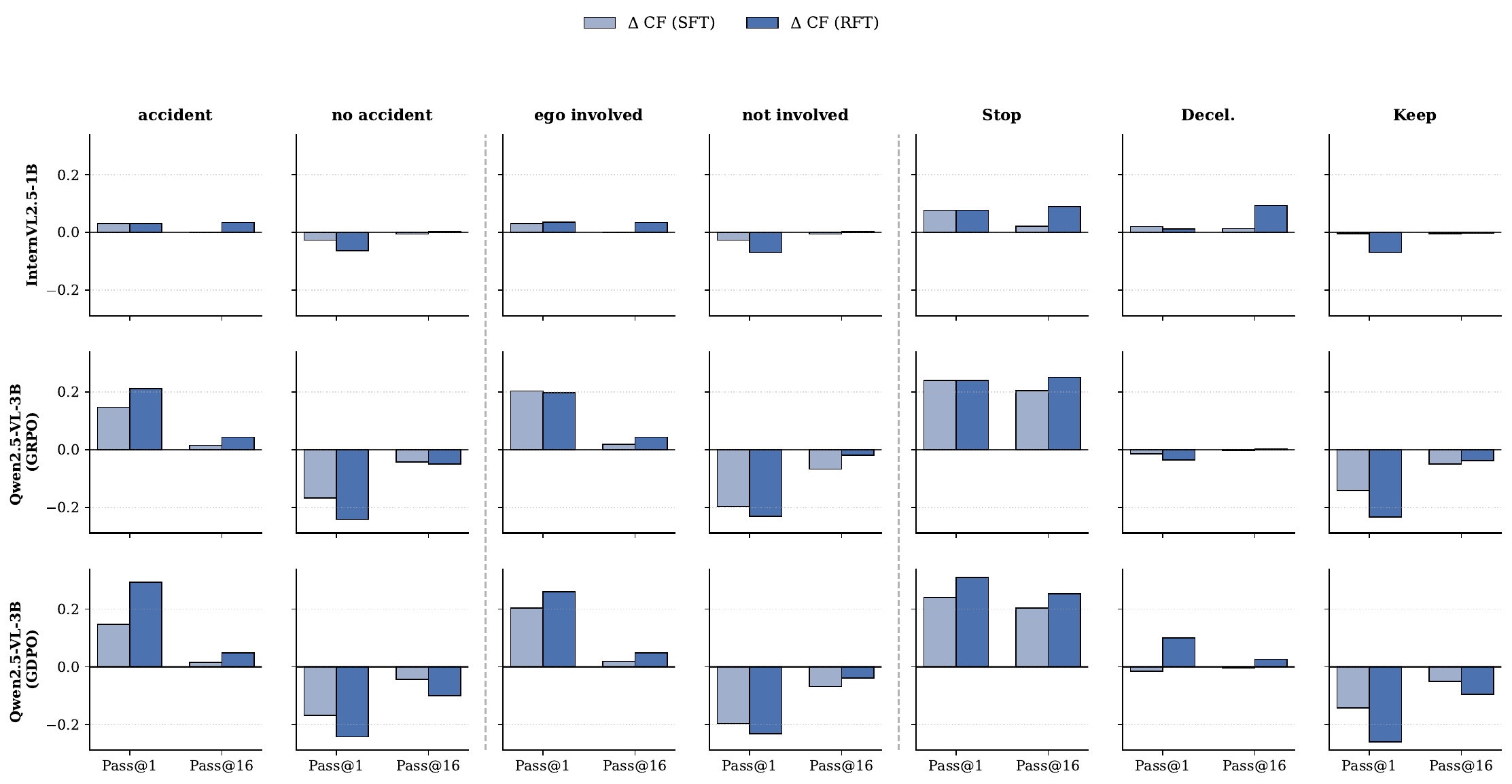}
    \caption{$\Delta$pass@$k$ ($P_{\text{w/ CF}} - P_{\text{w/o CF}}$) for SFT and RFT across varying model scales and tasks, showing persistent gaps in larger models.}
    \label{fig:passk_delta}
\end{figure}

\paragraph{Distribution Shift and Performance Transfer in RFT.}
SFT models served as references for the RFT stage. As illustrated in Fig. \ref{fig:passk_rft}, accident-relevant tasks ("accident", "ego involved") exhibit a significant pass@1 distribution shift, mirroring SFT trends. While this gap converges at $k=16$ for \texttt{InternVL2.5-1B}, larger model like \texttt{Qwen2.5-VL-3B} demonstrate a persistent trade-off between accident sensitivity and non-accident performance.

\paragraph{Amplification of Decision-Making Bias via RFT.}
As shown in Fig. \ref{fig:passk_delta}, RFT amplifies the behavioral patterns established during SFT. For the \texttt{Stop} in \texttt{Qwen2.5-VL-3B}, the $\Delta$pass@$k$ gap between models with and without CF reasoning is notably wider than that observed in the SFT stage. Similarly, for \texttt{Dec.} and \texttt{Keep}, the initial gap at $k=1$ is more pronounced than in SFT, although differences diminish as $k$ approaches 16. This suggests RFT reinforces risk-related priors, leading to a more ingrained safety-bias.

\clearpage
\subsection{Attention Map}
\begin{figure}[h]
    \centering
    \includegraphics[width=\columnwidth]{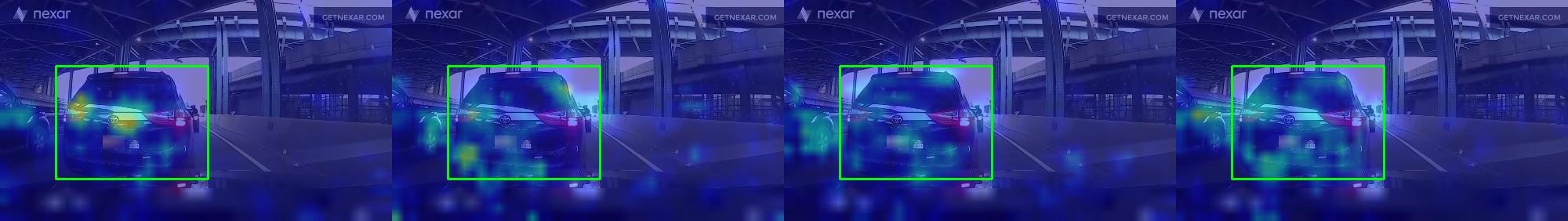}
    \includegraphics[width=\columnwidth]{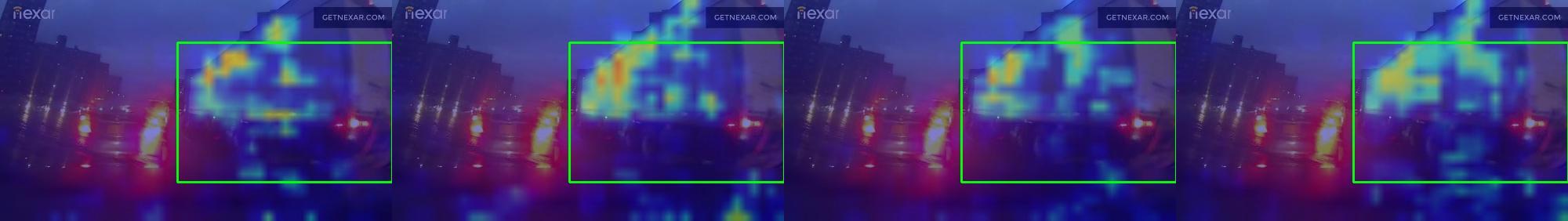}
    \includegraphics[width=\columnwidth]{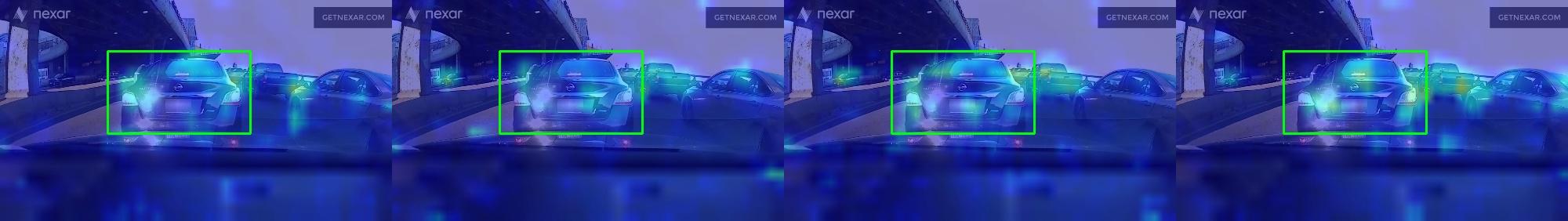}
    \includegraphics[width=\columnwidth]{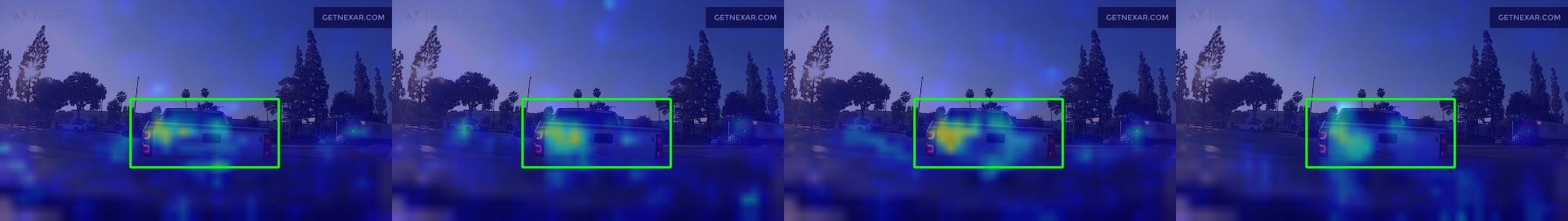}
    \includegraphics[width=\columnwidth]{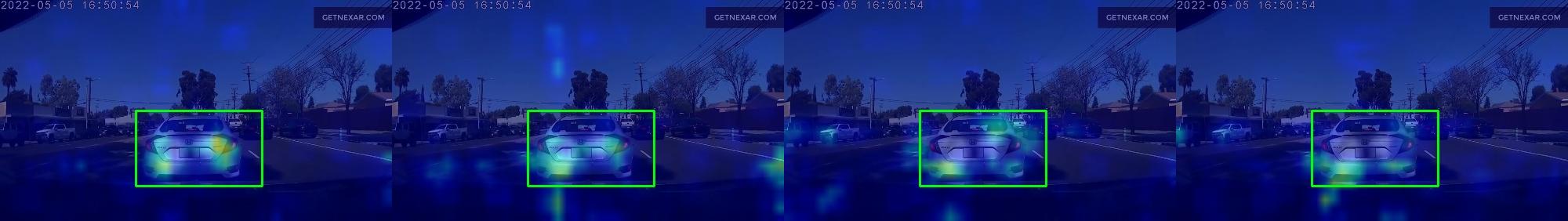}
    \includegraphics[width=\columnwidth]{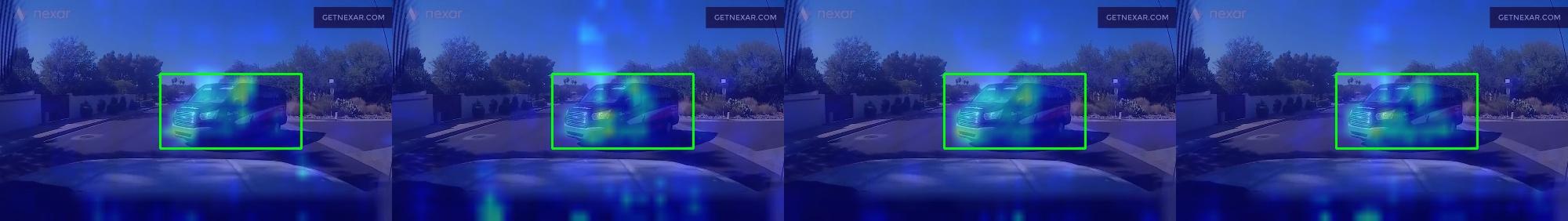}
    \includegraphics[width=\columnwidth]{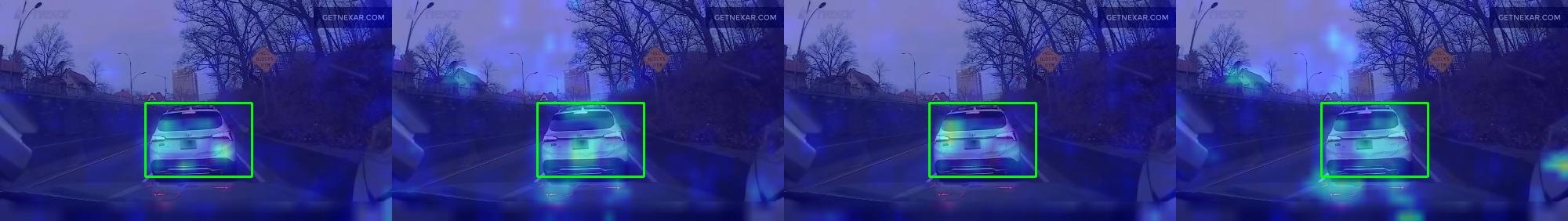}
    \includegraphics[width=\columnwidth]{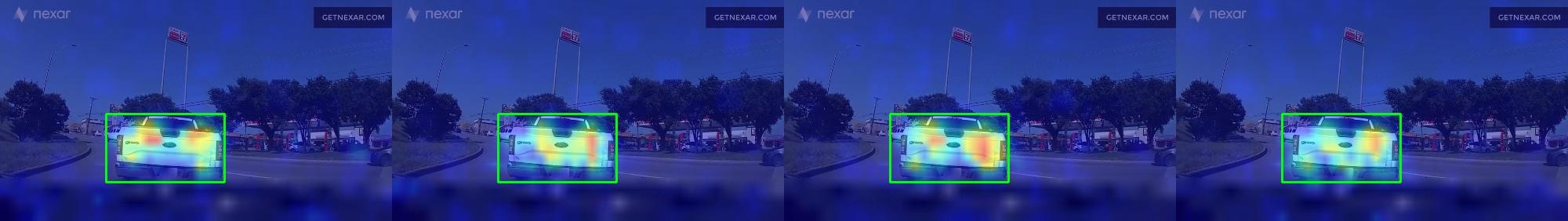}
    \includegraphics[width=\columnwidth]{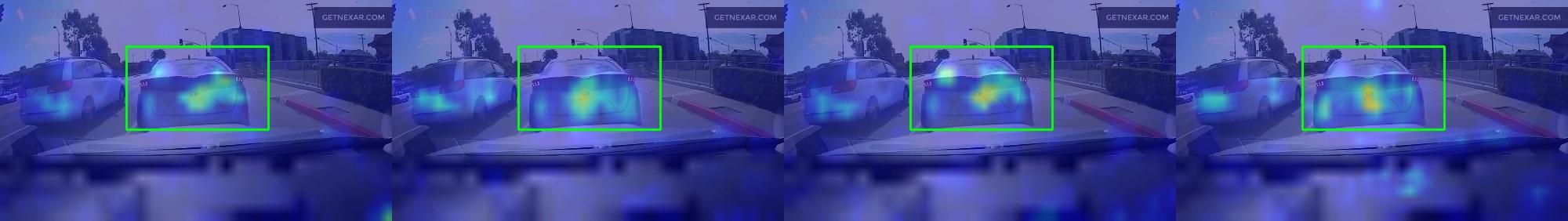}
    \caption{Qualitative comparison of interference activations among SFT w/o CF, SFT w/ CF, RFT w/o CF, and RFT w/ CF based on Qwen2.5-VL-3B. Visualization of activation maps for causal factors immediately before accidents shows no noticeable differences across the models.}
    \label{fig:attention_map}
\end{figure}



\end{document}